\PassOptionsToPackage{hyphens}{url}
\documentclass[pdflatex,sn-nature]{sn-jnl}
\usepackage{graphicx}
\usepackage{multirow}
\usepackage{amsmath,amssymb,amsfonts}
\usepackage{amsthm}
\usepackage{mathrsfs}
\usepackage[title]{appendix}
\usepackage{xcolor}
\usepackage{textcomp}
\usepackage{booktabs}
\usepackage{array}
\usepackage{tabularx}

\newcommand{\USD}{\mathrm{USD}}
\begin{document}

\title[Autonomous agents for electricity planning]{Autonomous agents for electricity-system planning}

\author[1]{Yoon Pyo Lee}
\author[1]{Samrendra Roy}
\author[1]{Jay Phil Yoo}
\author[1]{Kazuma Kobayashi}
\author[1,3,4]{Souvik Chakraborty}
\author*[1,2]{Syed Bahauddin Alam}\email{alams@illinois.edu}

\affil[1]{\orgdiv{Department of Nuclear, Plasma, and Radiological Engineering}, \orgname{The Grainger College of Engineering, University of Illinois Urbana-Champaign}, \orgaddress{\city{Urbana}, \state{IL}, \country{USA}}}
\affil[2]{\orgdiv{National Center for Supercomputing Applications}, \orgname{University of Illinois Urbana-Champaign}, \orgaddress{\city{Urbana}, \state{IL}, \country{USA}}}
\affil[3]{\orgdiv{Department of Applied Mechanics}, \orgname{Indian Institute of Technology Delhi}, \orgaddress{\city{New Delhi}, \country{India}}}
\affil[4]{\orgdiv{Yardi School of Artificial Intelligence}, \orgname{Indian Institute of Technology Delhi}, \orgaddress{\city{New Delhi}, \country{India}}}

\abstract{Planning electricity systems requires repeated decisions that connect regional resources, investment, operation and network constraints. Here we present an autonomous planning framework that carries prepared regional scenarios through proposal, numerical evaluation, feedback interpretation, redesign and final plan selection. Agents propose investment bounds, storage operating conditions and feeder interventions, interpret numerical feedback, and decide how to revise designs and when to stop. The framework automatically executes annual investment-and-dispatch optimization and representative-feeder verification, applies fixed acceptance criteria and retains the lowest-cost accepted candidate. We evaluate forty carbon-constrained trajectories across ten system--year scenarios in five electricity systems and four countries, using two model deployments and single- and multi-agent architectures. All forty retain candidates satisfying the implemented physical and fixed annual operational carbon-budget checks. Twenty-eight of forty matched reference designs exceed the new scenario ceilings. A separate archive of ninety-six trajectories under a common 1,000~Mt annual ceiling provides matched references and broader regional coverage. The results demonstrate autonomous planning under explicit emissions constraints using public regional data and published institutional demand outlooks.}

\keywords{Energy-system planning, Agentic artificial intelligence, Capacity expansion, Carbon budgets, Distribution validation}
\maketitle

\section{Introduction}\label{sec:introduction}

Electricity-system planning is becoming more urgent as AI data centres, electric vehicles, industrial electrification and rising cooling demand accelerate electricity use. The International Energy Agency projects global electricity demand to grow by an average of 3.6\% annually from 2026 to 2030, with data centres accounting for around half of the increase in the United States\cite{iea2026electricity}. This growth coincides with constrained grid connections, ageing infrastructure and increasing exposure to extreme weather. Large, concentrated data-centre loads link computing operations to local grid reliability and infrastructure costs\cite{colangelo2026gridassets}. Decisions about generation, storage and network expansion therefore shape both the reliability of electricity supply and the cost of long-lived infrastructure.

Capacity-expansion research examines these decisions through coupled investment and operation, including the representation of storage\cite{levin2023storage}. Spatial studies coordinate renewable deployment with transmission and storage\cite{wang2023transition}, and recent global models jointly optimize investment and hourly operation at fine spatial resolution\cite{zhu2026netzero}. Local network studies show that electrical constraints and spatially varying flexibility opportunities also affect deployment\cite{few2024localnetworks,brockway2021gridlimits}. An investment proposal changes how the regional system operates over the year. If the resulting dispatch exceeds a network limit, planners must decide whether to revise the portfolio, change storage operation or reinforce the network. The ability to repeat this process matters across countries with different electricity needs and analytical resources. Published work has documented the unequal geographic distribution of expertise in African energy modelling\cite{dioha2023africa}; comprehensive studies such as Net-Zero America illustrate the coordination required across generation and networks\cite{larson2021netzero}.

US offshore wind projections vary with assumptions about technology costs, transmission and policy\cite{beiter2023expanded}. The location of new interregional transmission also affects costs and reliability during extreme events\cite{senga2026transmission}. In western Europe, accounting for the effects of heat on demand and grid capacity reveals transmission bottlenecks\cite{liang2026heatwave}. Evaluating alternatives across regional systems requires repeated investment-and-dispatch optimization, network evaluation and revision of proposed designs. Full-year hourly operation and storage couple decisions across time, while finer temporal detail increases computational demands\cite{hoffmann2022pareto}. Surrogate models trained on numerical simulations can accelerate energy-system optimization\cite{prina2024surrogate}. Completing a planning study also requires interpreting numerical outputs, choosing design changes and deciding when to stop. We use pretrained language models to carry out these planning decisions through repeated numerical feedback.

Open models provide important components of this process. PyPSA optimizes investment and operation, PyPSA-Earth supports geographically extensive model construction, and OpenDSS resolves distribution voltages and currents\cite{brown2018pypsa,parzen2023pypsaearth,epriopendss}. Their integration raises a separate question: can agents autonomously carry a planning study from prepared regional inputs to a verified plan, interpreting physical feedback, revising designs and deciding when to stop?

Language-model agents can use feedback to run power-system simulations\cite{jia2025powersim}. Applications to distribution planning include a workflow with human oversight that evaluates reliability interventions on utility feeders\cite{zhuang2026reliability} and a multi-agent approach to planning photovoltaic generation and electric vehicle charging within distribution networks\cite{shao2026distribution}. Table~\ref{tab:planning-scope} compares the decisions addressed in these studies with those examined in recent electricity-system planning research.

Here we develop and evaluate an autonomous electricity-planning framework that connects regional investment and annual operation with representative-feeder verification. To our knowledge, no previous study has demonstrated a planning process across multiple regional electricity systems in which language-model agents revise investment proposals using feedback from both year-round dispatch and distribution network checks. The framework carries the planning process from prepared inputs through feedback-driven redesign and agent-directed stopping to final plan selection across heterogeneous regional and future conditions.

\begin{table}[tbp]
\centering
\caption{Planning decisions and scope of electricity-system studies and agent-assisted distribution planning.}
\label{tab:planning-scope}
\small
\begin{tabularx}{\linewidth}{@{}>{\raggedright\arraybackslash}p{0.21\linewidth}
  >{\raggedright\arraybackslash}X >{\raggedright\arraybackslash}X@{}}
\toprule
Study & Planning decision addressed & Scope demonstrated \\
\midrule
\multicolumn{3}{@{}l}{\textit{Electricity-system studies in Nature Energy}} \\
\addlinespace[2pt]
Zhu et al.\cite{zhu2026netzero} & How much generating capacity is needed to meet global electricity needs under net-zero pathways? & Joint capacity expansion and hourly operation in a spatially resolved global model. \\
Few et al.\cite{few2024localnetworks} & Where can local flexibility reduce network upgrades? & Network simulations quantify how upgrade costs and the value of flexibility vary across Great Britain. \\
Beiter et al.\cite{beiter2023expanded} & How much offshore wind is built under different planning assumptions? & US expansion scenarios quantify deployment across cost, transmission and policy assumptions. \\
Senga et al.\cite{senga2026transmission} & Where should interregional transmission expand? & US expansion strategies are compared on cost, emissions and reliability during extreme events. \\
Liang et al.\cite{liang2026heatwave} & Where does heat limit grid capacity? & Weather-dependent demand and equipment limits reveal transmission bottlenecks in western Europe. \\
\midrule
\multicolumn{3}{@{}l}{\textit{Agent-assisted distribution planning}} \\
\addlinespace[2pt]
Zhuang and Sadikovic\cite{zhuang2026reliability} & Which feeder interventions improve reliability? & Agent proposals undergo numerical evaluation with human oversight on 100 utility feeders. \\
Shao et al.\cite{shao2026distribution} & How should a distribution network accommodate solar generation and electric vehicle charging? & Agents negotiate stakeholder objectives within a search over network capacity and topology. \\
\midrule
\textbf{This study} & How can agents carry a regional scenario through design, evaluation and revision to a verified plan? & Agents revise proposals using annual dispatch and feeder checks. Across five systems all 40 carbon-constrained searches retain an accepted design. \\
\bottomrule
\end{tabularx}
\end{table}

Once a regional scenario is supplied, the framework carries out the planning loop autonomously. Agents interpret regional inputs and numerical feedback to choose technology permissions, investment bounds, storage operating conditions and feeder interventions. PySAM converts regional weather inputs into hourly photovoltaic generation profiles. A turbine power curve supplies the wind profiles. PyPSA uses these profiles to select investments and hourly dispatch within the proposed bounds. OpenDSS then screens the dispatch on representative feeders. The framework automatically executes the required system and feeder evaluations, returns their evidence to the agents and applies fixed acceptance criteria. Agents use this feedback to choose subsequent design changes and decide whether to continue or stop. The framework preserves the lowest-cost accepted candidate found so far in a separate reporting record and returns it as the final result. This adaptation takes place at inference time, without task-specific training or fine-tuning of model weights. Single agents maintain strategy notes, whereas specialist teams additionally maintain an advisory procedure graph inspired by procedural memory\cite{lu2026procedural}.

The primary evaluation comprises ten system--year scenarios across NYISO, MISO, South Africa, Mexico and Brazil. These five systems cover four countries and forty model zones. Each scenario is evaluated with GPT-6.1 Sol and Claude Opus 5.5 in both single- and multi-agent configurations, giving forty trajectories. The scenarios comprise NYISO in 2023, 2030 and 2035; MISO, South Africa and Brazil in 2030 and 2035; and Mexico in 2030. Four trajectories use 2023 demand and ERA5 resources, while thirty-six use future demand and resource profiles reconstructed for the matching target year from NASA NEX-GDDP-CMIP6 daily fields and ERA5 intraday templates. Published allowance budgets, planning constraints and emissions pathways define fixed annual operational CO$_2$ ceilings under the mappings in Methods. A separate archive contains ninety-six trajectories across eight systems under the original 1,000~Mt annual ceiling. Forty of those trajectories match the primary cohort by system, year, deployment and architecture, while the remaining fifty-six provide broader regional and temporal coverage.

Published expert plans provide a reference for interpreting the carbon-constrained portfolios in Brazil and South Africa in 2035. We compare technology-specific capacities where archived documents permit an explicit year and category mapping. The plans also supply some scenario inputs, including the adopted emissions ceilings, so the comparison describes portfolio differences under partially shared assumptions.

Across all forty carbon-constrained trajectories the agents completed the planning cycle and the framework retained a design that passed the implemented system, feeder and emissions checks. The separate loose-cap archive shows how agents revised their proposals and when feeder checks ruled out a lower-cost alternative.

\section{Results}\label{sec:results}

\subsection{Autonomous agents complete the planning cycle}

\begin{figure}[tbp]
\centering
\makebox[\textwidth][c]{\includegraphics[width=180mm]{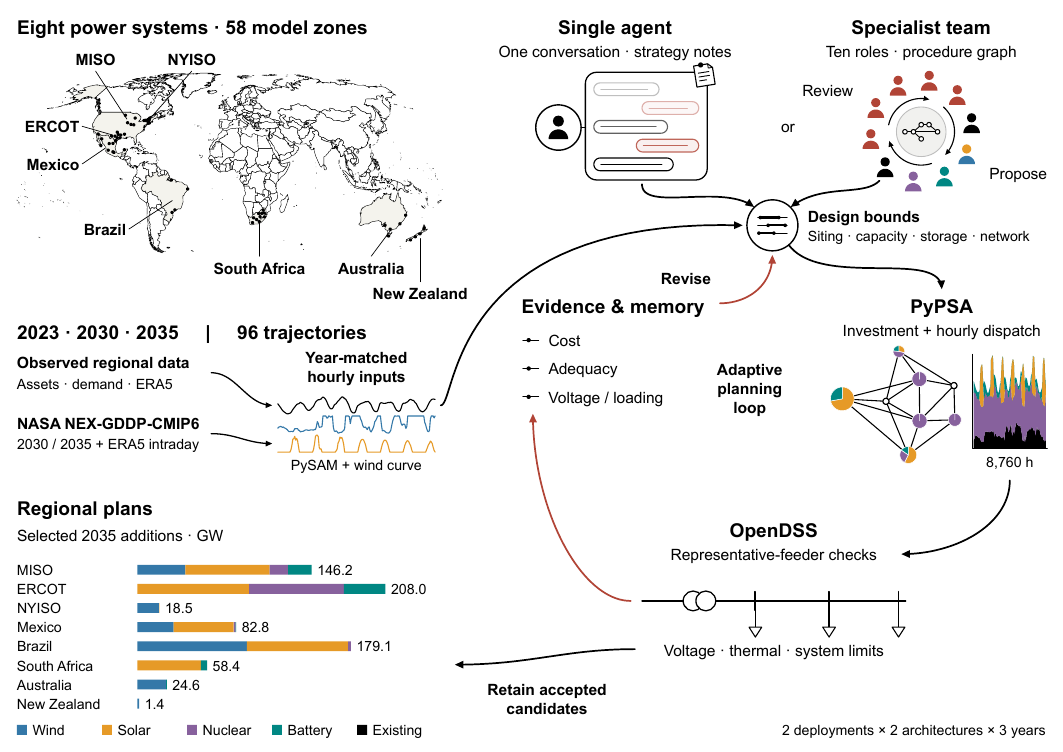}}
\caption{\textbf{Reality-grounded planning as an adaptive experimental loop.} The broader input library covers eight systems and 58 model zones. Regional records, inherited assets, network assumptions, costs and demand outlooks define scenarios. ERA5 supplies 2023 weather; NASA projections for 2030 and 2035 are reconstructed with ERA5 intraday templates. PySAM and a turbine curve produce power profiles. Agents propose design bounds through a single conversation with strategy notes or a ten-role specialist team. Specialists propose, an orchestrator commits one experiment, and reviewers evaluate, critique, refine a procedure graph, compress memory and judge convergence. PyPSA optimizes investment and annual dispatch; OpenDSS screens representative feeders. Evidence returns to guide subsequent experiments, while a separate record retains the lowest-cost accepted committed candidate. In the primary evaluation, PyPSA also enforces an immutable annual operational CO$_2$ budget. The ERCOT traces and schematic, regional 2035 additions and 96-trajectory count illustrate the loose-cap reference evaluation under the original 1,000~Mt ceiling (Tables~\ref{tab:complete-coverage} and~\ref{tab:all-selected-costs}). Regional bars select the lowest observed accepted cost across four reference trajectories per system. Black denotes inherited assets and output; colours identify added technologies. Connections have model-zone resolution and the feeder is schematic. Primary carbon-constrained outcomes are reported in Fig.~\ref{fig:carbon-outcomes}. The basemap uses Natural Earth\cite{naturalearth50m}.}
\label{fig:framework}
\end{figure}

The framework assigns distinct roles to resource, system and distribution models (Fig.~\ref{fig:framework}). PySAM converts irradiance to photovoltaic output, and a reference turbine curve converts wind speeds to generation potential. PyPSA selects additions and hourly dispatch within agent-specified technology permissions, siting bounds and storage operating envelopes. OpenDSS evaluates that dispatch on representative feeders. Agents use the resulting evidence to choose the next experiment.

In the primary evaluation, the annual operational CO$_2$ budget is a fixed scenario input enforced within each system optimization. The agent cannot relax this budget. Acceptance also checks the reported emissions against the fixed ceiling, alongside the system and representative-feeder requirements. The budget constrains direct operational emissions within the model boundary and does not establish compliance with the full institutional policy or planning framework from which its numerical value was drawn.

The single-agent architecture maintains one conversation and strategy notes. The multi-agent architecture separates specialist proposals, orchestration, evaluation, criticism, procedure refinement, compression and convergence judgement. Both architectures receive system and feeder evidence for every successful committed solve. In the multi-agent architecture, hypothetical system-only verifications support the judge's decision to continue or stop within the current cycle. A candidate is physically accepted only after its committed system and distribution results satisfy every applicable physical and carbon-budget check. Throughout the search, a separate reporting record retains the lowest all-in-cost committed candidate that has passed every applicable physical and carbon-budget check, without changing the agents' prompts or stopping decisions. This candidate is reported when the search ends through an agent stopping decision or the configured search ceiling.

Every search is initialized with the engine's default design, which permits all extendable technologies, sets capacity bounds to their maxima and uses four-hour batteries without feeder interventions. Agents may modify this design before the first committed simulation. In the primary cohort, every system optimization includes the scenario-specific carbon budget. Agents then revise design restrictions and feeder controls using numerical feedback. Final selection retains the lowest all-in-cost committed candidate satisfying every implemented physical check and the fixed carbon budget. This selection rule is applied independently of the last proposal evaluated and the agent's stopping decision.

\subsection{Planning outcomes under fixed annual carbon budgets}\label{sec:complete-results}
The primary cohort contains forty completed trajectories across ten system--year scenarios. All forty retain an accepted committed candidate under the implemented system, representative-feeder and carbon-budget checks. Each retained result is feasible within its modelled scenario and stated tolerances. Numerical optimality bounds apply to the optimizer's domain for that candidate, rather than to every design the agent could have proposed.

The budget is active to within $10^{-5}$~Mt in twenty-eight retained outcomes across seven scenarios. NYISO 2023, Mexico 2030 and South Africa 2030 retain emissions below their budgets in all four configurations. Complete outcome tables retain all four deployment--architecture trajectories for every scenario. Portfolio and geographic illustrations select the lowest observed all-in annual cost among the four accepted candidates for each displayed system and year. GPT-6.1 Sol single-agent supplies all ten displayed portfolios, attaining the unique observed minimum in six scenarios and a minimum shared with another configuration in four, using a USD~10 annual-cost tie tolerance. These illustrations describe achieved designs; they do not establish that one deployment or architecture is generally superior. Comparisons of deployment or architecture retain every matched pair and the candidate-specific numerical bounds.

\begin{figure}[tbp]
\centering
\makebox[\textwidth][c]{\includegraphics[width=170mm]{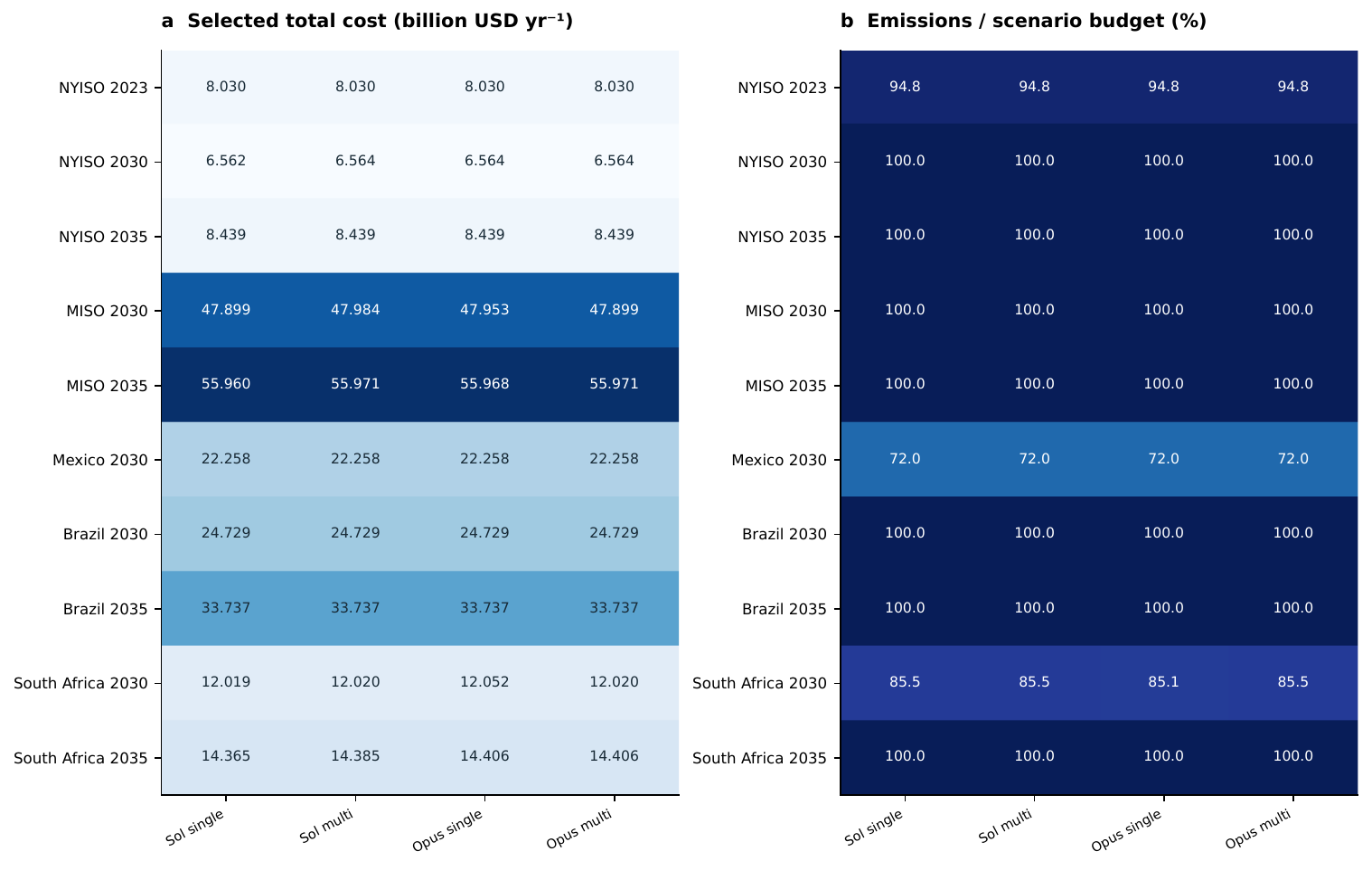}}
\caption{\textbf{Complete outcomes under annual operational carbon budgets.} The ten system--year scenarios each contain two model deployments and two agent architectures. All forty retained outcomes satisfy the implemented physical and carbon-budget checks. Selected annual cost includes representative-distribution interventions. Emissions are direct operational CO$_2$ expressed relative to the corresponding scenario budget. A budget ratio of one denotes an outcome at the ceiling within numerical tolerance. The budgets are research constraints adopted from sources with the distinctions listed in Table~\ref{tab:carbon-scenarios}.}
\label{fig:carbon-outcomes}
\end{figure}

Figure~\ref{fig:carbon-outcomes} reports costs and emissions relative to the fixed budgets for all forty outcomes. Table~\ref{tab:carbon-selected-costs} retains each selected cost, and Fig.~\ref{fig:carbon-bounds} shows candidate-specific numerical bounds.

\begin{tableorg}[tbp]
\caption{\textbf{Selected annual costs in all forty carbon-constrained trajectories.} Values are billion USD per year, including annualized representative-feeder interventions, rounded to six decimals. Each entry is one retained accepted committed candidate. S and M denote single- and multi-agent architectures. All four trajectories in every row pass the physical and fixed carbon-budget checks. Full precision, emissions, solver bounds and source identities accompany the archived numerical data.}
\label{tab:carbon-selected-costs}
\centering\small
\makebox[\textwidth][c]{\begin{tabular*}{170mm}{@{\extracolsep{\fill}}lrrrrrr@{}}
\toprule
System & Year & Sol S & Sol M & Opus S & Opus M & Accepted \\
\midrule
NYISO & 2023 & 8.030088 & 8.030121 & 8.030136 & 8.030121 & 4/4 \\
NYISO & 2030 & 6.561827 & 6.564198 & 6.564198 & 6.564198 & 4/4 \\
NYISO & 2035 & 8.438767 & 8.438767 & 8.438767 & 8.438767 & 4/4 \\
MISO & 2030 & 47.898514 & 47.984071 & 47.953352 & 47.898514 & 4/4 \\
MISO & 2035 & 55.959875 & 55.970535 & 55.968389 & 55.970535 & 4/4 \\
South Africa & 2030 & 12.019165 & 12.019593 & 12.051896 & 12.020077 & 4/4 \\
South Africa & 2035 & 14.365241 & 14.384747 & 14.405887 & 14.406203 & 4/4 \\
Mexico & 2030 & 22.257606 & 22.258090 & 22.258090 & 22.258090 & 4/4 \\
Brazil & 2030 & 24.729291 & 24.729291 & 24.729291 & 24.729291 & 4/4 \\
Brazil & 2035 & 33.736655 & 33.736655 & 33.736655 & 33.736655 & 4/4 \\
\bottomrule
\end{tabular*}}
\end{tableorg}

\subsection{Accepted carbon-constrained portfolios retain their geographic context}\label{sec:carbon-geographic-results}

\begin{figure}[p]
\centering
\makebox[\textwidth][c]{\includegraphics[width=170mm]{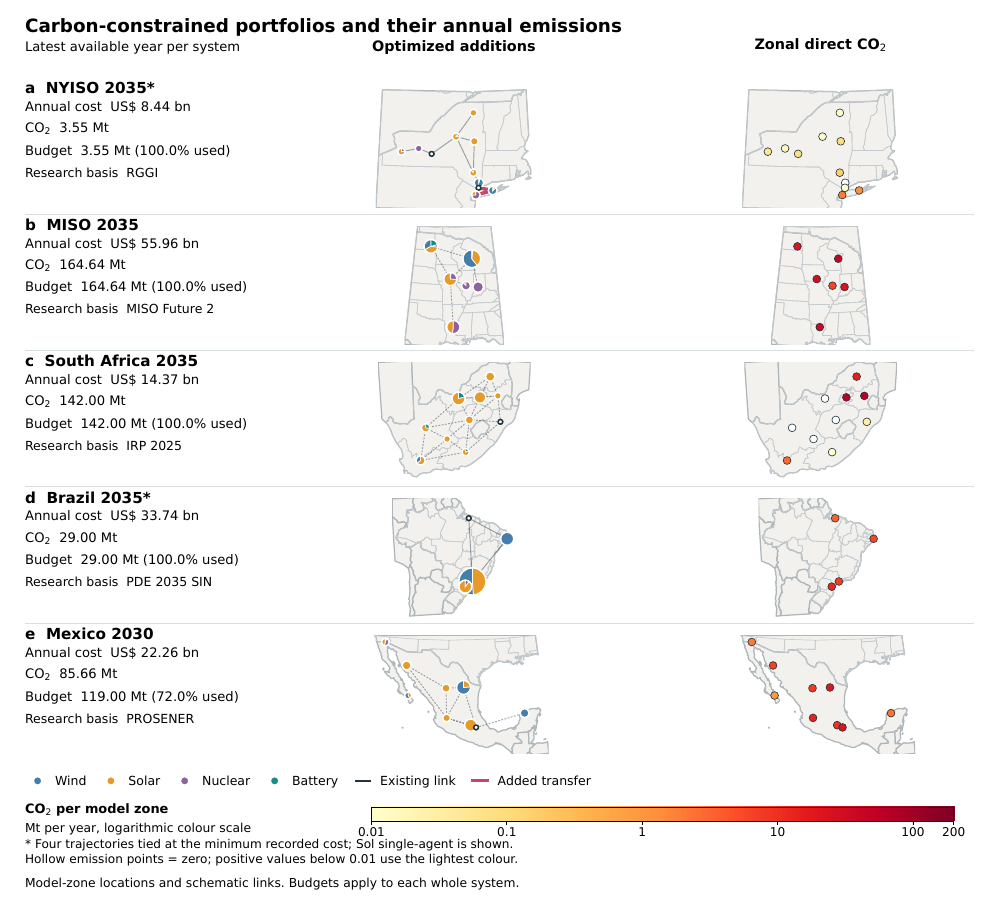}}
\caption{\textbf{Optimized grid additions and operational carbon emissions.} Paired maps show the minimum observed accepted-cost design for each system's latest evaluated year. GPT-6.1 Sol single-agent supplies all representatives; tied minima and years are marked. Pie colours distinguish technology additions; grey and magenta links show existing connections and selected transfer increments. Fixed-size emission circles use a shared logarithmic colour scale; hollow circles denote zero. Each row reports annual cost, direct operational CO$_2$, the whole-system budget and its source. Budget values and source-to-model mappings are in Table~\ref{tab:carbon-scenarios}. Emissions are modelled from retained dispatch and declared inventory factors. Locations represent model zones. The ten-case atlas retains all years. The basemap uses Natural Earth\cite{naturalearth50m}.}
\label{fig:carbon-geography}
\end{figure}

For each of the ten system--year scenarios, we select the lowest observed all-in annual cost among four candidates satisfying the physical and carbon-budget checks. GPT-6.1 Sol single-agent supplies each representative, with tied minima identified. Figure~\ref{fig:carbon-geography} pairs the resulting grid additions with operational emissions for the latest evaluated year in each system. NYISO, MISO, South Africa and Brazil use 2035, while Mexico uses 2030. The years are explicit because these panels do not form a common-year comparison. All ten scenarios remain in the outcome tables and the detailed atlas.

\begin{figure}[tbp]
\centering
\makebox[\textwidth][c]{\includegraphics[width=170mm]{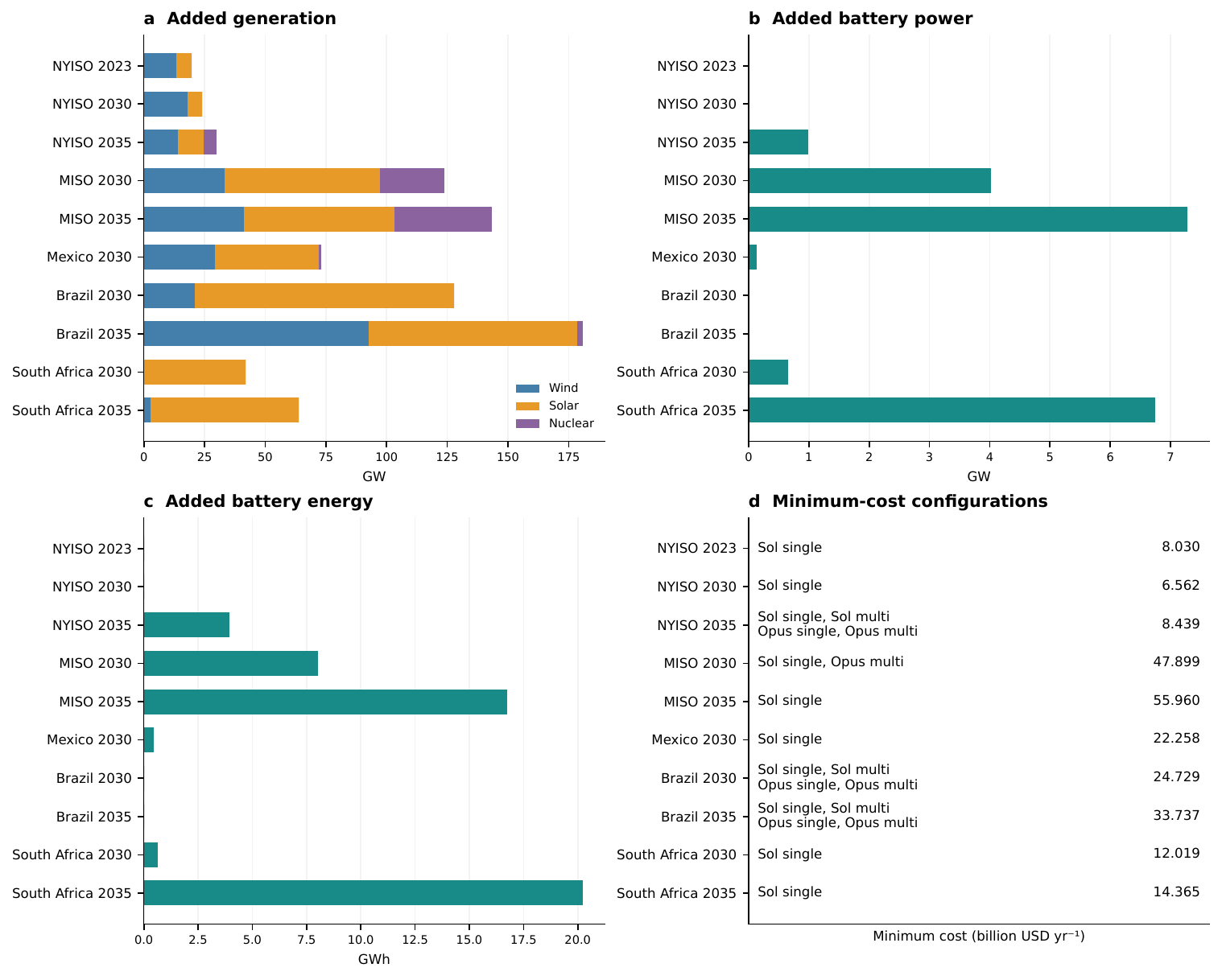}}
\caption{\textbf{Minimum observed accepted portfolios under carbon budgets.} Each system--year scenario shows the portfolio with the lowest observed all-in annual cost among its four accepted trajectories. Six scenarios have a unique observed minimum and four have tied minima within a USD~10 annual-cost tolerance. Panels a--c display the GPT-6.1 Sol single-agent representative, while panel d lists every deployment--architecture configuration attaining the observed minimum within this tolerance. Generation and storage values describe additions to the inherited fleet, with battery power and energy distinguished. Selection is conditional on the system, representative-feeder and carbon-budget checks. The complete comparisons retain all forty trajectories and their numerical bounds.}
\label{fig:carbon-portfolios}
\end{figure}

The paired maps connect the selected investments to the annual emissions of the retained dispatch. Each panel reports total emissions against its fixed whole-system budget; no zonal emissions ceiling is imposed. Figure~\ref{fig:carbon-portfolios} separately distinguishes added generation, battery power and battery energy across all ten scenarios. The detailed atlas provides the zonal capacity and emissions ledgers. These allocations have model-zone resolution.

\subsection{Carbon-constrained portfolios differ from published planning pathways}\label{sec:expert-comparison}

We compare the four retained accepted carbon-constrained outcomes for each of Brazil and South Africa in 2035 with technology-specific capacities from PDE 2035 and IRP 2025 (Table~\ref{tab:expert-capacity}). These eight candidates satisfy their physical checks and annual model emissions budgets of 29 and 142~MtCO$_2$, respectively. The comparisons use total capacity, including inherited assets, and retain all four deployment--architecture configurations.

Brazil's four outcomes contain 120.19~GW of wind, compared with 49.55~GW in PDE 2035. Their 96.66~GW of model solar is compared with the plan's 29.19~GW centralized-solar category, which excludes a further 78.10~GW of micro-/mini-distributed generation. The model does not represent a separate future distributed-generation rollout, so this category-level difference does not measure a difference in total national photovoltaic capacity. The plan includes 6.63~GW of batteries, whereas all four carbon-constrained outcomes contain zero battery capacity.

South Africa's outcomes contain 5.57--6.24~GW of wind and 63.02--64.24~GW of solar. The IRP schedule yields 26.39~GW of wind and 21.16~GW in its Solar PV column by 2035, with 16.07~GW in a separate distributed-generation column. The portfolios therefore retain a different technology balance while satisfying the adopted annual emissions ceiling. The numerical ceilings share institutional sources with the capacity references, while resource, retirement, technology and delivery assumptions remain distinct. These comparisons describe the resulting portfolios rather than validate them against withheld planning targets.

\begin{table}[tbp]
\caption{Published plans and carbon-constrained portfolios in 2035. Values are total capacities in GW, including inherited assets. Carbon ranges span all four retained accepted deployment--architecture trajectories per system. Differences are carbon-constrained agent values minus the published plan, calculated before rounding. Brazil and South Africa use annual model budgets of 29 and 142~MtCO$_2$, respectively.}
\label{tab:expert-capacity}
\centering\small
\setlength{\tabcolsep}{5pt}
\begin{tabularx}{\textwidth}{@{}lXrrr@{}}
\toprule
System & Capacity & Expert plan & Carbon range & Difference \\
\midrule
Brazil & Wind & 49.55 & 120.19 & +70.64 \\
Brazil & Solar$^\dagger$ & 29.19 & 96.66 & +67.47 \\
Brazil & Battery power & 6.63 & 0.00 & -6.63 \\
South Africa & Wind & 26.39 & 5.57--6.24 & \([-20.82,-20.14]\) \\
South Africa & Solar$^\dagger$ & 21.16 & 63.02--64.24 & \([+41.86,+43.08]\) \\
\bottomrule
\end{tabularx}
\smallskip
\par\footnotesize $^\dagger$ Brazil's PDE 2035 centralized-solar category excludes 78.099~GW of micro-/mini-distributed generation. South Africa's IRP 2025 Solar PV column excludes 16.065~GW in its separate distributed-generation column, CSP and hybrid capacity. Model solar has no separate future distributed-generation rollout. The Brazilian values describe December 2035; the South African values combine the end-2025 base and scheduled additions through 2035. Demand, resources, inherited fleets, retirements, technology options and delivery schedules retain their respective study assumptions. Source locations and mappings are given in Methods and Appendix~\ref{app:regional-sources}.
\end{table}

\subsection{Recorded effort in the carbon-constrained searches}\label{sec:carbon-search-process}

The forty carbon-constrained trajectories contain 202 logical system-evaluation requests, comprising 130 committed simulation requests and 72 hypothetical system-only verification requests. These counts include requests served through exact-result reuse. The archived summaries record 764 received model responses, 90,562,224 input tokens and 1,314,432 output tokens. Input includes reported cache reads and writes, and output includes the reported reasoning subset. Provider usage conventions, recovery histories and the result-reuse revision remain attached to the trajectory evidence. The counts describe search effort rather than a controlled wall-runtime benchmark.

\begin{figure}[tbp]
\centering
\makebox[\textwidth][c]{\includegraphics[width=170mm]{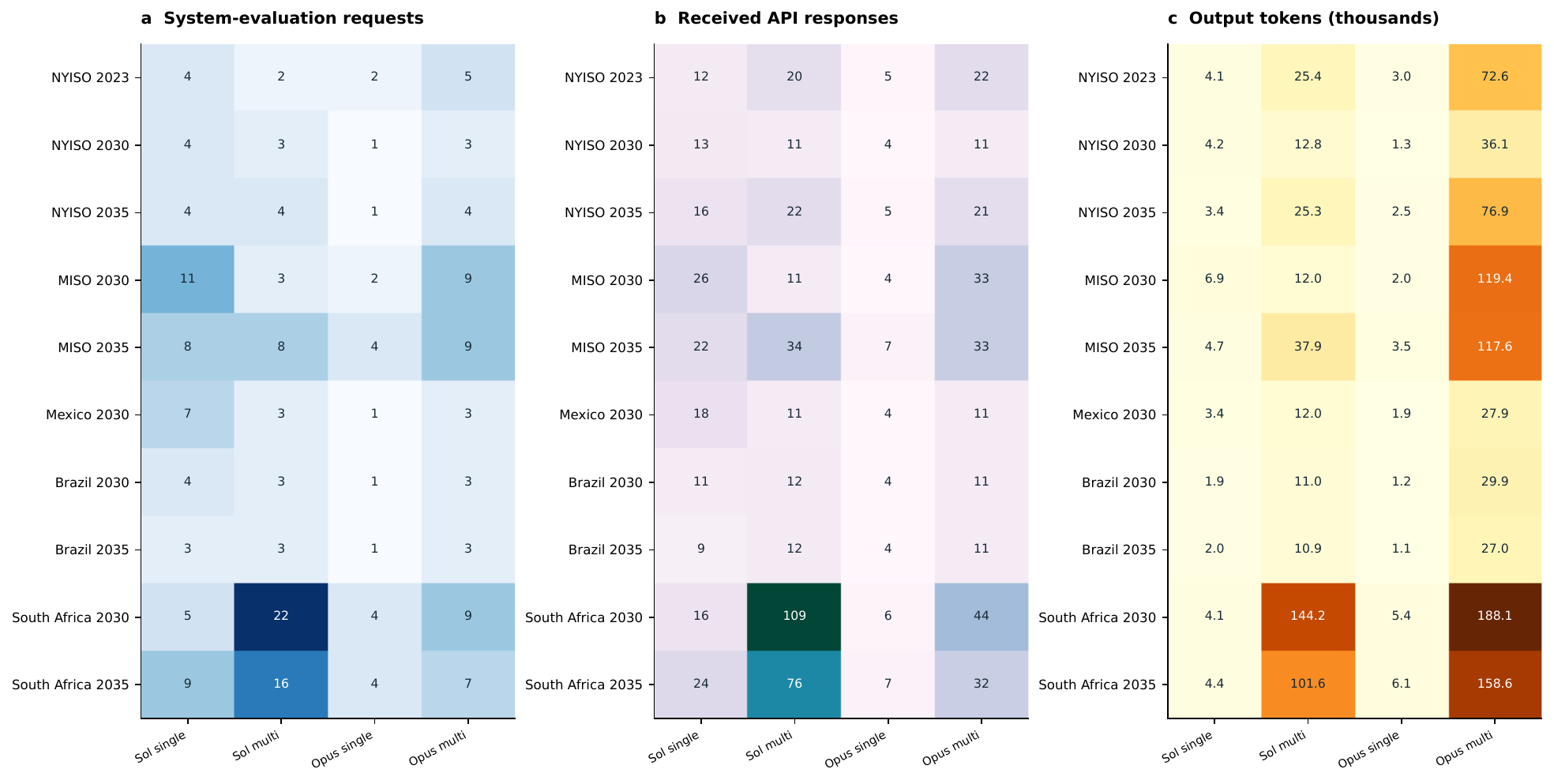}}
\caption{\textbf{Recorded effort in the carbon-constrained planning searches.} Logical system-evaluation requests, received API responses and provider-reported tokens describe different components of search effort. The forty trajectories contain 202 logical system-evaluation requests and 764 recorded received responses, with 90.562 million input tokens and 1.314 million output tokens. Logical requests include exact-result reuse and therefore are not counts of newly executed numerical optimizations. Recovery histories and the recorded reuse revision are retained in the evidence. These summaries do not constitute a wall-runtime or scaling benchmark.}
\label{fig:carbon-search}
\end{figure}

Figure~\ref{fig:carbon-search} retains the counts for every deployment--architecture--scenario trajectory. Received responses, logical evaluation requests and newly executed numerical optimizations measure different components of effort.

\subsection{Loose-cap reference scenarios distinguish imposed limits from observed emissions}\label{sec:carbon-ablation}
The original cohort used an annual CO$_2$ ceiling of 1,000~Mt. We retain its ninety-six trajectories as a separate reference dataset and compare the forty trajectories matching the primary cohort by system, year, deployment and architecture. Twenty-eight of these reference designs exceed the tighter budgets, corresponding to seven scenarios. The comparison reports achieved annual cost, operational emissions and investment portfolios for each matched pair. In NYISO 2035, the selected annual cost is USD~8.439 billion in all four carbon-constrained configurations, 23.56\% above each matching loose-cap outcome. These observed contrasts are conditional on the archived trajectories and their respective optimization domains. A source-derived ceiling need not change the selected outcome when the corresponding loose-cap design already emits below it. Differences in binding and non-binding scenarios must therefore be distinguished when interpreting the response to a tighter budget.

These are comparisons of archived agent trajectories under different scenario inputs. The 2023 Opus reference used a 16,000-token output ceiling and an earlier provider adapter, whereas the primary campaign used a 64,000-token ceiling and the updated adapter. Those two NYISO pairs are reported separately from the other matched pairs. Execution revisions and exact-result reuse are also recorded, and one trajectory per condition does not support estimates of stochastic variation or an isolated causal effect of the carbon ceiling on the full agent search.

Figure~\ref{fig:carbon-reference} shows the matched cost and emissions contrasts. Across the full loose-cap archive, 36 assessments in 16 searches had a lower annual objective than a previously accepted design but failed a feeder check. In one ERCOT 2035 search, removing a proposed line upgrade reduced the annual objective by 0.220\% but caused three hours of undervoltage. The agent team restored the upgrade in the next round and the framework retained the earlier accepted design (Fig.~\ref{fig:ercot-verified} and Appendix~\ref{app:loose-cap-results}).

\section{Discussion}\label{sec:discussion}

The central result is autonomous execution of a coupled electricity-planning process across heterogeneous regional systems. Starting from prepared inputs, agents interpret numerical feedback, revise investment and operating decisions, and determine when to end the search. Automated system and representative-feeder evaluations, acceptance checks and final selection carry these decisions through to retained plans. All forty primary trajectories retain candidates satisfying the implemented physical and carbon-budget checks. Twenty-eight matched loose-cap reference designs exceed the new scenario ceilings. The budgets are enforced by the optimizer; agents navigate design choices and interpret system and representative-feeder evidence within these fixed constraints.

Comparisons between the carbon-constrained designs and published plans make the interpretation of numerical acceptance more concrete. Brazil's larger modelled wind fleet and South Africa's solar-heavy portfolio persist under the adopted emissions ceilings. Satisfying a plan-derived annual carbon budget therefore does not reproduce the plan's technology mix. Model diagnostics document systematic differences in how energy models respond to scenario assumptions\cite{dekker2023fingerprints}; transmission studies also expose trade-offs among cost, emissions and reliability under policy-driven and cost-minimizing designs\cite{senga2026transmission}. Evidence from transition modelling further identifies roles for social acceptance, investment risk and infrastructure lock-in\cite{fischromito2025societal}. The portfolio differences identify retirement, resource, technology and staged-delivery assumptions for further sensitivity analysis.

Investment timing and available technologies remain important model boundaries. Apart from storage and the hydrogen chain, nuclear is the only new dispatchable technology in the simulator configuration; gas, coal and hydro plants exist only as inherited capacity. The single-period formulation makes selected additions available without a construction schedule, financing sequence or permitting constraint, while the inherited fleet remains available throughout the modelled year. Investment-pathway studies show why foresight and deployment timing matter for the transition\cite{heuberger2018myopic,wang2023transition}.

The carbon-constrained cohort adopts numerical limits from several kinds of institutional evidence. Allowance budgets, planning constraints and projected emissions are not interchangeable policy instruments. Our source-to-model mappings turn these quantities into fixed annual model constraints, while retaining differences in geographical coverage, emissions accounting, inherited assets and resource assumptions. Mexico and Brazil additionally use published CO$_2$-equivalent values as numerical proxies for direct operational CO$_2$. These choices define research scenarios and do not demonstrate legal compliance, lifecycle decarbonization or reproduction of the complete institutional plans.

Local data preparation, source interpretation and the predefined simulator interfaces remain essential expert contributions. Evaluation with regional practitioners would extend the study to systems with poorer network records, constrained financing or substantial unmet demand.

Further development can address network detail, climate uncertainty and replication. Network evaluation uses reduced representative feeders and a zonal transport formulation. The fixed annual reliability tolerance permits limited unmet demand. Detailed AC network, contingency and site-level analyses would extend this representation.

Weather-aware studies have coupled planning and operation across weather years or optimized under climate uncertainty and extremes\cite{zeyringer2018weather,perera2020extremes}. Recent work further connects climate-sensitive adequacy to renewable siting and system flexibility\cite{qiu2026siting,zheng2025climateresilient}. Our future weather uses one climate model, one emissions scenario and synthetic hourly chronology. The change of data product between 2023 and the future years is not separated from projected climate change. A same-GCM historical control, multiple weather realizations and a climate-model ensemble would extend the analysis to climate attribution and robustness.

Each deployment--architecture--scenario cell contains one trajectory, and campaign inference settings differ. Solver bounds apply to each candidate-specific optimization problem.

A controlled expert comparison would hold inputs, technology options, reliability criteria and computational resources fixed, then evaluate both expert-guided and agent-guided candidates through the same independent gate. A complementary ablation would compare the adaptive loop with a predefined sequence of system and feeder checks. These tests would separate the contribution of adaptive decisions from the numerical models and data preparation.

The present results demonstrate autonomous completion of electricity-planning studies under fixed operational carbon budgets, grounded in public regional data and institutional demand outlooks, with the reference archive extending regional and temporal coverage. Explicit regional assumptions, committed dispatch, rejected and accepted candidates, and recorded search histories make both the resulting plans and the decisions leading to them inspectable.

\section{Methods}\label{sec:methods}

\subsection{Regional systems and data basis}

The primary evaluation covers five electricity systems in four countries, drawn from the broader eight-system, six-country input library in Table~\ref{tab:regions}; the loose-cap reference evaluation covers all eight systems. Each system is represented as a coupled zonal planning problem in brownfield mode, with the inherited generation and storage fleet available for dispatch alongside optimized additions.

National data sources include EIA for the United States\cite{eiagridmonitor}, ESRG for South Africa\cite{merven2025demand}, AEMO for Australia\cite{aemonemweb}, EMI for New Zealand\cite{emi2026}, SENER for Mexico\cite{sener2024prodesen}, and ONS for Brazil\cite{onsopendata}. Specific source editions for demand, fleet, network and forecast inputs and expert-plan comparisons are listed with direct links in Appendix~\ref{app:regional-sources}.

The US demand inputs contain 2023 hourly observations for six MISO reporting groups, eight ERCOT weather zones and eleven NYISO zones. MISO's groups are LRZ 1, LRZ 4, LRZ 6, LRZs 2+7, LRZs 3+5 and LRZs 8+9+10. Existing generation is drawn from EIA-860 and mapped using geographic and identifier records, including NYISO's generator-zone assignments.

South Africa uses the ESRG demand model and its spatialized reference-scenario outputs. The ten electricity supply areas are Eastern Cape, Limpopo, Mpumalanga, Gauteng, Western Cape, Northern Cape, Hydra Central, KwaZulu-Natal, North West and Free State. The supply-area partition assigns Hydra Central separately. The published series contains 8,760 hourly model outputs per year for 2023--2050, informed by Eskom regional transformer data and national demand projections. The archived spatialization notebook uses 2017 provincial load profiles as calibration inputs.

The common base demand year is 2023. South African demand uses the spatialized model outputs, and Eskom ESK19622 supplies a national demand and load-reduction reference. Existing South African capacity follows Eskom's 2023 nominal fleet and independent-power-producer totals, with documented spatial allocation and operating assumptions.

\begin{table}[tbp]
\caption{Selected evaluation systems. Counts describe demand aggregation units within each coupled system. All reported trajectories use brownfield mode. The common base demand year is 2023, with different source types retained explicitly.}
\label{tab:regions}
\centering\small
\begin{tabularx}{\textwidth}{@{}lrX@{}}
\toprule
System & Units & Demand basis \\
\midrule
MISO & 6 & EIA-930 hourly observations, grouped LRZs \\
ERCOT & 8 & EIA-930 hourly observations, weather zones \\
NYISO & 11 & EIA-930 hourly observations, zones A--K \\
South Africa & 10 & ESRG reference scenario, corrected supply-area hourly profiles \\
Australia (NEM) & 5 & AEMO five-minute regional \texttt{TOTALDEMAND} from price-and-demand records, averaged to hourly \\
New Zealand & 5 & EMI metered grid offtake by point of connection, summed to the five EMI zones \\
Mexico & 9 & CENACE daily real-demand settlement reports, hourly by control area \\
Brazil & 4 & ONS hourly load curve by subsystem \\
\bottomrule
\end{tabularx}
\smallskip
\par\footnotesize The eight systems contain 58 coupled demand units in total.
\end{table}

Cost inputs retain the NREL Annual Technology Baseline basis for US electricity technologies and PyPSA technology-data for South Africa and the hydrogen chain\cite{nrel2024atb,pypsatechdata}; the Australian, New Zealand, Mexican and Brazilian packs use the ATB basis in US dollars as a declared assumption, and published project capex is converted at 2023 average exchange rates and annualized at a 7.5\% capital recovery factor. The South African cost inputs use a 10\% discount rate.

Input manifests record prepared data, raw extracts and source documents with SHA-256 digests. Baseline resources use one ERA5 point per demand node; NASA-weather input packs use the reconstruction in Section~\ref{sec:future-weather}. Each result records its input, code and dependency identities.

Existing assets are assembled from dated inventory records and mapped to the model's demand zones. The US packs use the 2023 EIA-860 inventory, including battery-energy schedules and the documented geographic and generator-identifier mappings. South African capacity follows Eskom's 31 March 2023 nominal-fleet and independent-power-producer totals, with WRI plant coordinates and historical capacity weights supporting spatial allocation\cite{wrigppd}. Australian registered capacities and maximum storage energies come from AEMO's archived Registration and Exemption List; the December 2023 MMSDM dispatch-unit records identify units registered on 31 December 2023, with identifier-stem matching for batteries whose registration format changed. New Zealand uses the EMI generation inventory with commissioning and decommissioning dates to select units operating during 2023. Mexico uses the state-level technology capacities in PRODESEN 2024--2038, Annex 1, mapped to the model's control regions. Brazil uses ONS unit-level effective capacities and entry/exit dates; the ANEEL SIGA register is retained as a coordinate reference in the source archive. Measured 2023 hydro generation supplies the annual hydro-energy budgets for New Zealand and Brazil. Availability factors, storage-duration fallbacks and regional allocation rules are declared modelling assumptions.

Inter-area transmission is a zonal abstraction constrained by published topology and limits where available. Each region carries a zonal network assembled from public documents (Table~\ref{tab:networks}). Only zone pairs that are electrically adjacent in the source topology are corridors. Existing capacities are taken from published values where they exist, and published limits that apply to several paths at once are kept as interface constraints on the summed flow rather than converted to single-line ratings. MISO publishes zonal import and export limits rather than pairwise ratings, and these enter as zone constraints. Official planned projects are separate build decisions that raise the limits they cross by published increments where available and by declared assumptions otherwise. A generic additional circuit along each existing route is also available as a labelled assumption. Where an operator publishes no pairwise capacity, the corridor carries a non-binding proxy so that the published limits are the binding constraints; the published zonal and interface constraints remain explicit in the model.

\begin{table}[tbp]
\caption{Zonal transmission networks in the evaluation package. Public records supply topology, interface limits and project descriptions. Corridor aggregation, estimated ratings, non-binding proxies and assigned project increments are modelling choices recorded in the input manifests.}
\label{tab:networks}
\centering\small
\begin{tabularx}{\textwidth}{@{}lr>{\raggedright\arraybackslash}X>{\raggedright\arraybackslash}X@{}}
\toprule
System & Corridors & Network basis & Candidate projects \\
\midrule
NYISO & 12 & Directional transfer limits; five model cut-sets assembled from operating and reliability studies & Propel NY (1,300~MW into Zone K) \\
MISO & 7 & Zonal import/export limits; Midwest--South physical transfer limits 3,000/2,500~MW & LRTP Tranche 1 and 2.1 lines, segments of one project bundled \\
ERCOT & 17 & Fixed study limits of 12,240~MW for West Texas Export and 5,595~MW for North-to-Houston & Permian 765~kV import paths and STEP eastern backbone; aggregate STEP increment assigned as described below \\
South Africa & 21 & NTCSA line GIS aggregated to supply areas; ratings estimated using the PyPSA-RSA method\cite{pypsarsa,horsch2017pypsaza} & 31 inter-area candidates extracted from the TDP \\
Australia & 5 & Static directional capabilities from AEMO's archived September 2025 edition (v5) & VNI West, QNI Connect, Marinus Link stages 1/2 and EnergyConnect; ISP Appendix 5 \\
New Zealand & 4 & HVDC inter-island capability from the Transmission Planning Report 2023; island backbones as non-binding proxies & HVDC cable replacement and possible fourth-cable upgrade to 1,400~MW northward \\
Mexico & 10 & PRODESEN Annex 1 circuit/voltage inventory aggregated to nine model control areas; MW ratings estimated by a declared voltage-rating rule & none in the base pack \\
Brazil & 4 & ONS pairwise interchange limits and Northeast/Southeast receiving and exporting interfaces & Gra\c{c}a Aranha--Silv\^ania HVDC candidate, represented by an assumed 4~GW bipole \\
\bottomrule
\end{tabularx}
\end{table}

The ERCOT fixed limits retain specific study assumptions. The 12,240~MW limit is the RTP economic study's 2027 West Texas Export value without outages, and 5,595~MW is the North-to-Houston base value with planned facilities. The source applies an hourly profile to the latter; our model holds the base value fixed. The 3,500~MW improvement reported for the full 765~kV STEP Core Plan, from 12,700 to 16,200~MW, is assigned to the Permian import-path candidate as an aggregate model approximation. Applied to the model's 12,240~MW baseline, that assigned increment gives 15,740~MW before any additional generic-circuit increments.

NYISO's published Propel NY gain is 1,300~MW of additional import capability into Long Island. The model assigns 1,300~MW to both directions of the I--K corridor and to its selected ConEd--LIPA and Dunwoodie/Y49/Y50 interface constraints; reverse interface limits receive the increment where present. These mappings are proxies for the aggregate project benefit. New Zealand's HVDC candidate increases northward capacity from 1,200 to 1,400~MW and assumes a southward increase from 850 to 1,000~MW. The possible fourth-cable enhancement in the TPR supports the northward target; the 150~MW southward increment is a study assumption.

MISO's 3,000/2,500~MW constraints represent the physical regional directional transfer limits, separately from the auction's subregional import/export allowances. South African corridor counts and the 31 TDP candidates are study aggregations of GIS records and planned lines; expected in-service years determine eligibility under the study's stated rule. Australia's archived nominal interconnector capabilities are held fixed, although realized capability depends on operating conditions. New Zealand's 1,400~MW candidate represents the possible fourth-cable enhancement discussed alongside cable replacement, including the need to assess reactive support.

The representative feeders are calibrated so that bare peak load gives approximately 0.96 per unit on the 12.47~kV US feeder and 0.98 per unit on the 22~kV South African feeder. Demand, photovoltaic output, and existing and new battery operation are mapped to those circuits. Line-current limits are 400~A for the normal circuit and 600~A for the reinforced circuit.

\subsection{Geographic representation of committed outcomes}\label{sec:geographic-methods}

The carbon-constrained geographic figures read retained committed candidates without new inference, optimization or power-flow simulation. Each of the ten system--year scenarios displays its minimum observed accepted-cost portfolio across the four deployment--architecture trajectories. GPT-6.1 Sol single-agent supplies the representative in all ten scenarios, with tied minima identified explicitly. The carbon maps show model-zone additions and schematic connections. The main figure displays the latest evaluated year in each of the five systems, while the atlas retains all ten system--year scenarios. Capacity-pie area increases with added generation and battery power, subject to a minimum visible area. For each model zone $z$, annual emissions are $E_z=10^{-6}\sum_{g:\,\mathrm{bus}(g)=z}\epsilon_g\sum_t p_{g,t}\Delta t$, calculated from the retained accepted dispatch and checked by summing the zonal values back to the reported system total. Fixed-size emissions circles share a logarithmic scale from 0.01 to 200~Mt per year; smaller positive values receive the minimum colour and zero values are hollow. This is generation-based direct operational accounting, not consumption-based allocation, lifecycle accounting or measured plant-level emissions. The budget applies to the system total rather than individual zones.

The reference figures~\ref{fig:geographic-planning} and \ref{fig:ercot-verified} read committed candidates without new inference, optimization or power-flow simulation. Figure~\ref{fig:geographic-planning} and Appendix~\ref{app:regional-atlas} select the minimum observed accepted 2035 cost across the four deployment--architecture trajectories per system; tied minima retain the displayed representative. Figure~\ref{fig:ercot-verified} separately follows the Opus multi-agent ERCOT 2035 repair trajectory. New and inherited capacity totals are aggregated from the retained asset records by model node. Added transfer capacity sums selected generic-circuit and official-project increments on each model corridor.

Natural Earth version 5.1.1, at 1:50 million scale, supplies public-domain country and US state boundaries\cite{naturalearth50m}; each panel uses a local Lambert azimuthal equal-area projection. Model-node coordinates come from the prepared regional input packs. They represent grouped-zone centroids, administrative proxies or reference cities, as recorded for each node. In the reference figures, US operating-plant points join the archived 2023 inventory (status OP) to plant coordinates and are deduplicated by plant code. The other five systems use model-zone coordinates. New capacity is allocated by zone, and model connections are drawn as straight schematic corridors.

The reference ERCOT verification panels use annual extrema, currents, ratings and violation counts already committed in the initial and selected Far West representative-feeder outputs. Endpoint colours show annual feeder voltage extrema; line colours show annual per-line peak current divided by the corresponding candidate's rating. Extrema may occur at different hours. The stylized three-bus circuit depicts the numerical feeder topology. A frozen geographic evidence file and SHA-256 source manifest record all extracted values, coordinate files and basemap archives.

\subsection{Brownfield investment and dispatch}

Existing generation and storage retain their locations and capacities while dispatch and additional investment are optimized. Historical construction costs are treated as sunk; specified existing fixed costs and operating costs remain included. The model represents one planning year with exogenous demand. Each demand area has one representative bus with hourly load, photovoltaic and wind profiles.

Inter-area transmission uses a zonal transport formulation. Each corridor is represented by directed links with published or declared capacity and a wheeling cost of 0.1~USD/MWh. Published multi-path limits enter as cut-set constraints, while zonal import and export limits constrain net nodal interchange. Candidate circuits and official projects are binary build decisions. An investment raises the constraints it crosses by a published increment or a declared capacity assumption. Equivalent project choices are ordered to remove symmetric branches without changing the set of feasible networks.

\subsection{Task formulation and acceptance criteria}

Let $w$ denote scenario assumptions, $d$ an agent-selected design and $x$ the capacities and dispatch optimized by the system model. For each candidate, the optimization problem computes
\begin{equation}
x^\star(d;w)\in\arg\min_{x\in\mathcal{X}(d;w)}
\left[\sum_k a_k K_k+\sum_{t,k}m_k p_{k,t}\Delta t
+v\sum_t s_t\Delta t\right],
\label{eq:lp}
\end{equation}
where $a_k$ is the annualized capacity coefficient, $K_k$ the associated capacity, $m_k$ the marginal production cost, $s_t$ unserved power and $v$ the value of lost load. Conversion and storage capacities enter the first term with their corresponding power or energy units. Each of the 8,760 snapshots represents one hour. In the regional package the problem is a mixed-integer programme. Binary variables $y_c$ for candidate circuits and $z_p$ for official projects fix the corresponding link capacities, and each interface $I$ with member corridors $c\in I$, directions $\sigma_c$, base limit $L_I$ and increments $\delta_{p,I}$ satisfies
\begin{equation}
\sum_{c\in I}\sigma_c f_{c,t}\le L_I+\sum_{p}\delta_{p,I}z_p+\sum_{c\in I}r_c y_c\quad\forall t,
\label{eq:interface}
\end{equation}
with $f_{c,t}$ the signed corridor flow and $r_c$ the rating of an assumed expansion. Zonal import and export limits take the same form on net nodal import. The feasible set further includes nodal power balance, technology availability, capacity bounds, cyclic storage dynamics, conversion efficiencies, annual energy budgets for hydro and concentrating solar, and a fixed annual CO$_2$ budget.

For an annual budget $B_{r,y}$ in MtCO$_2$, direct operational emissions satisfy
\begin{equation}
E(d;w)=\frac{1}{10^6}\sum_{g\in\mathcal{G}_{\mathrm{emit}}}\sum_{t=1}^{8760}
\epsilon_g p_{g,t}\Delta t\leq B_{r,y},
\label{eq:operational-carbon-budget}
\end{equation}
where $p_{g,t}$ is generator dispatch in MW, $\epsilon_g$ is the supplied direct emissions factor in tonnes of CO$_2$ per MWh, and $\Delta t=1$~h. The sum covers generators with nonzero inventory emissions factors. Lifecycle emissions and emissions associated with external electricity imports are outside this accounting boundary. The annual budget is immutable through the agent action interface. Short diagnostic horizons receive a proportional budget, but all reported planning outcomes use 8,760 hours. The inventory emissions factors are declared technology-level modelling defaults per MWh of electrical output rather than plant-level operating measurements. For the emitting technologies represented here, they are 0.95, 0.42, 0.75, 0.40, 0.04 and 0.50~tCO$_2$/MWh for coal, gas, oil, waste, geothermal and other generation, respectively; biomass is assigned zero direct emissions under the inventory convention.

The agent evaluates complete designs using
\begin{equation}
J(d;w)=J_{\mathrm{system}}(d;w)+C_{\mathrm{dist}}(d;w).
\label{eq:allin}
\end{equation}
Acceptance requires a valid solver solution, unmet energy below $10^{-4}$ of demand, passing voltage, convergence and thermal checks at every representative feeder, and nodal, storage and interface residuals within the declared tolerances. For the primary cohort, a separate carbon-budget audit checks finite reported emissions, the fixed scenario and result ceilings, source identity and the 8,760-hour horizon. The emissions reporting tolerance is $\max(10^{-9},10^{-8}B_{r,y})$~Mt, and does not relax the optimizer's budget. Gurobi uses eighteen threads, a barrier root relaxation, a requested relative MIP gap of 0.1\% and a 24-hour ceiling per solve\cite{gurobi}. Each numerical cost interval spans the solver bound and incumbent objective for the candidate's fixed optimization domain, with the selected distribution cost added to both endpoints. Unserved energy is penalized at $v=50{,}000\ \USD/\mathrm{MWh}$ and checked against the acceptance threshold after optimization.

Let $\mathcal{A}$ denote committed candidates satisfying all applicable physical and carbon-budget acceptance checks. The selected result is
\begin{equation}
\hat d\in\arg\min_{d\in\mathcal{A}}J(d;w).
\label{eq:selection}
\end{equation}
Equal-cost candidates retain the first recorded accepted result. Hypothetical system-only evaluations are outside $\mathcal{A}$. This rule is applied to the complete committed trajectory for every reported deployment, independently of its final attempted design or stopping decision.

\subsection{Carbon-budget scenarios and source mappings}\label{sec:carbon-scenarios}
Ten researcher-selected system--year scenarios adopt annual emissions ceilings from archived official policy and planning sources (Table~\ref{tab:carbon-scenarios}). We restricted the carbon-constrained evaluation to system--year pairs with a documented mapping from an official electricity-sector emissions budget or pathway to an annual model ceiling, excluding cases such as MISO 2023 and Mexico 2035. These ceilings are implemented through the existing \texttt{operational\_co2\_budget} constraint. The original input packs remain unchanged on disk, and a fixed scenario overlay records the source value, unit conversion, mapping assumptions and content hashes. The baseline policy mode leaves this independent constraint active. Agents can revise design variables but cannot alter the budget, its source identity or the acceptance requirements. The official sources determine the scenario ceilings, whereas achieved emissions are calculated from optimized dispatch and the declared inventory coefficients.

NYISO uses New York's RGGI base allowance budgets as model-wide annual ceilings. US short tons are converted to metric tonnes using the exact multiplier 0.90718474. Regulated-unit coverage, interstate allowance trading, banking, reserve supply and multiyear compliance are not recreated. MISO combines the 2025 Future 2 aggregate planning emissions constraint with the 2026 LTLF Current Trajectory demand inputs. South Africa combines the IRP 2025 balanced-plan emissions pathway with ESRG spatialized reference demand excluding exports. These combinations preserve the published source values without claiming to reproduce each official scenario in full.

Mexico adopts the PROSENER 2030 electricity-emissions projection, combining it with the existing PRODESEN demand inputs. The model excludes SIMUL/Mulege and does not infer a proportional emissions allocation to that excluded subsystem. Brazil uses the PDE 2035 SIN projection, corresponding to the model's four interconnected subsystems, rather than the larger whole-sector projection. The Brazilian plan reports these quantities as projected outcomes rather than imposed emissions limits, and its average-hydrology assumption differs from the model's inherited hydro-energy assumptions. Mexico and Brazil report CO$_2$-equivalent emissions. We adopt their numerical values as direct CO$_2$ ceilings as an explicit proxy, without converting between greenhouse gases. Exact document locations and source-to-model assumptions are preserved in the carbon-scenario archive.

\begin{table}[tbp]
\caption{\textbf{Annual operational CO$_2$ budgets in the primary evaluation.} Each scenario contains both deployments and both architectures. Displayed values are rounded; the archive retains exact source values and conversions. The source categories are retained because an allowance budget, a planning-model constraint and a projected emissions outcome have different institutional meanings. Mexico and Brazil use published CO$_2$-equivalent values as numerical proxies for direct operational CO$_2$.}
\label{tab:carbon-scenarios}
\centering\small
\begin{tabularx}{\textwidth}{@{}lrr>{\raggedright\arraybackslash}X@{}}
\toprule
System & Year & Budget (Mt yr$^{-1}$) & Source basis \\
\midrule
NYISO & 2023 & 24.761865 & New York RGGI base allowance budget \\
NYISO & 2030 & 12.425308 & New York RGGI base allowance budget \\
NYISO & 2035 & 3.550088 & New York RGGI base allowance budget \\
MISO & 2030 & 200.033343 & Future 2 aggregate planning constraint \\
MISO & 2035 & 164.641838 & Future 2 aggregate planning constraint \\
South Africa & 2030 & 168 & IRP 2025 balanced-plan emissions pathway \\
South Africa & 2035 & 142 & IRP 2025 balanced-plan emissions pathway \\
Mexico & 2030 & 119 & PROSENER electricity-emissions projection \\
Brazil & 2030 & 27 & PDE 2035 SIN emissions projection \\
Brazil & 2035 & 29 & PDE 2035 SIN emissions projection \\
\bottomrule
\end{tabularx}
\end{table}

\subsection{Resource and demand construction}

Photovoltaic capacity factors use the PVWatts v8 model through PySAM\cite{pvwattsv8,pysampvwattsv8}. Baseline profiles use ERA5 from the Open-Meteo historical archive, including global, direct-normal and diffuse irradiance, temperature and 100~m wind speed\cite{era5,openmeteo}. Arrays are fixed racks facing the equator, with tilt equal to absolute latitude, a DC-to-AC ratio of one and 14\% losses. Radiation averages are aligned to the load-interval start. The same conversion settings are applied to reconstructed future weather.

Wind capacity factors use the archived 2020 ATB NREL reference turbine curve, rated at 4~MW with a 150~m rotor\cite{nrel_reference4mw150_curve}. For the ERA5-weather evaluation, each node's ERA5 100~m wind-speed series feeds this curve directly. The same reference curve is used across regions, while the node-specific weather produces distinct hourly resource profiles.

US demand uses EIA-930 hourly subregion records aligned to UTC. EIA-930 bulk-system delivery and the South African sent-out demand include distribution losses in their accounting basis. South African demand retains the published regional values and 8,760-hour model-year convention. Each prepared pack aligns load and resource profiles on its declared hourly index.

\subsection{Cost and storage representation}

US electricity-technology costs are derived from the locally archived 2024 ATB version 4, using the Market financing case, Moderate scenario and 2030 technology values with a 30-year cost-recovery period\cite{nrel2024atb}. The battery power and energy coefficients are fitted from the available duration-specific products using the derivation script. South African technology costs and the hydrogen chain use archived PyPSA technology-data inputs\cite{pypsatechdata}; Australia, New Zealand, Mexico and Brazil use the declared ATB electricity-cost basis in US dollars. Annualization assumptions and per-value provenance are retained in the derived cost JSON files.

Battery power $P_b$ and energy $E_b$ are related by the agent-selected duration $h_b$, so $E_b=h_bP_b$ and the annualized investment contribution is
\begin{equation}
C_b=(a_P+h_b a_E)P_b.
\label{eq:battery}
\end{equation}
New batteries use charging and discharging efficiencies of 0.95 each; existing pumped storage and other storage use declared round-trip efficiencies. Storage is cyclic over the simulated horizon. Hydrogen is represented by an electrolyzer link, an energy store on a hydrogen bus and a fuel-cell link with independently extendable capacities. New nuclear is an extendable generator with a minimum output of 70\% of installed capacity; existing nuclear is dispatched flexibly. Existing thermal, hydro and biomass fleets are fixed generators with declared availability, marginal cost and emission rate, and hydro and concentrating solar carry annual energy budgets. Historical construction costs of existing assets are sunk; declared annual fixed costs remain in the objective. The primary cohort uses the scenario-specific annual CO$_2$ budgets in Table~\ref{tab:carbon-scenarios}; the reference archive retains the original 1,000~Mt annual ceiling.

\subsection{Representative-feeder model}

The representative feeder has a source bus, two serial line segments and two loads allocated 40\% and 60\% of feeder demand, with photovoltaic generation at the downstream bus and battery charging and discharging split between the two buses by a design variable. Both loads use a nominal power factor of 0.95. US feeders are 12.47~kV with 2~km sections and South African feeders 22~kV with 3~km sections, at 0.25 and 0.33~ohm/km; these lengths were calibrated so that the bare peak load sits inside the band. A node's demand is mapped to a 5~MW feeder at the node's peak, with photovoltaic and battery dispatch mapped by the same share. Hourly photovoltaic injection and battery power follow the optimized system dispatch.

At each hour, OpenDSS solves a snapshot and returns bus-voltage magnitudes and line currents. The acceptance band is 0.95--1.05 per unit, non-convergent hours count as failures, and line currents are checked against 400~A normal and 600~A reinforced ratings. If reactive control is enabled and the initial maximum exceeds 1.03 per unit, the photovoltaic generator absorbs reactive power at the configured 0.95 power factor and the hour is solved again. The line-upgrade intervention halves line impedance and raises the rating. Its annualized cost is the configured per-feeder cost multiplied by
\begin{equation}
N_{\mathrm{feeders}}=\left\lceil\frac{D_{\mathrm{peak}}\,1000}
{P_{\mathrm{feeder,peak}}}\right\rceil,
\end{equation}
with system demand in MW and feeder peak power in kW. This representative-circuit replication scales intervention costs with each zone's peak demand. Distribution controls are evaluated on the retained system dispatch using OpenDSS steady-state snapshots\cite{epriopendss}.

\subsection{Future-demand sources and scenario construction}\label{sec:demand-scenarios}

\begin{table}[tbp]
\caption{Annual electricity demand in the year-matched inputs. The 2023 demand is paired with ERA5 2023 weather; 2030 and 2035 demand are paired with the corresponding NASA-derived weather. Energy is in TWh; changes are relative to 2023. Both architectures share each input. South African values are spatialized model outputs, Australian growth is mapped between consumption definitions, and New Zealand uses an island winter-growth annualization.}
\label{tab:future-demand-inputs}
\centering\small
\begingroup\setlength{\tabcolsep}{4pt}
\begin{tabular}{@{}lrrrrr@{}}
\toprule
System & Demand 2023 & Demand 2030 & Change (\%) & Demand 2035 & Change (\%)\\
\midrule
MISO & 640.968 & 868.579 & 35.5 & 961.187 & 50.0\\
ERCOT & 446.802 & 983.792 & 120.2 & 1136.972 & 154.5\\
NYISO & 146.997 & 160.160 & 9.0 & 170.170 & 15.8\\
South Africa & 227.005 & 257.571 & 13.5 & 294.539 & 29.7\\
Australia & 180.155 & 203.655 & 13.0 & 252.242 & 40.0\\
New Zealand & 37.630 & 44.919 & 19.4 & 51.399 & 36.6\\
Mexico & 345.292 & 410.615 & 18.9 & 460.419 & 33.3\\
Brazil & 645.656 & 854.450 & 32.3 & 1003.957 & 55.5\\
\bottomrule
\end{tabular}
\endgroup
\end{table}

Demand inputs comprise eight regional packs for each of 2023, 2030 and 2035. Forecast editions published in 2024--2026 determine the target-year scale, and each pack supplies the same demand to both architectures. Table~\ref{tab:future-demand-inputs} reports annual energy and growth relative to 2023. Demand year, resource-weather source and hourly-template year are recorded separately. Reported cases pair 2023 demand with ERA5 2023 resources and future demand with the matching NASA projection year. The ERA5 2023 hourly template supplies intraday shape for future resources, not their projection year.

\paragraph{Hourly demand transformation.}
Let $D_{z,t}^{2023}$ be the prepared demand in MW for zone $z$ and hourly snapshot $t$, and let $E_{z,y}$ be that zone's target annual energy in GWh. For $y\in\{2030,2035\}$, the transformation is
\begin{equation}
E_{z,2023}=\frac{\Delta t}{1000}\sum_{t=1}^{8760}D_{z,t}^{2023},\qquad
g_{z,y}=\frac{E_{z,y}}{E_{z,2023}},\qquad
D_{z,t}^{y}=g_{z,y}D_{z,t}^{2023},
\label{eq:demand-scaling}
\end{equation}
with $\Delta t=1$ hour. Thus $\sum_t D_{z,t}^{y}\Delta t/1000=E_{z,y}$. The zonal multiplier preserves the 2023 load chronology and seasonal shape. Source-specific mappings determine $E_{z,y}$ or, equivalently, $g_{z,y}$.

\paragraph{Published regional annual targets.}
MISO uses the Current Trajectory annual net-load forecast in the 2026 Long-Term Load Forecast regional supplementary workbook. Its ten local resource zones are aggregated into the six model groups. ERCOT uses the 2025 ERCOT-adjusted hourly Long-Term Load Forecast; the eight weather-zone net-load columns are summed over each target year to obtain annual energy. NYISO uses the eleven zonal annual-energy values in Table I-2 of the 2026 Gold Book Baseline Forecast Tables. Mexico uses the planning-scenario net-consumption values in Table 3.5 of SENER's PRODESEN 2024--2038 for the nine modelled control areas. Brazil uses the reference-scenario subsystem loads in Table 2-3 of the 2026 final report of PDE 2035. Each value in average MW is multiplied by 8,760 hours and divided by 1,000 to obtain GWh for the four model subsystems.

\paragraph{Australian consumption-growth mapping.}
Australia uses the Step Change scenario in AEMO's 2026 National Electricity Market Electricity Statement of Opportunities (ESOO) results workbook. Workbook values in TWh are converted to GWh. Let $C_{z,F}^{\mathrm{ESOO}}$ be regional delivered consumption in GWh for financial year $F$. Its growth relative to 2022--23 sets
\begin{equation}
g_{z,y}=\frac{C_{z,F_y}^{\mathrm{ESOO}}}{C_{z,2022\text{--}23}^{\mathrm{ESOO}}},\qquad
E_{z,y}=g_{z,y}E_{z,2023},
\label{eq:australia-demand}
\end{equation}
where $F_{2030}=2029\text{--}30$ and $F_{2035}=2034\text{--}35$. We apply the ESOO delivered-consumption growth factors to the prepared calendar-2023 \texttt{TOTALDEMAND} series as a modelling assumption. This transfers the published annual growth to the baseline series while retaining its chronology and dispatch-demand definition at the regional reference node.

\paragraph{New Zealand winter-growth annualization.}
New Zealand uses Medium Demand in Transpower's 2026 Security of Supply Assessment (SOSA) supplementary data. Let $W_{i,y}^{\mathrm{SOSA}}$ be target-year April--September grid-exit-point offtake in GWh for island $i$. The corresponding model baseline and multiplier are
\begin{equation}
W_{i,2023}=\frac{\Delta t}{1000}\sum_{z\in i}\sum_{t\in\mathrm{Apr\text{--}Sep}}D_{z,t}^{2023},\qquad
g_{i,y}=\frac{W_{i,y}^{\mathrm{SOSA}}}{W_{i,2023}}.
\label{eq:new-zealand-demand}
\end{equation}
All zones on an island receive $D_{z,t}^{y}=g_{i,y}D_{z,t}^{2023}$ for all 8,760 hours. North Island comprises UNI, CNI and LNI; South Island comprises USI and LSI. The island winter-growth factors are thereby annualized while retaining the within-island spatial shares and each zone's hourly pattern.

\paragraph{South African spatialized reference demand.}
South Africa uses the University of Cape Town ESRG Hourly Demand Model v2025 by Province and its spatialized IRP REF series, with exports excluded. The source's hourly model outputs are summed over each target year separately for the ten supply areas to obtain $E_{z,y}$. Equation~\ref{eq:demand-scaling} maps those annual targets onto the corresponding prepared 2023 templates. Both the baseline and future inputs retain the source's status as spatially allocated model outputs informed by national demand projections and regional calibration data.

\paragraph{Fixed assumptions and input checks.}
Demand totals retain the source assumptions about electrification, industrial activity and large loads, including NYISO's large-load tables, AEMO's data-centre outlook and Transpower's DataGrid step-load assumptions. The 2023 evaluation uses ERA5 2023 resource profiles; the future packs use the corresponding target-year NASA profiles described below. All cases retain the inherited fleet, baseline policy and ATB/PyPSA technology-cost basis\cite{nrel2024atb,pypsatechdata}. The primary evaluation applies its fixed carbon-budget overlay independently of this policy mode. Each target year is a single-period brownfield problem, and representative-feeder replication follows the scenario peak.

South African 2030 cases apply commissioning-year eligibility to the official transmission projects in the NTCSA Transmission Development Plan. A project is eligible when its documented in-service year is at or before 2030; an indivisible project with a date range uses the final year. Other region--year routes retain their inherited project permissions. Generic candidate-circuit expansion follows the declared circuit bounds.

Preparation records source editions, scenarios, extraction coordinates, geographic mappings, unit conversions and file hashes. Demand checks confirm 8,760 finite non-negative values per zone, agreement with annual-energy targets and identical inputs across architectures.

\subsection{Future-weather inputs and hourly reconstruction}\label{sec:future-weather}

For the combined input library, we prepared 16 future-weather input packs covering eight systems in 2030 and 2035, comprising 116 annual node profiles. The primary cohort uses nine of these future system--year packs. NASA NEX-GDDP-CMIP6 v2.0 supplies daily, bias-corrected projections at $0.25^\circ$ resolution\cite{nasa2022nexcmip6,nasa2025nexv2}. The selected realization is MPI-ESM1-2-HR, member r1i1p1f1, under SSP2-4.5. At each of the 58 representative locations, we selected the nearest grid cell with complete records for all four variables, comprising minimum and maximum temperature (\texttt{tasmin}, \texttt{tasmax}; K), daily-mean shortwave irradiance (\texttt{rsds}; W~m$^{-2}$) and near-surface wind speed (\texttt{sfcWind}; m~s$^{-1}$).

Target-year daily values are combined with the corresponding month--day ERA5 2023 hourly patterns\cite{era5,openmeteo}. Each retrieval includes the preceding 31 December and following 1 January; their templates are 2022-12-31 and 2024-01-01. Both projection years have 365 days. Complete UTC days are reconstructed before selecting 8,760 hours using each node's fixed model-clock offset. Solar geometry uses target-year dates and interval midpoints. The profiles retain the 2023 hourly index and inherited target-year demand values (Table~\ref{tab:future-weather-inputs}).

Let $N$ denote NASA daily values, $E$ the ERA5 template, $d$ a UTC day and $h$ an hour. After converting temperature to $^\circ$C, hourly temperature and wind are
\begin{align}
T^{y}_{d,h}&=T^{N,\min}_{d}
+\bigl(T^{N,\max}_{d}-T^{N,\min}_{d}\bigr)
\frac{T^{E}_{d,h}-\min_j T^{E}_{d,j}}
{\max_j T^{E}_{d,j}-\min_j T^{E}_{d,j}},\label{eq:future-temperature}\\
u^{y}_{\ell,d,h}&=u^{E}_{\ell,d,h}
\frac{\overline{u}^{N}_{d}}{\overline{u}^{E}_{10,d}},
\qquad \ell\in\{10,100\}\ {\rm m},\label{eq:future-wind}
\end{align}
where overbars denote daily means. The common wind multiplier retains the template's hourly variation and vertical wind-speed ratio.

Daily solar energy is allocated proportionally to the ERA5 shortwave pattern, aligned from preceding-hour averages to interval starts. Allocation is bounded by extraterrestrial horizontal irradiance, with remaining energy redistributed across unsaturated daylight hours to satisfy
\begin{equation}
\sum_{h=1}^{24}G^{y}_{d,h}\Delta t
=24\,{\rm h}\,\overline{R}^{N}_{d},\qquad
0\leq G^{y}_{d,h}\leq I_{0,d,h}\max(\cos\theta_{d,h},0),
\label{eq:future-radiation}
\end{equation}
where $\Delta t=1$~h, $G$ is global-horizontal irradiance, $\overline R^N$ is \texttt{rsds}, $I_0$ is extraterrestrial normal irradiance and $\theta$ is midpoint solar zenith. The pvlib Erbs decomposition supplies direct-normal irradiance, capped at $I_0$; diffuse-horizontal irradiance is recomputed as $G-\mathrm{DNI}\max(\cos\theta,0)$.

We apply PVWatts v8 through PySAM (NREL-PySAM 7.1.1.post1 in the archived generation environment)\cite{pvwattsv8,pysampvwattsv8} with a 1~MW DC fixed array, DC/AC ratio one, equator-facing orientation, tilt equal to absolute latitude and 14\% losses as study settings. Synthetic 100~m wind supplies its wind-speed input. Photovoltaic capacity factors are AC output divided by 1~MW DC. Wind capacity factors are the interpolated NREL 4~MW, 150~m-rotor reference-curve output at synthetic 100~m wind, divided by 4~MW\cite{nrel_reference4mw150_curve}. Checks verify daily temperature extrema, solar energy and near-surface mean wind, complete hourly profiles and preserved demand. ERA5 and NEX-GDDP-CMIP6 differ in their wind and irradiance climatology at the study locations, so mean capacity factors differ between the 2023 and future-year packs (Table~\ref{tab:resource-cf}). Differences between 2023 and future-year outcomes therefore combine demand growth, projected climate and the change of data product. Source, processing-code and output SHA-256 hashes accompany each pack.

\begin{table}[tbp]
\caption{Mean resource capacity factors in the reported input packs. Values are unweighted means of annual node capacity factors in each system. The 2023 packs use ERA5; the 2030 and 2035 packs use NASA NEX-GDDP-CMIP6 fields reconstructed as described above.}
\label{tab:resource-cf}
\centering\small
\begin{tabular}{@{}lrrrrrr@{}}
\toprule
 & \multicolumn{3}{c}{Wind} & \multicolumn{3}{c}{Solar PV} \\
\cmidrule(lr){2-4}\cmidrule(lr){5-7}
System & 2023 & 2030 & 2035 & 2023 & 2030 & 2035 \\
\midrule
MISO & 0.321 & 0.255 & 0.271 & 0.168 & 0.202 & 0.204 \\
ERCOT & 0.339 & 0.256 & 0.245 & 0.179 & 0.211 & 0.210 \\
NYISO & 0.230 & 0.422 & 0.441 & 0.149 & 0.184 & 0.189 \\
South Africa & 0.252 & 0.228 & 0.218 & 0.201 & 0.205 & 0.206 \\
Australia (NEM) & 0.249 & 0.439 & 0.449 & 0.170 & 0.180 & 0.180 \\
New Zealand & 0.398 & 0.596 & 0.594 & 0.159 & 0.176 & 0.178 \\
Mexico & 0.075 & 0.198 & 0.195 & 0.203 & 0.212 & 0.211 \\
Brazil & 0.140 & 0.168 & 0.195 & 0.163 & 0.170 & 0.167 \\
\bottomrule
\end{tabular}
\end{table}

\begin{table}[tbp]
\caption{Weather inputs and time coordinates. Future resource profiles combine the indicated NASA daily projection with ERA5 2023 intraday patterns. Each row covers eight systems and 58 nodes; every profile contains 8,760 hours. The two future rows comprise 16 prepared input packs and 116 annual profiles.}
\label{tab:future-weather-inputs}
\centering\small
\begin{tabular}{@{}l >{\raggedright\arraybackslash}p{0.37\linewidth} c c@{}}
\toprule
Demand year & Weather source & Hourly template & Hourly index\\
\midrule
2023 & ERA5 2023 reanalysis & 2023 & 2023\\
2030 & NASA MPI-ESM1-2-HR SSP2-4.5, 2030 projection & ERA5 2023 & 2023\\
2035 & NASA MPI-ESM1-2-HR SSP2-4.5, 2035 projection & ERA5 2023 & 2023\\
\bottomrule
\end{tabular}
\end{table}

\subsection{Agent architectures, models and evidence retention}

The shared action interface includes technology permissions, global and node-level capacity caps, battery duration, charging and discharging fractions, distribution reinforcement, reactive control, battery placement, candidate-circuit bounds and project permissions. These variables constrain the optimizer. Demand, weather, costs, existing assets, network assumptions, policy and the annual carbon budget are immutable through the agent interface. Code reconstructs fixed roles, tools and acceptance requirements independently of editable guidance.

The single agent maintains a continuous conversation and free-form strategy notes. At each turn, it chooses among state inspection, design modification, system simulation, feeder rechecking, note revision and termination. The agent determines the order of these actions, subject to fixed tool preconditions. The current notes are reinserted into the system prompt at every turn, while the preceding conversation remains available.

At the start of each multi-agent cycle, four specialists in system architecture, renewable generation, storage and firm generation receive the same shared state and advisory procedure-graph guidance. They independently propose changes without access to one another's proposals for that cycle. The orchestrator receives all four proposals and selects one experiment, which may combine proposed changes or retain the current design. A valid selection is evaluated by PyPSA, followed by OpenDSS if the system solve succeeds. The evaluator interprets the cycle record, and the critic reviews that record together with the evaluation. The procedure refiner then updates the advisory graph, the compressor records a structured summary, and the convergence judge decides whether to continue or stop. Roles use separate model-call contexts and exchange information through explicitly supplied records. Across cycles, the framework retains the current design, latest system and feeder evidence, accumulated cycle summaries and advisory graph. The judge additionally receives a code-computed outcome trajectory.

The procedure graph contains planning steps with directed edges, conditions, guidance and pitfalls, inspired by procedural memory\cite{lu2026procedural}. It begins with a six-node expert skeleton and is limited to thirty nodes and sixty edges with bounded text fields. Structurally valid edits are applied and logged. Their contents provide advice without altering tools or numerical requirements.

Every successful committed system evaluation triggers an OpenDSS check. A distribution-only edit invalidates feeder evidence while preserving a matching system dispatch, whereas a system-design edit invalidates both. The convergence judge can request up to two system-only verifications of hypothetical alternatives within a judge session. Their outputs are logged and used for that session's decision to continue or stop. They are not included in subsequent-cycle prompts and do not establish physical acceptance. The single agent requests termination through \texttt{finish\_design}, and the multi-agent judge issues its verdict through \texttt{judge\_convergence}. These stopping decisions are subject to current-evidence requirements, and the configured search ceilings limit further execution. Stopping does not itself establish physical acceptance. The selected accepted candidate and the last attempt are recorded separately.

Provider adapters translate a common action vocabulary to OpenAI Responses and Anthropic Messages. The reported deployments are GPT-6.1 Sol (\texttt{gpt-6.1-sol}) and Claude Opus 5.5 (\texttt{claude-opus-5-5})\cite{anthropic2026opus55}. Within each multi-agent run, all roles use the same model deployment, with specialization provided by role-specific instructions and inputs. Hosted models use their provider's \texttt{high} effort setting, with the recorded inference configurations in Appendix~\ref{app:deployments}. We use the pretrained deployments without updating model weights or constructing a task-specific training set. Adaptation is confined to the recorded conversation, strategy notes, advisory procedure graph and proposed actions.

The primary and completed hosted-weather campaigns permit 120 single-agent turns, forty committed simulation requests and 120 multi-agent cycles for both deployments. The default response ceiling is 64,000 tokens, with smaller role-specific ceilings retained. Each multi-agent cycle contains several role-specific responses. The earlier base-year trajectories retain their recorded execution settings in Appendix~\ref{app:deployments}.

The runtimes record exact system-result reuse separately from numerical optimization. Eligible reuse requires matching system design, scenario, solver identity and solver settings, and preserves the archived optimal dispatch and objective evidence. Distribution-only controls can be changed while reusing a matching system dispatch, after which the requested feeder evaluation is performed. During the primary campaign, a separately recorded execution revision extended this matching to system-equivalent designs and retained the latest committed dispatch independently of the recent trial cache. The source and deployment manifests identify the affected trajectories. Logical evaluation requests, actual numerical solves and cache hits are distinct measures of effort. Reuse does not change the numerical acceptance requirements, but runtime and trajectory comparisons retain the execution-revision distinction. This revision applies from the interrupted NYISO 2035 GPT-6.1 Sol multi-agent verification onward, covering that trajectory and the seven subsequent GPT-6.1 Sol multi-agent trajectories.

\subsection{Outcome extraction and comparison}

The primary reporting manifest selects forty completed carbon-constrained trajectories. A separate reference manifest retains the ninety-six trajectories from the original loose-cap evaluation. For each carbon trajectory, extraction reads its retained committed design, system result, distribution evidence, carbon-budget audit and numerical bound. The reported outcome is the lowest-cost accepted committed candidate selected by the retention rule in Eq.~\ref{eq:selection}. Physical and emissions checks are distinguished from the agent's stopping decision. Source manifests preserve the result paths, scenario and implementation identities, and links to numerical and model-response records.

Search histories, received-response tokens and process durations are reconstructed separately from the annual power-system objective (Section~\ref{sec:search-methods}).

Reference comparisons match system, demand year, resource profiles, deployment and architecture. Input-file hashes match for all forty pairs, with the source-derived carbon ceiling applied through the recorded scenario overlay. The reported contrasts describe observed trajectories, and any differences in provider configuration, recovery history and result-reuse revision remain attached to the pair. Numerical bounds are conditional on the optimization domain of each selected design. The comparison contains one trajectory per condition and does not establish a distribution of performance across repeated searches.

\subsection{Reconstruction of search effort and token usage}\label{sec:search-methods}

We analyse the forty primary trajectories and the ninety-six loose-cap reference trajectories separately. A committed assessment is the chronological evaluation of a candidate after a system-simulation request or an additional feeder-only check; repeated designs and rechecks remain separate assessments. For the reference cohort, candidate reconstruction reproduces the hashes of all 382 assessed designs and links each archived selected design and objective to its first recorded assessment. The primary counts use the completed response ledgers and recorded committed and hypothetical system-evaluation requests. Native single-agent turns and multi-agent cycles are retained in the evidence; common comparisons use assessment order. A rejected improvement is an assessment whose recorded all-in objective is lower than an already accepted incumbent beyond the declared numerical comparison tolerance ($10^{-6}$ in the recorded billion-per-year objective units), but which fails the acceptance gate. Such an objective does not represent a feasible saving. Uncommitted system-only probes are tabulated separately because they lack feeder validation.

We deduplicate received responses by provider and response identifier and reconcile 764 received responses in the primary cohort and 2,257 in the reference cohort with their archived ledgers and journals. Input tokens are normalized as uncached input plus cache reads plus cache writes; repeated context is counted each time it is supplied. Output tokens include the reported reasoning subset, so reasoning is not added again to the total. Tokens describe the complete selected search histories, including inherited responses exactly once within each retained trajectory. Responses from separate discarded attempts are excluded from these trajectory totals. Provider tokenizers and usage-reporting conventions differ.

Matched comparisons pair single- and multi-agent trajectories within deployment, system and year. Aggregate ratios divide sums across a cohort, whereas median paired ratios summarize within-scenario quotients. Reference-cohort process-duration summaries use 92 trajectories and 44 pairs; four resumed reference histories lack full uninterrupted duration records. The primary campaign is not treated as a controlled wall-runtime benchmark because pauses, recovery and the result-reuse revision affect execution histories. Process duration covers the recorded workflow, and logged solver time is reported separately. Evaluation requests count logical calls, including exact-result reuse. All summaries are descriptive, with one trajectory per deployment--architecture--scenario cell.

\subsection{Comparison with published expert plans}\label{sec:expert-methods}

We compare the retained carbon-constrained designs for Brazil and South Africa in 2035 with archived institutional planning outputs, without new inference or simulation. These two systems have explicit archived 2035 capacity mappings for the categories in Table~\ref{tab:expert-capacity}. All four accepted deployment--architecture trajectories per system are included. Capacity values are read from each retained candidate's total-capacity record and checked against the final reported output. Minima and maxima span the four observed outcomes rather than statistical uncertainty. For capacity category $k$, the descriptive difference is $\Delta C_k=C^{\mathrm{agent}}_k-C^{\mathrm{plan}}_k$ in GW.

Brazil uses the PDE 2035 chapter-three workbook, \texttt{Figura 3-25}, December 2035 row 8, with wind from E8, centralized solar from F8 and batteries from I8. MMGD capacity G8 is kept separate. South Africa uses the IRP 2025 user-friendly edition, Table 1, printed page 31 (PDF page 41). The wind and Solar PV columns are reconstructed as the end-2025 current base plus scheduled 2026--2035 additions. Separate distributed-generation, CSP and hybrid columns are excluded. The model inherits its 2023 fleet, whereas the South African plan's capacity base includes assets available by end-2025.

Each comparison retains the published capacity category. Model solar is a zonal allocation without a separate future distributed-generation rollout, so the solar differences cannot be interpreted as national photovoltaic surpluses. Demand, retirements, resources, costs, permitted technologies and delivery schedules retain the respective study assumptions. PDE 2035 supplies Brazilian demand and the numerical emissions ceiling adopted here; IRP 2025 supplies the South African emissions pathway, while South African demand uses ESRG. The comparison thus shares planning inputs with its institutional references and is not independent validation against withheld targets. The carbon-budget mappings, including Brazil's CO$_2$-equivalent proxy, are specified in Section~\ref{sec:carbon-scenarios}. The comparison archive retains all eight candidate identities, capacities, acceptance records, source hashes, workbook cells or page locators and calculations.

\backmatter
\bmhead{Data availability}
Demand and fleet sources include EIA, ESRG, Eskom, AEMO, the New Zealand Electricity Authority, Transpower, CENACE, SENER, ONS and ANEEL. ERA5 provides the baseline weather, and NASA NEX-GDDP-CMIP6 v2.0 provides the future daily climate fields\cite{era5,openmeteo,nasa2025nexv2,nasa2022nexcmip6}. The country-indexed source catalogue in Appendix~\ref{app:regional-sources} provides direct document links. The study archive retains source snapshots, geographic mappings, forecast targets, hourly resource profiles and input hashes. The manuscript evidence links each table and figure to its selected planning outcomes and comparison records. The carbon-constrained expert-plan comparison retains the eight Brazilian and South African 2035 candidates, document editions, workbook cells, category mappings and source hashes. The search-and-token archive contains the 96 reference assessment histories and the separately identified carbon-cohort response and evaluation records, with normalized response counts, token totals, role attribution, matched comparisons and source hashes. Carbon scenario records preserve the source-derived budgets, conversion assumptions and acceptance audits.

\bmhead{Code availability}
The planning and analysis code is publicly available at \url{https://github.com/sixticket/grid-agent}. The repository contains the system simulator, regional scenario preparation, NASA-to-hourly reconstruction, model adapters, launchers and dependency records. The manuscript package includes the 96-trajectory reporting manifest drawn from the frozen Opus base-year and hosted-weather evidence, read-only committed-history extractors, table and figure generators, the frozen-evidence search-and-token table builder, and processing provenance for the 16 future-weather packs. A separate forty-trajectory manifest, fixed carbon-budget overlay, source catalogue, execution-revision manifest and carbon figure generators accompany the primary evaluation.

\bmhead{Acknowledgements}
This work was supported by the U.S. National Science Foundation (NSF) through CAREER Award No.~ECCS-2543177, by the U.S. Department of Energy through Distinguished Early Career Award No.~DE-NE0009635, and by the U.S. Nuclear Regulatory Commission under Grant No.~31310024M0041.

Computing resources on Delta and DeltaAI at the National Center for Supercomputing Applications at the University of Illinois Urbana-Champaign were provided through two ACCESS allocations: CIS262451 (\emph{AI Agents for Safety-Critical Energy Systems}) and CIS261154 (\emph{Deployable Energy-Efficient Agentic Foundation Model with Edge AI for Energy Systems}). The State of Illinois and NSF support Delta (OAC-2005572) and DeltaAI (OAC-2320345). NSF supports the ACCESS program\cite{boerner2023access} through grants 2138259, 2138286, 2138307, 2137603 and 2138296.

We thank the organisations whose public data made the regional package possible. The U.S. Energy Information Administration provided hourly demand (EIA-930) and the generator inventory (EIA-860). The New York Independent System Operator, the Midcontinent Independent System Operator and the Electric Reliability Council of Texas published the reliability, operating-study, resource-adequacy and transmission-planning documents from which zonal constraints and planned projects were taken, and the Public Utility Commission of Texas the 765~kV import-path approvals. The National Transmission Company South Africa published the Transmission Development Plan 2024 and Eskom the Integrated Report 2023 and station data. The Energy Systems Research Group at the University of Cape Town (G.~Ireland and colleagues) published the spatialized South African demand model, and Meridian Economics and the PyPSA-ZA authors (J.~H\"orsch and J.~Calitz) the open PyPSA-RSA model whose topology method we replicated. The Australian Energy Market Operator published the NEMWEB demand data, the registration lists, the interconnector capabilities and the Integrated System Plan; the New Zealand Electricity Authority the EMI datasets and Transpower the Transmission Planning Report; Mexico's Centro Nacional de Control de Energ\'ia the real-demand reports and the Secretar\'ia de Energ\'ia the PRODESEN; Brazil's Operador Nacional do Sistema El\'etrico the open data portal and the PEN, and ANEEL the SIGA plant register. The World Resources Institute provided the Global Power Plant Database and geoBoundaries the administrative boundaries. ERA5 reanalysis was produced by ECMWF under the Copernicus Climate Change Service and served through the Open-Meteo archive. NASA provided the NEX-GDDP-CMIP6 projections, including the MPI-ESM1-2-HR simulation used here. The National Laboratory of the Rockies (formerly NREL) provided the Annual Technology Baseline, the reference turbine power curves and the System Advisor Model through PySAM, and the PyPSA project the technology-data repository. We used PyPSA, linopy, OpenDSS through OpenDSSDirect.py, HiGHS and Gurobi under an academic licence.

\bmhead{Author contributions}
Y.P.L. conceived the study, developed the framework, prepared the regional data and performed the experiments. S.C. and S.B.A. supervised the work. All authors wrote and revised the manuscript.

\bmhead{Competing interests}
The authors declare no competing interests.

\begin{appendices}
\renewcommand{\theHtable}{appendix.\Alph{section}.\arabic{table}}
\renewcommand{\theHfigure}{appendix.\Alph{section}.\arabic{figure}}
\section{Carbon-constrained outcomes and matched reference evidence}\label{app:carbon-evidence}
\setcounter{figure}{0}

The following figures retain the forty matched cost and emissions comparisons and candidate-specific numerical bounds. Main-text Fig.~\ref{fig:carbon-geography} combines grid additions and emissions for the latest evaluated year in each system. The full-precision forty-row ledger and ten-case grid-and-emissions atlas are provided in the evidence archive. The original 1,000~Mt reference ceiling is nonbinding in every matched reference outcome. Configuration differences are retained with the pairs, so these figures describe matched retrospective comparisons rather than a controlled carbon-only ablation.

\begin{figure}[tbp]
\centering
\makebox[\textwidth][c]{\includegraphics[width=170mm]{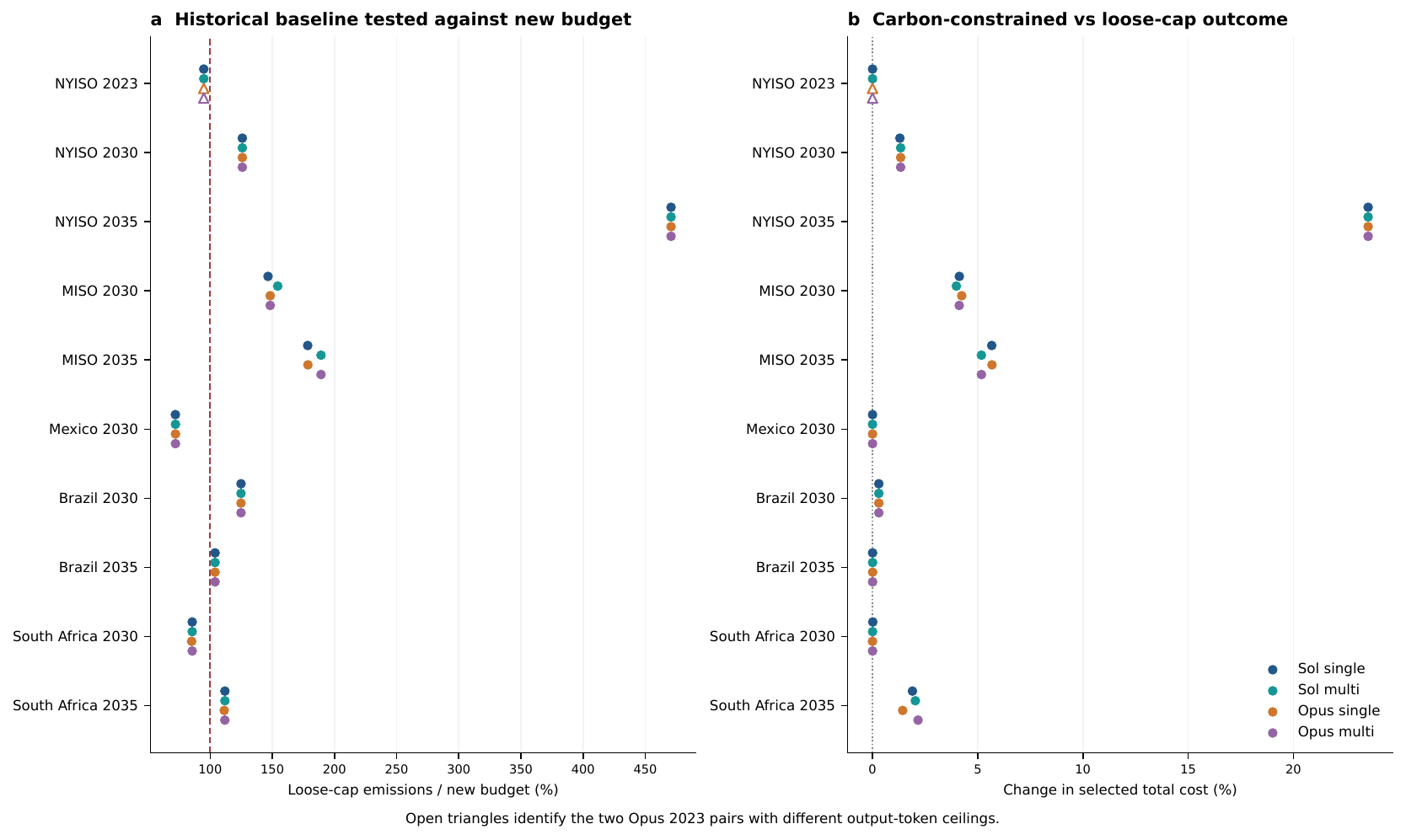}}
\caption{\textbf{Carbon-constrained outcomes and matched loose-cap references.} Comparisons pair each primary trajectory with the archived trajectory having the same system, year, deployment and architecture under the original 1,000~Mt annual ceiling. Twenty-eight reference designs exceed the source-derived ceilings. Costs include annualized representative-feeder interventions, and emissions are direct operational CO$_2$ within the model boundary. Open triangles identify the two 2023 Opus pairs with distinct output ceilings and provider configurations. The points describe observed trajectories and are not repeated samples.}
\label{fig:carbon-reference}
\end{figure}

\begin{figure}[tbp]
\centering
\makebox[\textwidth][c]{\includegraphics[width=170mm]{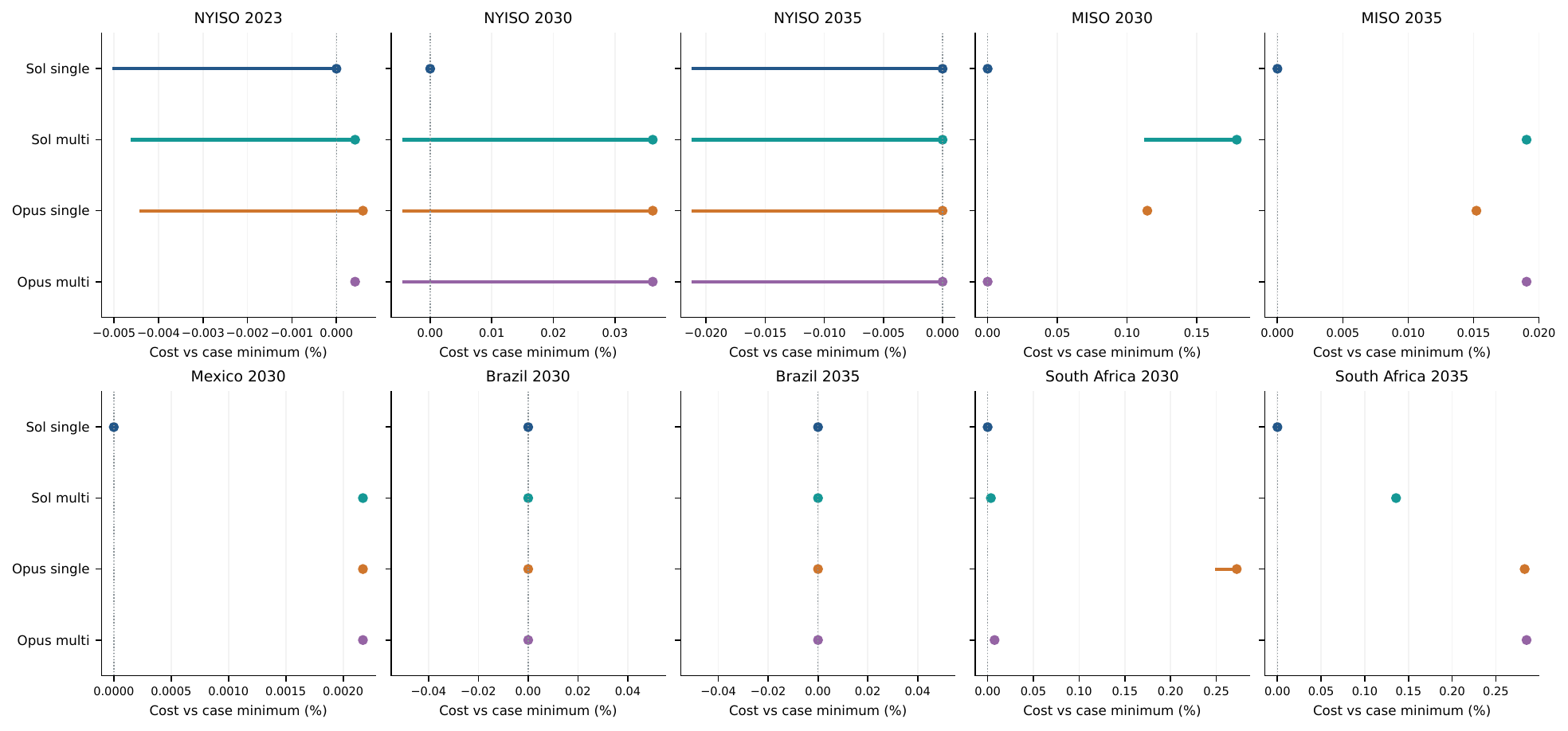}}
\caption{\textbf{Candidate-specific numerical bounds for the selected carbon-constrained outcomes.} Each interval spans the numerical solver lower bound and achieved system objective for the selected candidate's fixed optimization domain, adding the selected distribution cost at both endpoints. These intervals describe numerical uncertainty within each candidate-specific problem. They are not confidence intervals or bounds on the full agent search space. All forty selected outcomes pass the physical and carbon-budget checks.}
\label{fig:carbon-bounds}
\end{figure}

\clearpage
\section{Execution configurations}\label{app:deployments}
\setcounter{figure}{0}
The primary evaluation contains forty carbon-constrained trajectories. Both deployments use a 64,000-token default output ceiling, high effort, 120 single-agent turns, forty committed single-agent simulation requests and 120 multi-agent cycles, with the recorded role-specific response ceilings. The scenario manifests retain the fixed budgets and the completed trajectory identities. Pause, recovery and system-result reuse records remain attached to the affected trajectories. A separate reference archive contains the ninety-six GPT-6.1 Sol and Claude Opus 5.5 trajectories described below. The sixteen Opus trajectories with 2023 demand and ERA5 2023 weather come from the earlier base-year campaign (frozen on 29 September 2026 at 23:35:46 UTC). The remaining eighty trajectories retain the settings of the later hosted campaign. Table~\ref{tab:model-settings} records the earlier Opus configuration.

\begin{table}[htbp]
\caption{Execution settings for the sixteen reported Opus baseline trajectories. Evaluation counts include requests served by an exact matching cached result.}
\label{tab:model-settings}
\centering\small
\begin{tabularx}{\textwidth}{@{}l>{\raggedright\arraybackslash}X@{}}
\toprule
Setting & Claude Opus 5.5\\
\midrule
Inference & Anthropic Messages API\\
Completed trajectories & 16 for demand year 2023\\
Single-agent ceiling & 120 turns; 40 committed evaluations\\
Multi-agent ceiling & 120 cycles\\
Default output ceiling & 16,000 tokens\\
Generation & Adaptive thinking; high effort; standard-only tier\\
Numerical solver & Gurobi 13.0.3; 18 threads; 0.1\% requested gap\\
Planning horizon & 8,760 hourly snapshots\\
Selection & Lowest-cost physically accepted committed candidate\\
\bottomrule
\end{tabularx}
\end{table}

GPT-6.1 Sol links successive Responses API requests through \texttt{previous\_response\_id}. Opus uses adaptive thinking, high effort and a five-minute prompt cache. Its revised adapter uses the \texttt{thinking-binding-controls-2026-08-01} header and \texttt{prefix\_mismatch\_behavior=drop\_block} so that strategy-note edits can update the system prompt under the preserved-thinking protocol\cite{anthropic2026preservedthinking}. Original response blocks remain in saved history; the server records exclusions of prefix-incompatible thinking blocks in \texttt{input\_transformations}. Trajectory records identify the applied adapter, retained response history and provider-reported thinking transformations.

\paragraph{Completed hosted NASA-weather campaign.}
A separate completed campaign reports eighty hosted trajectories, with its final case completed on 3 October 2026 at 06:08:08 UTC (Table~\ref{tab:nasa-model-settings}). It includes sixty-four cases with the prepared MPI-ESM1-2-HR SSP2-4.5 weather inputs for 2030 and 2035, and sixteen GPT-6.1 Sol controls with 2023 demand and ERA5 2023 weather. Both deployments use a 64,000-token default output ceiling, with 12,000-token critic requests; the single-agent ceiling is 120 turns and forty committed evaluations, and the multi-agent ceiling is 120 cycles. They share the agent workflow, action interface and physical acceptance rules, with the 8,760-hour horizon and numerical solver settings in Table~\ref{tab:model-settings}. GPT-6.1 Sol uses high effort and the standard tier; Opus uses the revised thinking protocol above with high effort and the standard-only tier. Original execution identities and complete response ledgers remain attached to each trajectory. Three resumed trajectories have no uninterrupted wall-runtime estimate; their retained numerical evidence and unique received responses are counted once.

\begin{table}[htbp]
\caption{Completed cases in the new hosted weather campaign. Each listed demand year contains eight systems and two agent architectures. The sixteen earlier Opus base-year trajectories are reported separately.}
\label{tab:nasa-model-settings}
\centering\small
\begin{tabularx}{\textwidth}{@{}l>{\raggedright\arraybackslash}Xr@{}}
\toprule
Model & Demand and weather inputs & Completed cases\\
\midrule
GPT-6.1 Sol & 2023 demand with ERA5 2023 weather; 2030 and 2035 demand with the corresponding NASA-derived weather & 48\\
Claude Opus 5.5 & 2030 and 2035 demand with the corresponding NASA-derived weather & 32\\
\bottomrule
\end{tabularx}
\end{table}
\section{Planning outcomes under the loose-cap reference}\label{app:loose-cap-results}
\setcounter{figure}{0}

This section preserves the original ninety-six trajectories under a common annual CO$_2$ ceiling of 1,000~Mt. Forty trajectories match the primary cohort by system, year, deployment and architecture; the remaining fifty-six extend regional or temporal coverage. All numerical results, maps and search histories in this section belong to the reference evaluation.

\subsection{Initial designs and retained reference candidates}

In the loose-cap reference evaluation, the engine's default design permits all extendable technologies, sets capacity bounds to their maxima and uses four-hour batteries without feeder interventions. This default is the least-cost plan that the optimizer returns without agent restrictions under the original carbon ceiling. It passes the gate in 17 of the 24 system--year problems and fails only the distribution component in seven. The 28 trajectories for those seven problems comprise twelve South African, eight ERCOT, four Mexican and four NYISO trajectories; all subsequently retain an accepted candidate. In 46 of the 56 trajectories that retain the default design, agents test 99 alternatives before stopping, comprising 60 hypothetical system-only checks and 39 committed designs; none is cheaper. The remaining 12 trajectories in the passing problems select modified designs that are up to 0.483\% cheaper. Twelve initial rejections occur in the 2023 cohort and eight in each future-year cohort. The records show transitions from rejected to accepted planning candidates in multiple systems.

\subsection{Reference experiment coverage and retained planning outcomes}\label{sec:reference-complete-results}
The three-year GPT-6.1 Sol and Claude Opus 5.5 coverage contains 96 trajectories across eight systems, three matching demand/weather years, two deployments and single- and multi-agent architectures. All 96 trajectories are reported in Table~\ref{tab:complete-coverage}. The Opus 2023 results come from the earlier base-year campaign; deployment versions and recorded inference settings remain distinct.

Every reported trajectory retains a physically accepted committed candidate. Table~\ref{tab:all-selected-costs} reports its selected total annual cost for every system, deployment, year and architecture. These outcomes demonstrate operation of the framework across the full evaluated scope. Detailed costs, candidate-specific solver bounds and effort records remain in Appendix~\ref{app:outcomes}.

The geographic portfolios in Fig.~\ref{fig:geographic-planning} show the lowest-cost accepted 2035 design observed for each system across both deployments and architectures, with tied costs explicitly identified. The Opus ERCOT repair in Fig.~\ref{fig:ercot-verified} is a separate process example. These spatial outputs complement the complete results in Tables~\ref{tab:complete-coverage} and~\ref{tab:all-selected-costs}.

\begin{table}[htp]
\caption{\textbf{Loose-cap reference coverage and physical acceptance.} Each cohort contains eight systems and both architectures. The complete GPT-6.1 Sol--Opus grid contains 96 trajectories. Accepted counts refer to trajectories with a retained committed candidate passing the implemented system and representative-feeder gate. Baseline resources are ERA5 2023; future resources are reconstructed for the indicated NASA projection year.}
\label{tab:complete-coverage}
\centering\small
\begin{tabular}{@{}llrrrr@{}}
\toprule
Deployment & Demand / weather & Single & Multi & Accepted & Total \\
\midrule
GPT-6.1 Sol & 2023 / ERA5 & 8 & 8 & 16 & 16 \\
GPT-6.1 Sol & 2030 / NASA & 8 & 8 & 16 & 16 \\
GPT-6.1 Sol & 2035 / NASA & 8 & 8 & 16 & 16 \\
Opus 5.5 & 2023 / ERA5 & 8 & 8 & 16 & 16 \\
Opus 5.5 & 2030 / NASA & 8 & 8 & 16 & 16 \\
Opus 5.5 & 2035 / NASA & 8 & 8 & 16 & 16 \\
\midrule
Total & Year matched & 48 & 48 & 96 & 96 \\
\bottomrule
\end{tabular}
\end{table}
\begin{tableorg}[p]
\caption{\textbf{Selected planning outcomes for all 96 loose-cap reference trajectories.} Entries are total annual power-system costs in billion USD per year, including annualized representative-feeder interventions, rounded to three decimals. S and M denote single- and multi-agent architectures. The Accepted column counts retained physically accepted trajectories within each system--deployment row. Full numerical precision, candidate-specific solver bounds and evaluation effort are retained in Appendix~\ref{app:outcomes}.}
\label{tab:all-selected-costs}
\centering\small
\makebox[\textwidth][c]{\begin{tabular*}{170mm}{@{\extracolsep{\fill}}llrrrrrrr@{}}
\toprule
System & Deployment & \multicolumn{2}{c}{2023} & \multicolumn{2}{c}{2030} & \multicolumn{2}{c}{2035} & Accepted \\
\cmidrule(lr){3-4}\cmidrule(lr){5-6}\cmidrule(lr){7-8}
 & & S & M & S & M & S & M & \\
\midrule
MISO & GPT-6.1 Sol & 28.293 & 28.293 & 46.001 & 46.146 & 52.960 & 53.217 & 6/6 \\
 & Opus 5.5 & 28.293 & 28.293 & 46.004 & 46.004 & 52.961 & 53.217 & 6/6 \\
\addlinespace[3pt]
ERCOT & GPT-6.1 Sol & 18.225 & 18.225 & 68.058 & 68.076 & 82.980 & 82.958 & 6/6 \\
 & Opus 5.5 & 18.225 & 18.225 & 68.407 & 68.408 & 83.763 & 83.151 & 6/6 \\
\addlinespace[3pt]
NYISO & GPT-6.1 Sol & 8.030 & 8.030 & 6.478 & 6.478 & 6.830 & 6.830 & 6/6 \\
 & Opus 5.5 & 8.030 & 8.030 & 6.478 & 6.478 & 6.830 & 6.830 & 6/6 \\
\addlinespace[3pt]
South Africa & GPT-6.1 Sol & 10.647 & 10.648 & 12.018 & 12.019 & 14.098 & 14.098 & 6/6 \\
 & Opus 5.5 & 10.652 & 10.656 & 12.052 & 12.020 & 14.202 & 14.101 & 6/6 \\
\addlinespace[3pt]
Australia (NEM) & GPT-6.1 Sol & 7.053 & 7.053 & 6.861 & 6.861 & 8.853 & 8.854 & 6/6 \\
 & Opus 5.5 & 7.053 & 7.053 & 6.861 & 6.861 & 8.854 & 8.854 & 6/6 \\
\addlinespace[3pt]
New Zealand & GPT-6.1 Sol & 0.569 & 0.569 & 0.647 & 0.647 & 0.800 & 0.800 & 6/6 \\
 & Opus 5.5 & 0.569 & 0.569 & 0.647 & 0.647 & 0.800 & 0.800 & 6/6 \\
\addlinespace[3pt]
Mexico & GPT-6.1 Sol & 19.552 & 19.552 & 22.258 & 22.258 & 26.409 & 26.412 & 6/6 \\
 & Opus 5.5 & 19.553 & 19.555 & 22.258 & 22.258 & 26.409 & 26.412 & 6/6 \\
\addlinespace[3pt]
Brazil & GPT-6.1 Sol & 15.870 & 15.870 & 24.655 & 24.655 & 33.736 & 33.736 & 6/6 \\
 & Opus 5.5 & 15.870 & 15.870 & 24.655 & 24.655 & 33.736 & 33.736 & 6/6 \\
\bottomrule
\end{tabular*}}
\end{tableorg}

\clearpage
\subsection{Accepted planning portfolios retain their geographic context}\label{sec:geographic-results}

\begin{figure}[p]
\centering
\makebox[\textwidth][c]{\includegraphics[width=167mm]{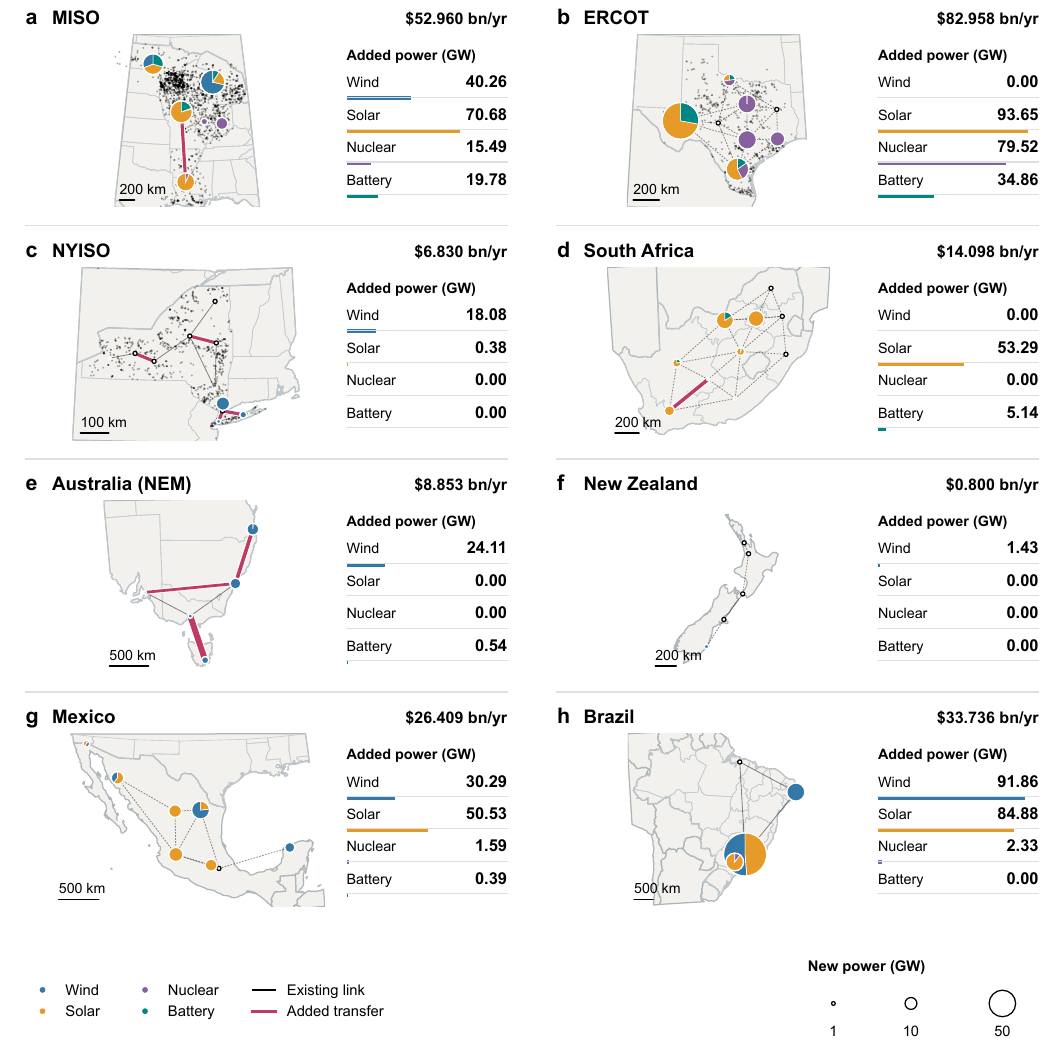}}
\caption{\textbf{Lowest-cost accepted regional designs for 2035.} a--h, Minimum accepted annual costs among four trajectories per system (billion USD). Power bars share a 0--100~GW scale; sector area encodes added generation and battery power. Colours identify technologies; hollow points mark additions below 0.005~GW. Black points locate US plants; black corridors are existing, dashed for assumed ratings. Magenta widths encode added transfer capacity. New allocations and straight corridors have model-zone resolution. All 96 outcomes remain tabulated; Appendix~\ref{app:regional-atlas} gives zonal amounts. The basemap uses Natural Earth\cite{naturalearth50m}.}
\label{fig:geographic-planning}
\end{figure}

The same planning loop produces accepted investment-and-dispatch candidates in all eight regional systems. Tables~\ref{tab:complete-coverage} and~\ref{tab:all-selected-costs} retain all 96 trajectories. For the geographic views, we compare the four completed 2035 trajectories in each system---GPT-6.1 Sol and Opus 5.5, each with single- and multi-agent architectures---using identical regional demand, resource, inventory and network inputs. Figure~\ref{fig:geographic-planning} maps the lowest observed selected total annual cost among their physically accepted candidates. GPT-6.1 Sol single-agent supplies the displayed MISO, South Africa, Australia and Mexico designs, and GPT-6.1 Sol multi-agent supplies ERCOT. In NYISO, New Zealand and Brazil, all four trajectories have the same reported minimum cost; an Opus multi-agent portfolio is retained as a stable representative of each tie. Additional inter-area transfer capacity is selected in MISO (1.00~GW), NYISO (4.00~GW), South Africa (1.20~GW) and Australia (4.61~GW). These values sum selected corridor increments rather than net import capabilities. Appendix~\ref{app:regional-atlas} provides enlarged maps and a complete zonal installation ledger for each displayed design.

Across these eight selected designs, new generation ranges from 1.43~GW in New Zealand to 179.07~GW in Brazil. The geographic pattern reflects the declared zonal demand, resource realization, inherited fleet and investment bounds. It demonstrates that the framework returns explicit regional planning outputs across heterogeneous input packs. New assets are represented by zonal allocations.

Opus multi-agent designs represent four-way cost ties in NYISO, New Zealand and Brazil.

\subsection{Feeder evidence changes the accepted planning output}\label{sec:ercot-spatial}

\begin{figure}[tbp]
\centering
\makebox[\textwidth][c]{\includegraphics[width=165mm]{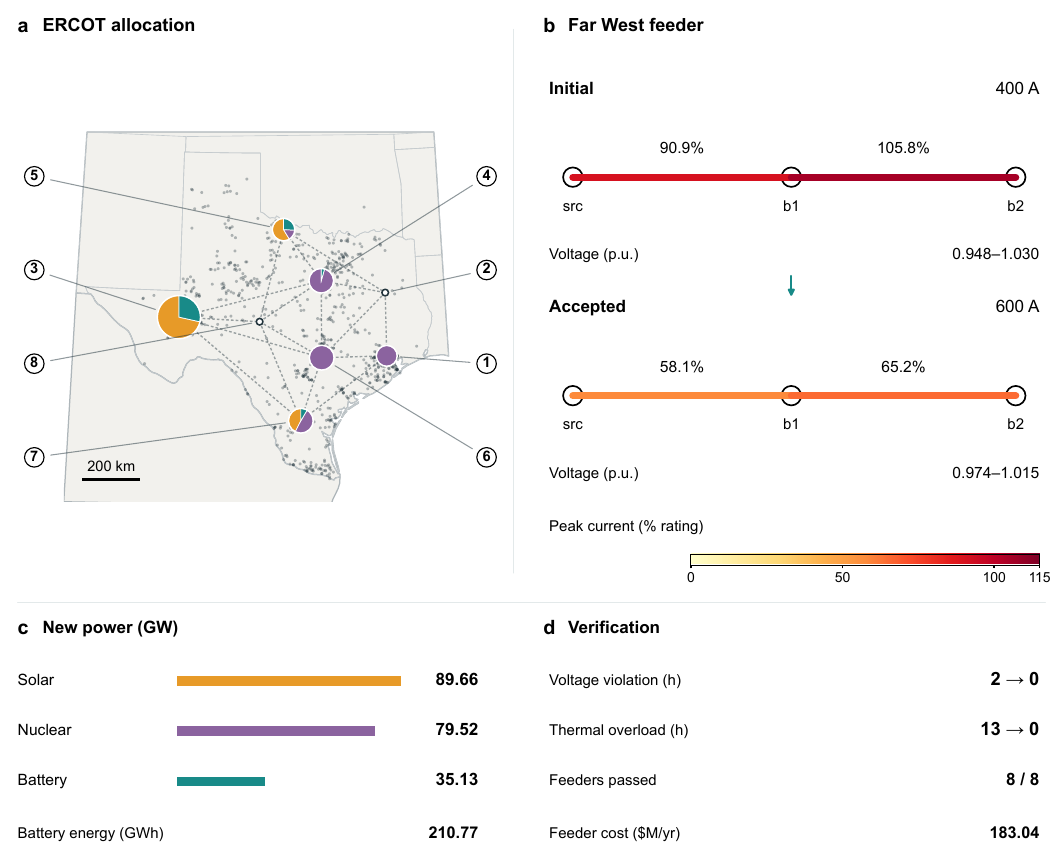}}
\caption{\textbf{ERCOT allocation and representative-feeder repair in an Opus multi-agent trajectory.} This process example is distinct from the lower-cost Sol ERCOT design in Fig.~\ref{fig:geographic-planning} and Appendix~\ref{app:regional-atlas}. a, New generation and battery power at numbered model zones, with the same technology colour key. b, Initial and selected Far West three-bus numerical circuits; line ratings increase from 400 to 600~A and impedances are halved under the intervention model. Line colours and percentages use annual peak current divided by each line's own rating. Annual voltage ranges are shown separately; extrema can occur at different hours. c, Selected additions and battery energy; no new inter-area transfer is selected. d, Recorded violation hours, accepted feeder count and annualized intervention cost (million USD per year). All eight selected representative feeders pass. The selected design combines feeder reinforcement with a revised portfolio and dispatch.}
\label{fig:ercot-verified}
\end{figure}

A separate Opus 5.5 multi-agent ERCOT 2035 trajectory illustrates how the framework links regional capacity choices to a physical check outside the annual system objective (Fig.~\ref{fig:ercot-verified}). Its initial Far West representative feeder has two voltage-violation hours and thirteen thermal-overload hours. The annual minimum voltage is 0.948~p.u., below the 0.95~p.u. lower limit, and the second line reaches 105.8\% of its 400~A rating. This candidate fails the distribution gate.

The selected accepted candidate combines a changed portfolio and battery operation with the Far West line-upgrade decision. The upgrade increases both representative-line ratings from 400 to 600~A and halves their impedances under the declared intervention model. The final Far West feeder has no voltage violations, thermal overloads or nonconverged hours; its annual voltage range is 0.974--1.015~p.u., and the second line's annual peak current is 65.2\% of its upgraded rating. All eight ERCOT representative feeders pass.

The committed regional result contains 89.662~GW of new solar, 79.519~GW of new nuclear and 35.128~GW of new battery power with 210.770~GWh of battery energy. It selects no additional inter-area transmission capacity. The annualized distribution intervention is USD~183.04 million. These outcomes connect zonal additions, a named feeder intervention and the resulting acceptance record.

\subsection{Search histories distinguish selection from continued exploration}\label{sec:search-process}

The 96 trajectories contain 382 committed candidate assessments, including 369 system-simulation requests and thirteen additional feeder-only assessments. Of these, 307 pass the recorded acceptance gate and 75 fail its distribution component. The median trajectory contains three assessments (range, one to eighteen). The selected candidate first appears at assessment one in 56 trajectories and later in 40; 35 trajectories improve the objective after their first accepted candidate. In 28 trajectories, the selected candidate's source assessment precedes the last assessment. Fifteen of these end by rechecking the same design, whereas thirteen end with a different design. Thus, the source of the selected outcome and the final evaluated proposal convey different information about the search.

Sixteen trajectories produce 36 subsequent assessments with a lower recorded objective than an already accepted incumbent but fail distribution validation. In the Opus ERCOT 2035 multi-agent example (Fig.~\ref{fig:ercot-verified}), round four supplies the selected accepted design. Removing feeder reinforcement in round six lowers the objective by 0.220\% but introduces three undervoltage hours. Round seven restores the accepted design, leaving round four as its selected source. A further 228 uncommitted system-only probes inform exploration across the cohort; their results cover the system model and are reported separately from committed feeder assessments. These records show the predefined validation-and-retention rule operating throughout agent-directed exploration.

\subsection{Coverage and reproducible numerical evidence}\label{sec:coverage}

The year-matched evaluation contains 96 completed annual trajectories. Thirty-two pair 2023 demand with ERA5 2023 resources; sixty-four pair 2030 or 2035 demand with resource profiles reconstructed for the same target year. Every trajectory yields a committed candidate satisfying the implemented system and representative-feeder acceptance gate. The eighty newly completed hosted trajectories also undergo independent reconstruction of committed candidate selection and recomputation of the physical gate from their retained numerical outputs. The evidence package preserves input and implementation identities, accepted candidates, costs, stopping decisions and recorded search effort. Appendix~\ref{app:outcomes} reports the complete descriptive outcome tables and deployment comparisons.

\subsection{Token usage characterizes the effort of the planning loop}\label{sec:token-results}

The complete reported search histories contain 2,257 distinct received model responses, totalling 276.386 million input tokens and 3.834 million output tokens (Table~\ref{tab:search-token-usage}). Input includes cache reads and writes; output includes the reported reasoning subset. Across 48 trajectories per architecture, multi-agent input and output totals are respectively 2.08 and 23.34 times their single-agent counterparts. Every one of the 48 pairs matched by deployment, system and year uses more output tokens in the multi-agent configuration. The median paired output ratio is 5.94 for Sol and 25.10 for Opus. Input increases in 28 pairs; for Sol, the median paired input ratio is 0.74 despite an aggregate ratio of 1.55, reflecting a small number of large search histories. Totals, means and medians therefore describe distinct aspects of the observed effort. These comparisons describe search effort in the configured deployments.

\begin{tableorg}[tbp]
\caption{\textbf{Input and output tokens across all 96 loose-cap reference trajectories.} Input columns use millions of tokens; output columns use thousands. Means and medians are per trajectory. Counts cover complete search histories, including repeated context and inherited responses counted once. Input includes cache reads and writes; output includes reasoning. The archived evidence retains integer token counts for every response.}
\label{tab:search-token-usage}
\centering\small
\makebox[\textwidth][c]{\begin{tabular*}{170mm}{@{\extracolsep{\fill}}llrrrrrrr@{}}
\toprule
Deployment & Architecture & $n$ & \multicolumn{3}{c}{Input tokens (million)} & \multicolumn{3}{c}{Output tokens (thousand)} \\
\cmidrule(lr){4-6}\cmidrule(lr){7-9}
 & & & Total & Mean & Median & Total & Mean & Median \\
\midrule
GPT-6.1 Sol & Single & 24 & 74.854 & 3.119 & 1.756 & 91.667 & 3.819 & 3.437\\
GPT-6.1 Sol & Multi & 24 & 116.006 & 4.834 & 0.495 & 1,364.790 & 56.866 & 11.160\\
Opus 5.5 & Single & 24 & 14.914 & 0.621 & 0.428 & 65.864 & 2.744 & 2.165\\
Opus 5.5 & Multi & 24 & 70.611 & 2.942 & 0.724 & 2,311.868 & 96.328 & 33.580\\
\bottomrule
\end{tabular*}}
\end{tableorg}

Appendix~\ref{app:search-tokens} gives assessment counts, selected assessment indices, rejected assessments, probes and token usage for every trajectory. The histories also record 142 applied multi-agent advisory-graph edits and 129 single-agent strategy updates. These records capture changes to the search procedure. Among the 44 matched pairs with complete duration records, the median multi/single runtime ratio is 1.84 for Sol and 4.01 for Opus; the four resumed trajectories are excluded from duration summaries but retain their full reconstructed token histories.

\subsection{Reconstruction of the reference outcome evidence}

The reported evidence comprises sixteen Claude Opus 5.5 trajectories with 2023 demand and ERA5 2023 weather, plus forty-eight GPT-6.1 Sol and thirty-two Opus trajectories from the completed hosted-weather campaign. The latter includes sixteen Sol 2023 ERA5 controls and sixty-four future-weather cases. A reporting manifest selects these 96 trajectories, each with matching demand and resource-weather years. Extraction reconstructs committed candidates, applies the common acceptance gate and selects the minimum-cost accepted candidate in Eq.~\ref{eq:selection}. Source manifests link the selected design to its system result, dispatch, feeder checks and model-response records. For all eighty new cases, reconstruction of the full committed history yields the same selected design and cost as the archived final; physical acceptance is recomputed from the recorded numerical results without new inference or optimization.

The thirty-two Sol--Opus pairs in the future-weather cohort share input and physical-code hashes and numerical settings, and all sixteen single-agent initial exposed states match. The 2023 controls retain campaign-specific inference settings. Comparisons pair deployment outcomes by system, demand year, weather source and architecture, with candidate-specific numerical bounds retained. All comparisons contain one trajectory per cell. Solver bounds apply to the optimization problem within each selected design domain.

\clearpage
\section{Loose-cap reference outcomes and evaluation effort}\label{app:outcomes}
\setcounter{figure}{0}
These records preserve the achieved outcomes and execution effort of the ninety-six loose-cap reference trajectories. They document the verified workflow under the recorded deployment and inference settings.
\renewcommand{\thefigure}{\Alph{section}\arabic{figure}}
\renewcommand{\theHfigure}{appendix.\Alph{section}.\arabic{figure}}
\setcounter{figure}{0}
\subsection{Opus planning with 2023 demand and ERA5 2023 weather}\label{sec:opus-base}
\begin{figure}[tbp]
\centering
\includegraphics[width=\textwidth]{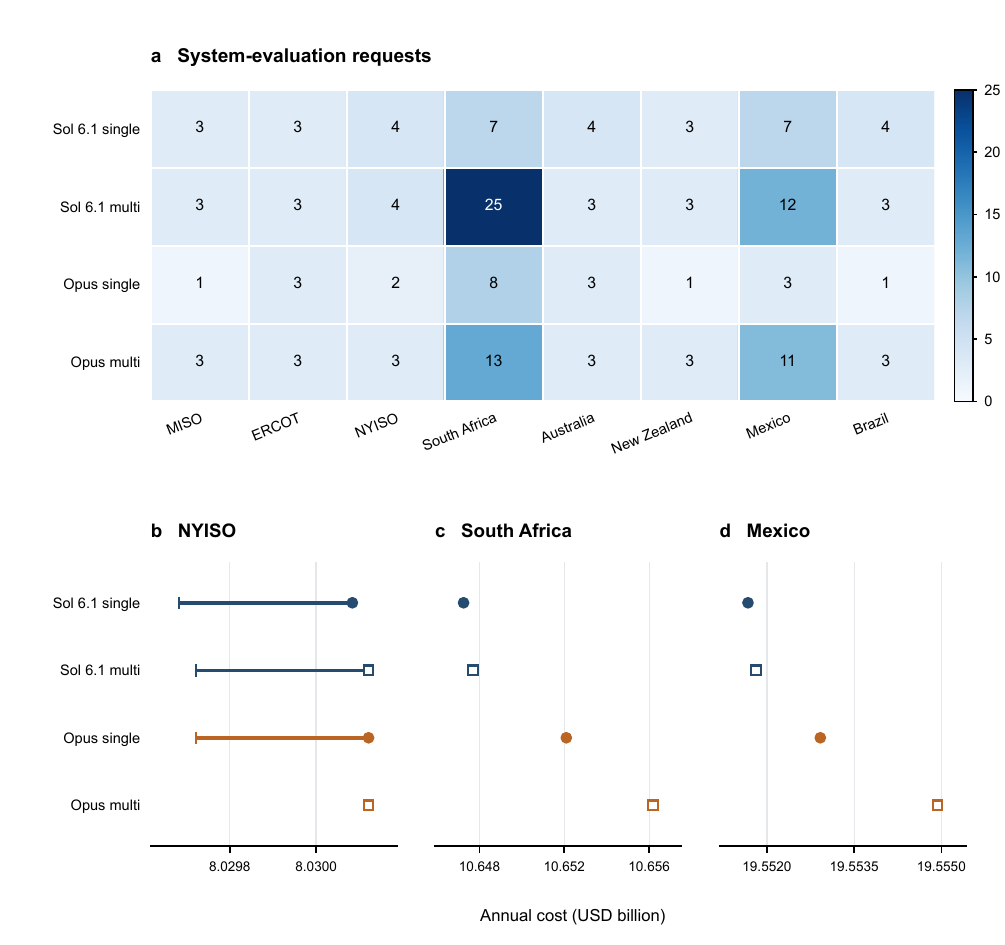}
\caption{Two deployments under 2023 demand and ERA5 2023 weather. a, System-evaluation requests across 32 trajectories, comprising committed simulations and hypothetical verifications, including exact-result reuse. Cell values are exact and colour uses one linear scale. b--d, Selected all-in annual costs in NYISO, South Africa and Mexico on panel-specific scales. Filled circles denote single-agent and open squares multi-agent trajectories. Blue denotes GPT-6.1 Sol and orange Claude Opus 5.5. Horizontal intervals extend from the candidate's solver lower bound plus its distribution cost to its selected total cost; the magnified axes condition on each selected design.}
\label{fig:model-comparison}
\end{figure}

Claude Opus 5.5 runs sixteen trajectories with 2023 demand and ERA5 2023 weather, with both architectures in each of the eight systems (Table~\ref{tab:opus-2023}). Every selected candidate passes the same code-computed system and representative-feeder conditions. Selection retains the lowest all-in annual cost among physically accepted committed candidates. Together with the sixteen GPT-6.1 Sol 2023 controls, these results give thirty-two baseline trajectories (Fig.~\ref{fig:model-comparison}).

\begin{table}[tbp]
\caption{Claude Opus 5.5 planning with 2023 demand and ERA5 2023 weather. Entries are single/multi. Selected costs are billion USD per year, including representative-distribution interventions. System evaluations count committed and uncommitted verification requests, including exact-result reuse; model requests count unique received responses in the selected trajectory ledger. }
\label{tab:opus-2023}
\centering\small
\begin{tabular}{@{}lrrr@{}}
\toprule
System & Selected cost & System evaluations & Model requests\\
\midrule
MISO & 28.293256/28.293256 & 1/3 & 3/11\\
ERCOT & 18.225239/18.225239 & 3/3 & 6/11\\
NYISO & 8.030121/8.030121 & 2/3 & 5/21\\
South Africa & 10.652093/10.656181 & 8/13 & 11/55\\
Australia & 7.052820/7.052820 & 3/3 & 7/11\\
New Zealand & 0.569150/0.569150 & 1/3 & 4/11\\
Mexico & 19.552919/19.554932 & 3/11 & 6/44\\
Brazil & 15.869564/15.869564 & 1/3 & 3/11\\
\bottomrule
\end{tabular}
\end{table}

The 2023-demand cohort records 38 committed evaluations, 26 verification requests and 220 received model responses. Solver cost intervals condition on each selected design. Each cell contains one trajectory. Model protocols, search ceilings and implementations are listed in Appendix~\ref{app:deployments}.
\subsection{Planning with projected weather and matched demand}\label{sec:weather-results}
The completed hosted campaign adds eighty annual trajectories, comprising sixteen GPT-6.1 Sol controls with 2023 demand and ERA5 2023 weather, and sixty-four future-weather trajectories with GPT-6.1 Sol and Claude Opus 5.5. Each future deployment covers eight systems, 2030 and 2035 demand, and both architectures. NASA MPI-ESM1-2-HR SSP2-4.5 daily fields and ERA5 2023 intraday templates supply the target-year resource profiles (Methods, Section~\ref{sec:future-weather}). All eighty selected candidates pass the recomputed system and representative-feeder acceptance gate. Full committed-history reconstruction confirms that each selected result is the lowest-cost physically accepted committed candidate. All trajectories end through their single-agent finish decision or multi-agent judge.

\begin{table}[tbp]
\caption{Selected annual cost under 2030 demand and NASA-derived 2030 weather. Entries are single/multi, in billion USD per year including annualized representative-distribution interventions. Each cell is one trajectory; candidate-specific solver intervals are retained in the comparison evidence.}
\label{tab:weather-2030}
\centering\small
\begingroup\setlength{\tabcolsep}{4pt}
\begin{tabular}{@{}lrr@{}}
\toprule
System & GPT-6.1 Sol & Claude Opus 5.5\\
\midrule
MISO & 46.000616/46.145542 & 46.003694/46.003694\\
ERCOT & 68.057741/68.076200 & 68.407019/68.407550\\
NYISO & 6.477653/6.477653 & 6.477653/6.477653\\
South Africa & 12.017688/12.019324 & 12.052166/12.020040\\
Australia & 6.861099/6.861099 & 6.861099/6.861099\\
New Zealand & 0.646734/0.646734 & 0.646734/0.646734\\
Mexico & 22.257606/22.258090 & 22.258090/22.258090\\
Brazil & 24.655438/24.655438 & 24.655438/24.655438\\
\bottomrule
\end{tabular}
\endgroup
\end{table}

\begin{table}[tbp]
\caption{Selected annual cost under 2035 demand and NASA-derived 2035 weather. Entries are single/multi, in billion USD per year including annualized representative-distribution interventions. Each cell is one trajectory; candidate-specific solver intervals are retained in the comparison evidence.}
\label{tab:weather-2035}
\centering\small
\begingroup\setlength{\tabcolsep}{4pt}
\begin{tabular}{@{}lrr@{}}
\toprule
System & GPT-6.1 Sol & Claude Opus 5.5\\
\midrule
MISO & 52.959760/53.216539 & 52.961390/53.216539\\
ERCOT & 82.980096/82.957622 & 83.762708/83.150660\\
NYISO & 6.829947/6.829947 & 6.829947/6.829947\\
South Africa & 14.097805/14.098005 & 14.202359/14.101251\\
Australia & 8.853329/8.854495 & 8.854495/8.854495\\
New Zealand & 0.799669/0.799669 & 0.799669/0.799669\\
Mexico & 26.408564/26.411673 & 26.408751/26.411673\\
Brazil & 33.735917/33.735917 & 33.735917/33.735917\\
\bottomrule
\end{tabular}
\endgroup
\end{table}

\begin{figure}[tbp]
\centering
\includegraphics[width=\textwidth]{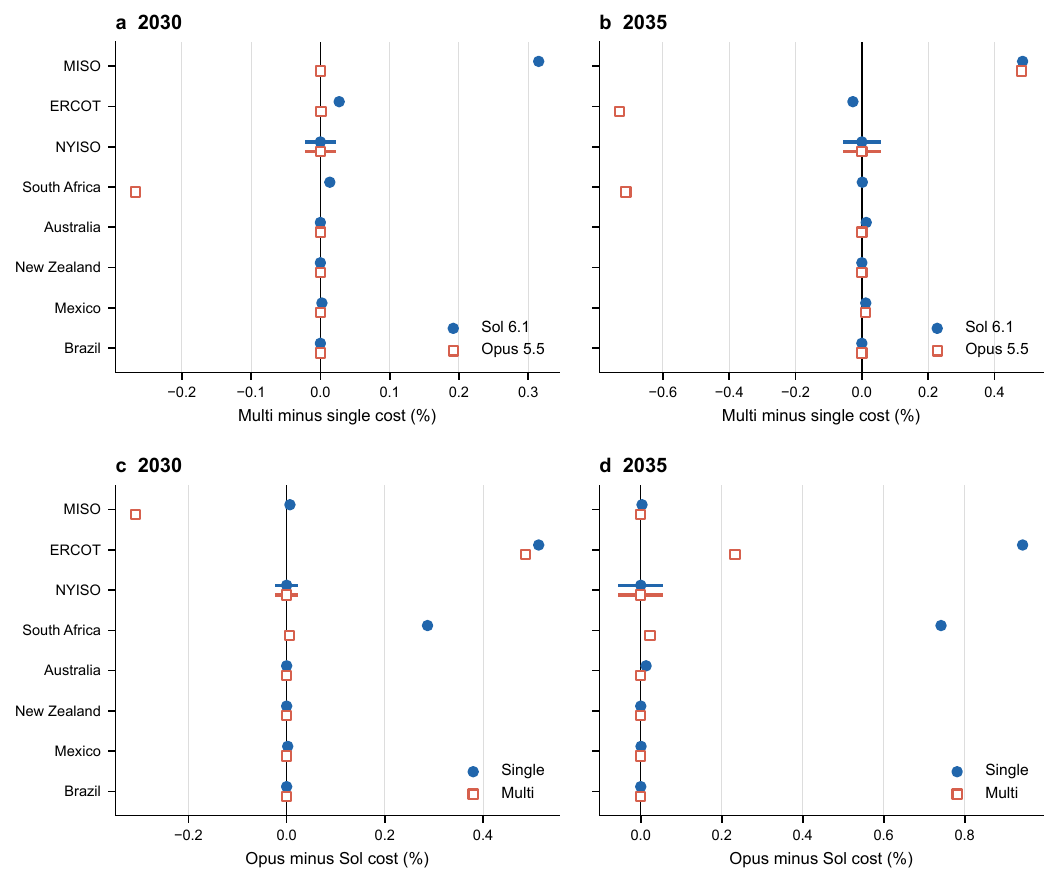}
\caption{Selected-cost comparisons with NASA-derived future weather. a,b, Multi minus single selected annual cost, as a percentage of the single-agent cost, for GPT-6.1 Sol and Claude Opus 5.5 under 2030 and 2035 demand and weather. c,d, Opus minus Sol selected annual cost as a percentage of the Sol cost, for the two architectures. Intervals extend from the first candidate's solver lower bound minus the second candidate's achieved cost to the first candidate's achieved cost minus the second candidate's lower bound. Each candidate's distribution cost is included at both endpoints. The intervals describe numerical bounds conditional on the two selected design domains. Negative values favour multi in a,b and Opus in c,d. Each point represents one matched pair.}
\label{fig:weather-comparison}
\end{figure}

Across the 32 matched future-weather deployment pairs, Opus selects a lower annual cost in 1, an equal cost to the nearest cent in 18, and a higher cost in 13. The candidate-specific solver intervals overlap in 18 pairs. Shared demand, weather, inventory and network hashes, numerical settings and physical-code hashes are verified for all 32 pairs; all sixteen single-agent initial exposed states also match. Figure~\ref{fig:weather-comparison} shows achieved costs and their candidate-conditional numerical bounds.

Selected-cost differences vary across architecture pairs. Across the sixteen future region--year architecture pairs per deployment, GPT-6.1 Sol's multi-agent candidate is cheaper in one, equal to the nearest cent in seven and more expensive in eight. Opus's corresponding counts are three, ten and three. In Opus ERCOT 2035, multi-agent selection reduces annual cost from USD 83.762708 billion to USD 83.150660 billion (USD 612.048 million; 0.731\%). In South Africa 2035, it reduces cost from USD 14.202359 billion to USD 14.101251 billion (USD 101.108 million; 0.712\%). Conversely, the MISO 2035 single-agent selections cost approximately 0.48\% less than their multi-agent counterparts in both deployments. These are achieved outcomes of individual searches, with feasibility, economic improvement and stopping effort reported separately.

\begin{table}[tbp]
\caption{Evaluation effort in the completed eighty-trajectory hosted campaign. Each row contains sixteen trajectories. Evaluations are logical requests, including exact-result reuse, rather than physical optimizer invocations. The 2023 control row uses ERA5 2023 weather; the future rows use the corresponding NASA-derived weather.}
\label{tab:weather-effort}
\centering\small
\begingroup\setlength{\tabcolsep}{4pt}
\begin{tabular}{@{}lrrrr@{}}
\toprule
Deployment/year & Committed & Verify & Responses & Median cycles\\
\midrule
Sol 6.1 2023 & 62 & 29 & 398 & 1\\
Sol 6.1 2030 & 91 & 50 & 578 & 1\\
Sol 6.1 2035 & 91 & 53 & 517 & 1\\
Opus 5.5 2030 & 41 & 29 & 254 & 1\\
Opus 5.5 2035 & 46 & 41 & 290 & 1\\
\bottomrule
\end{tabular}
\endgroup
\end{table}

Across the new hosted campaign, GPT-6.1 Sol records 244 committed and 132 verification requests, with 1,493 unique received responses. Opus records 87 committed and 70 verification requests, with 544 responses. The median multi-agent search length is one cycle in each deployment/year cohort; the longest new searches occur in Sol ERCOT 2030 (eighteen cycles) and South Africa 2030 (fifteen). Ledger attribution counts each inherited or newly received response once; three resumed trajectories have no uninterrupted wall-runtime estimate.

The sixteen GPT-6.1 Sol base-year controls also yield accepted candidates (Table~\ref{tab:sol61-control}). Five systems retain identical selected costs across architectures; NYISO, Mexico and South Africa have slightly higher multi-agent selected costs. The sixteen Opus 2023 trajectories complete the 32-case baseline comparison. Their recorded default response ceiling is 16,000 tokens, compared with 64,000 tokens in the GPT-6.1 Sol campaign.

\begin{table}[tbp]
\caption{GPT-6.1 Sol controls with 2023 demand and ERA5 2023 weather. Entries are single/multi. Selected costs are billion USD per year including distribution interventions; evaluations include committed and hypothetical requests.}
\label{tab:sol61-control}
\centering\small
\begingroup\setlength{\tabcolsep}{4pt}
\begin{tabular}{@{}lrrr@{}}
\toprule
System & Selected cost & Evaluations & Responses\\
\midrule
MISO & 28.293256/28.293256 & 3/3 & 9/12\\
ERCOT & 18.225239/18.225239 & 3/3 & 9/11\\
NYISO & 8.030084/8.030121 & 4/4 & 14/21\\
South Africa & 10.647249/10.647689 & 7/25 & 23/149\\
Australia & 7.052820/7.052820 & 4/3 & 12/11\\
New Zealand & 0.569150/0.569150 & 3/3 & 9/11\\
Mexico & 19.551675/19.551810 & 7/12 & 21/64\\
Brazil & 15.869564/15.869564 & 4/3 & 11/11\\
\bottomrule
\end{tabular}
\endgroup
\end{table}
\clearpage
\section{Loose-cap reference regional installation atlas}\label{app:regional-atlas}
\setcounter{figure}{0}
These eight 2035 portfolios retain the original 1,000~Mt annual ceiling. Carbon-constrained geography is reported separately in Fig.~\ref{fig:carbon-geography} and the accompanying carbon atlas.
\renewcommand{\thefigure}{\Alph{section}\arabic{figure}}
\renewcommand{\theHfigure}{appendix.\Alph{section}.\arabic{figure}}
\setcounter{figure}{0}
\begin{figure}[!ht]
\centering
\makebox[\textwidth][c]{\includegraphics[width=155mm]{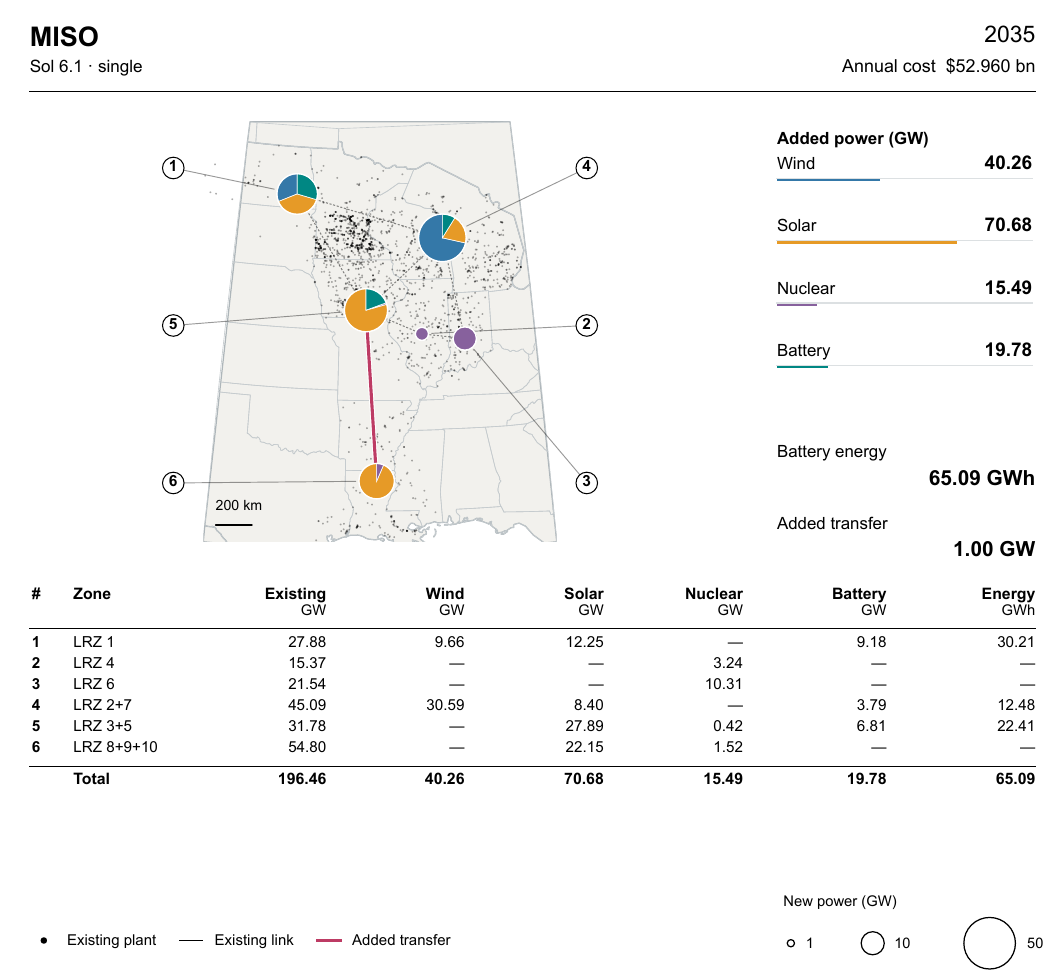}}
\caption{\textbf{Lowest-cost accepted 2035 design for MISO.} Lowest observed selected total annual cost among four physically accepted trajectories using identical 2035 demand and NASA resource inputs. The displayed portfolio is GPT-6.1 Sol, single-agent, at USD 52.960 billion per year. Map symbols follow Fig.~\ref{fig:geographic-planning}; numbers link to the zonal ledger. Existing is inherited generation; Energy is added battery energy; dashes are below 0.005 in the displayed units. Additional transfer capacity is 1.00~GW on 0035--8910. All represented feeder checks pass. Locations and corridors have model-zone resolution. The basemap uses Natural Earth.}
\label{fig:atlas-miso}
\end{figure}

\clearpage
\begin{figure}[p]
\centering
\makebox[\textwidth][c]{\includegraphics[width=170mm]{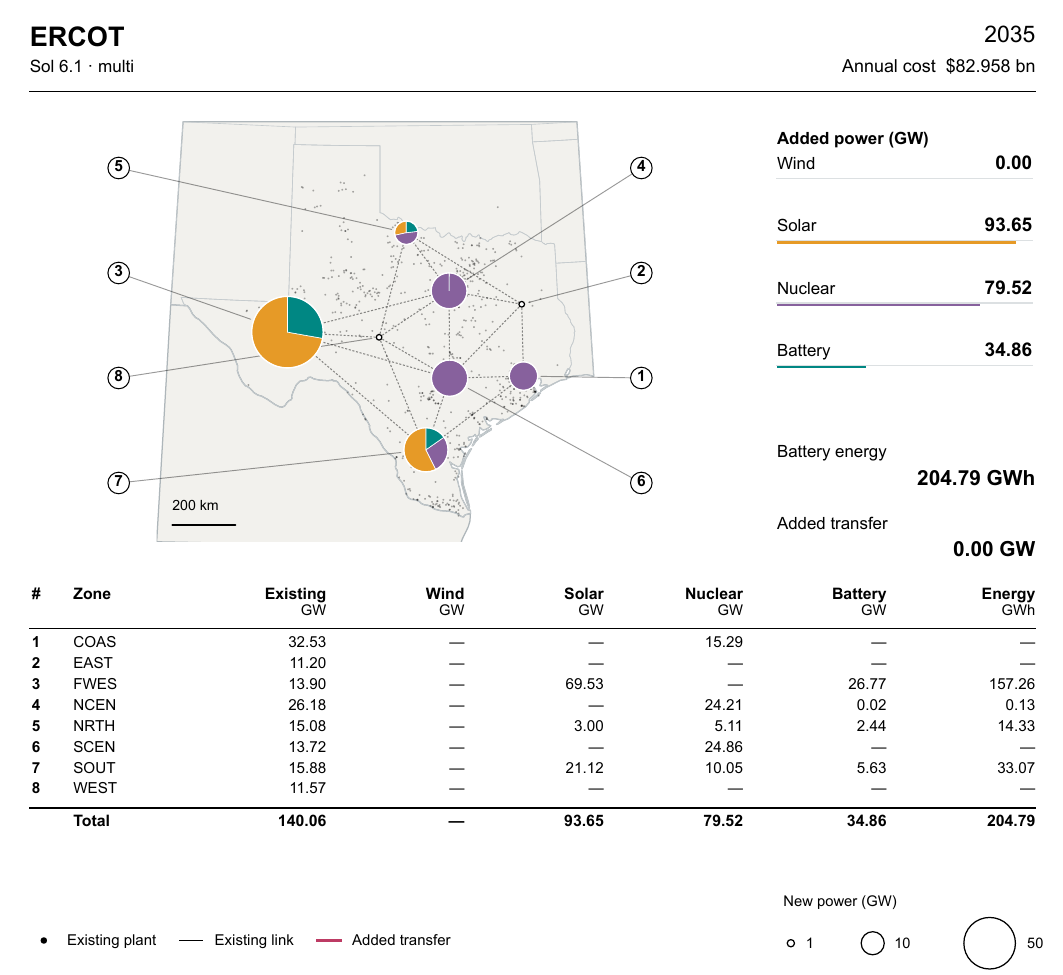}}
\caption{\textbf{Lowest-cost accepted 2035 design for ERCOT.} Lowest observed selected total annual cost among four physically accepted trajectories using identical 2035 demand and NASA resource inputs. The displayed portfolio is GPT-6.1 Sol, multi-agent, at USD 82.958 billion per year. Map symbols follow Fig.~\ref{fig:geographic-planning}; numbers link to the zonal ledger. Existing is inherited generation; Energy is added battery energy; dashes are below 0.005 in the displayed units. No additional transfer capacity is selected. All represented feeder checks pass. Locations and corridors have model-zone resolution. The basemap uses Natural Earth.}
\label{fig:atlas-ercot}
\end{figure}

\clearpage
\begin{figure}[p]
\centering
\makebox[\textwidth][c]{\includegraphics[width=170mm]{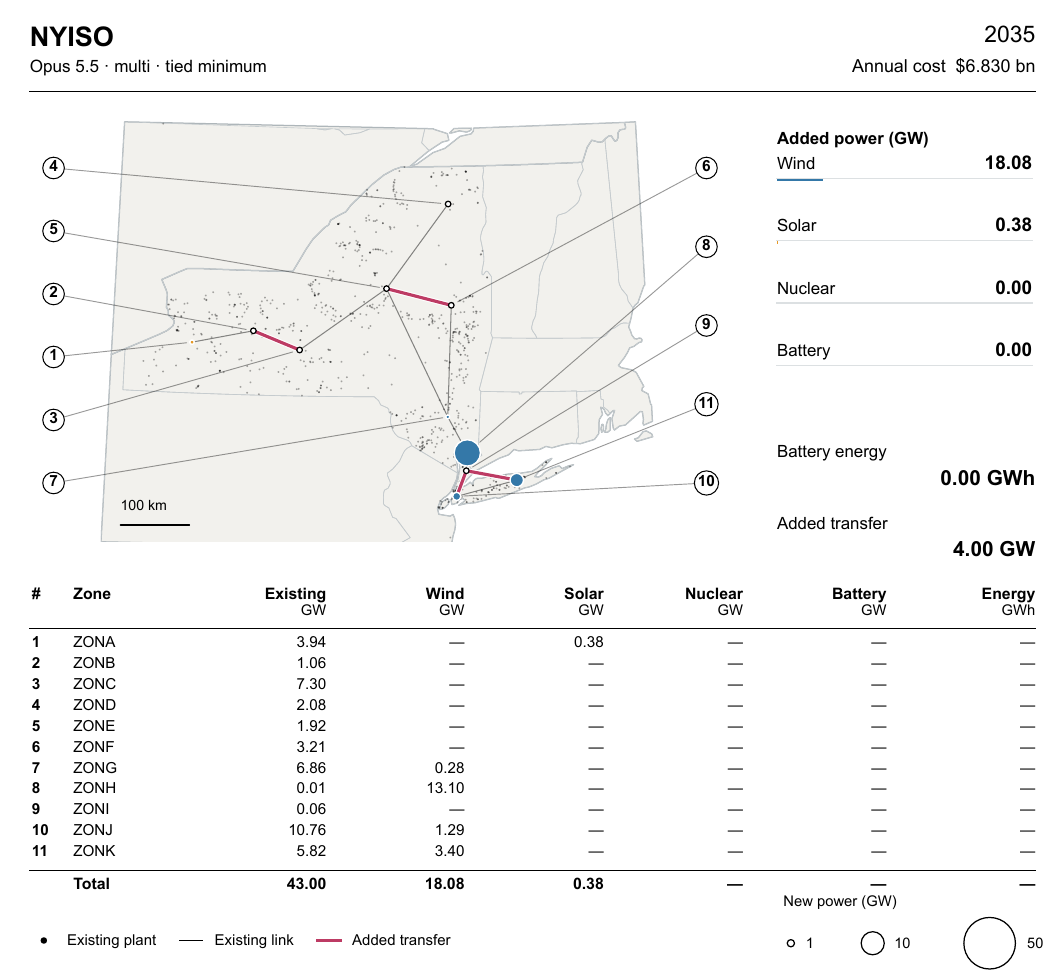}}
\caption{\textbf{Lowest-cost accepted 2035 design for NYISO.} Lowest observed selected total annual cost among four physically accepted trajectories using identical 2035 demand and NASA resource inputs. The displayed portfolio is Claude Opus 5.5, multi-agent, at USD 6.830 billion per year. All four trajectories have the same reported minimum cost; this portfolio is a stable representative of the tie. Map symbols follow Fig.~\ref{fig:geographic-planning}; numbers link to the zonal ledger. Existing is inherited generation; Energy is added battery energy; dashes are below 0.005 in the displayed units. Additional transfer capacity is 1.00~GW on ZONB--ZONC, 1.00~GW on ZONE--ZONF, 1.00~GW on ZONI--ZONJ and 1.00~GW on ZONI--ZONK. All represented feeder checks pass. Locations and corridors have model-zone resolution. The basemap uses Natural Earth.}
\label{fig:atlas-nyiso}
\end{figure}

\clearpage
\begin{figure}[p]
\centering
\makebox[\textwidth][c]{\includegraphics[width=170mm]{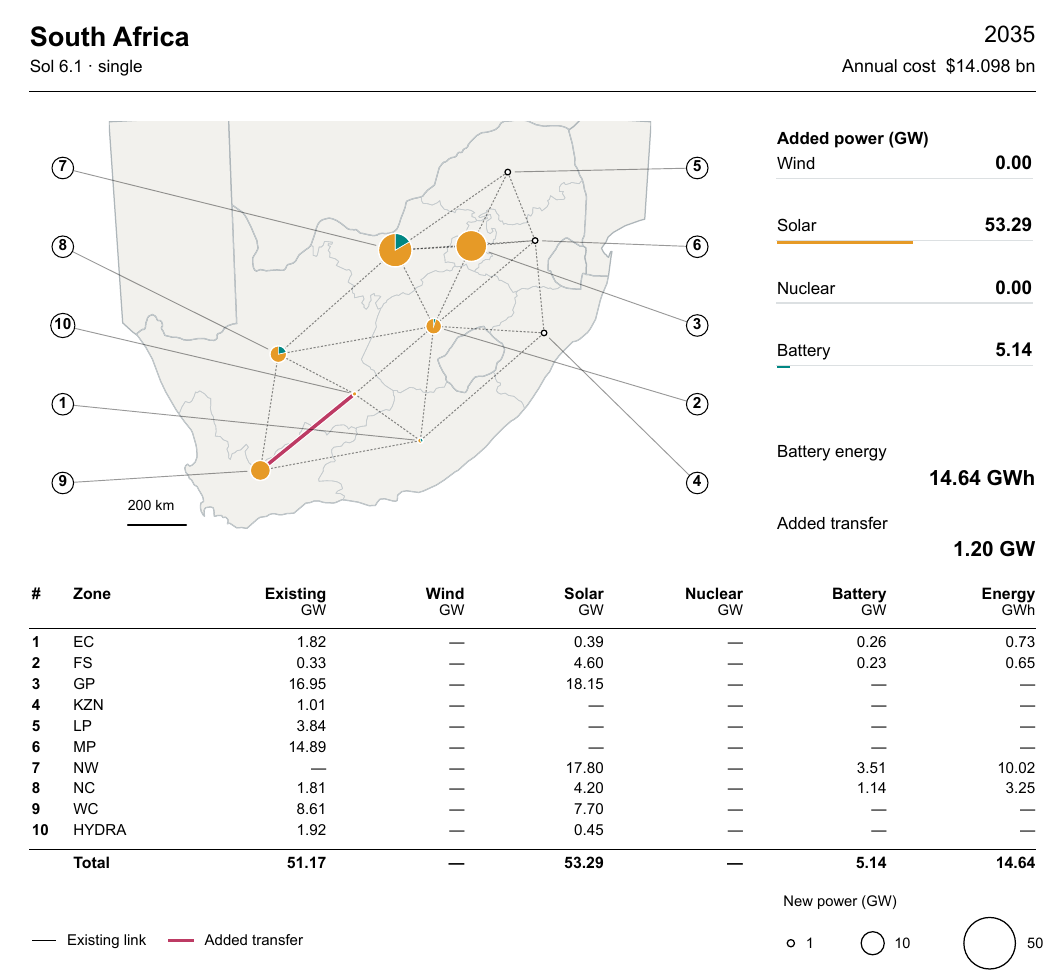}}
\caption{\textbf{Lowest-cost accepted 2035 design for South Africa.} Lowest observed selected total annual cost among four physically accepted trajectories using identical 2035 demand and NASA resource inputs. The displayed portfolio is GPT-6.1 Sol, single-agent, at USD 14.098 billion per year. Map symbols follow Fig.~\ref{fig:geographic-planning}; numbers link to the zonal ledger. Existing is inherited generation; Energy is added battery energy; dashes are below 0.005 in the displayed units. Additional transfer capacity is 1.20~GW on HYDRA--WC. All represented feeder checks pass. Locations and corridors have model-zone resolution. The basemap uses Natural Earth.}
\label{fig:atlas-south_africa}
\end{figure}

\clearpage
\begin{figure}[p]
\centering
\makebox[\textwidth][c]{\includegraphics[width=170mm]{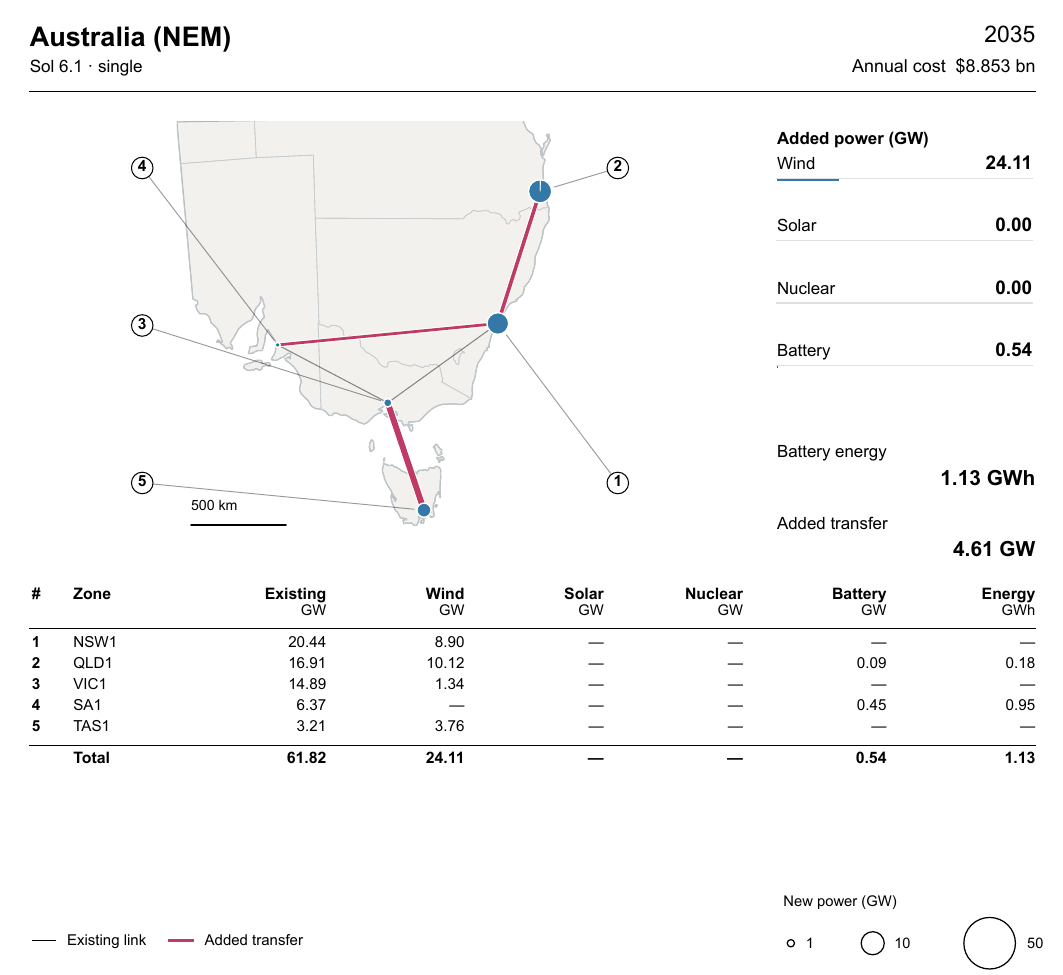}}
\caption{\textbf{Lowest-cost accepted 2035 design for Australia (NEM).} Lowest observed selected total annual cost among four physically accepted trajectories using identical 2035 demand and NASA resource inputs. The displayed portfolio is GPT-6.1 Sol, single-agent, at USD 8.853 billion per year. Map symbols follow Fig.~\ref{fig:geographic-planning}; numbers link to the zonal ledger. Existing is inherited generation; Energy is added battery energy; dashes are below 0.005 in the displayed units. Additional transfer capacity is 1.26~GW on NSW1--QLD1, 2.55~GW on VIC1--TAS1 and 0.80~GW on NSW1--SA1. All represented feeder checks pass. Locations and corridors have model-zone resolution. The basemap uses Natural Earth.}
\label{fig:atlas-australia}
\end{figure}

\clearpage
\begin{figure}[p]
\centering
\makebox[\textwidth][c]{\includegraphics[width=170mm]{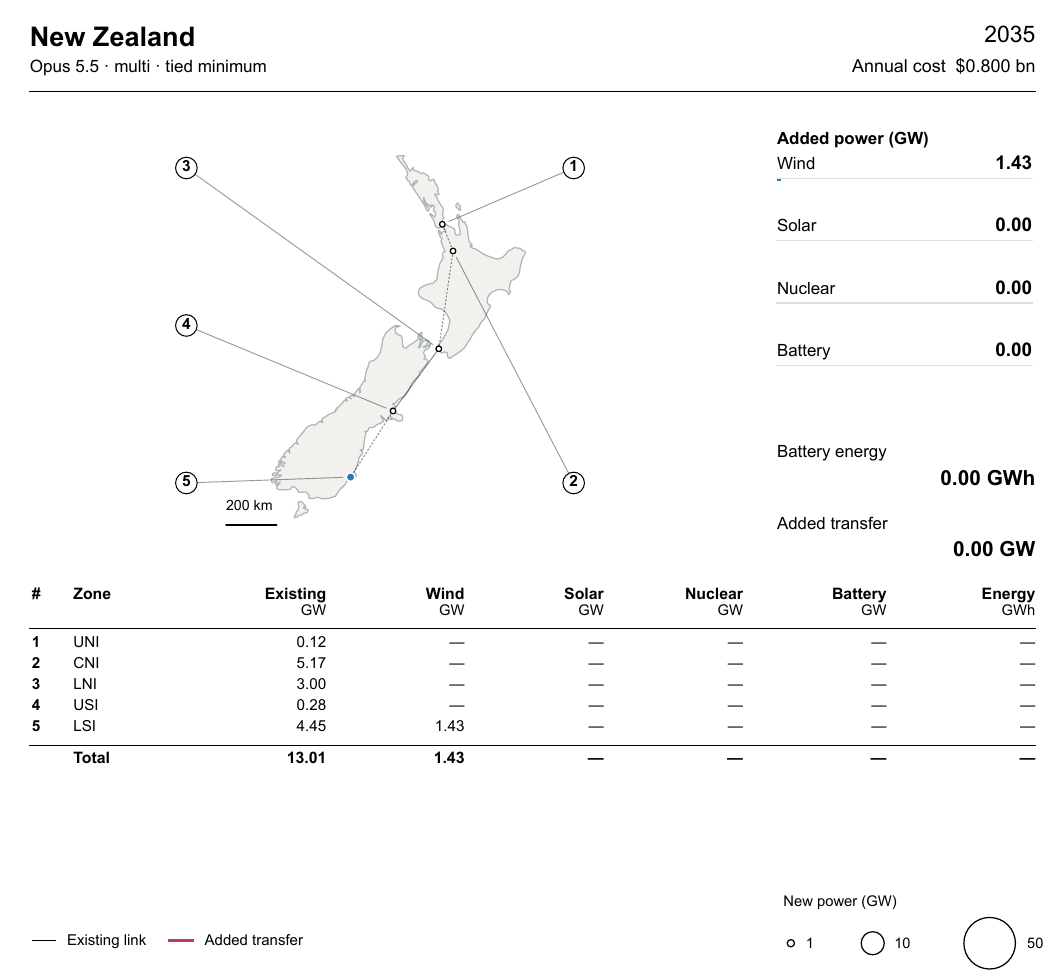}}
\caption{\textbf{Lowest-cost accepted 2035 design for New Zealand.} Lowest observed selected total annual cost among four physically accepted trajectories using identical 2035 demand and NASA resource inputs. The displayed portfolio is Claude Opus 5.5, multi-agent, at USD 0.800 billion per year. All four trajectories have the same reported minimum cost; this portfolio is a stable representative of the tie. Map symbols follow Fig.~\ref{fig:geographic-planning}; numbers link to the zonal ledger. Existing is inherited generation; Energy is added battery energy; dashes are below 0.005 in the displayed units. No additional transfer capacity is selected. All represented feeder checks pass. Locations and corridors have model-zone resolution. The basemap uses Natural Earth.}
\label{fig:atlas-new_zealand}
\end{figure}

\clearpage
\begin{figure}[p]
\centering
\makebox[\textwidth][c]{\includegraphics[width=170mm]{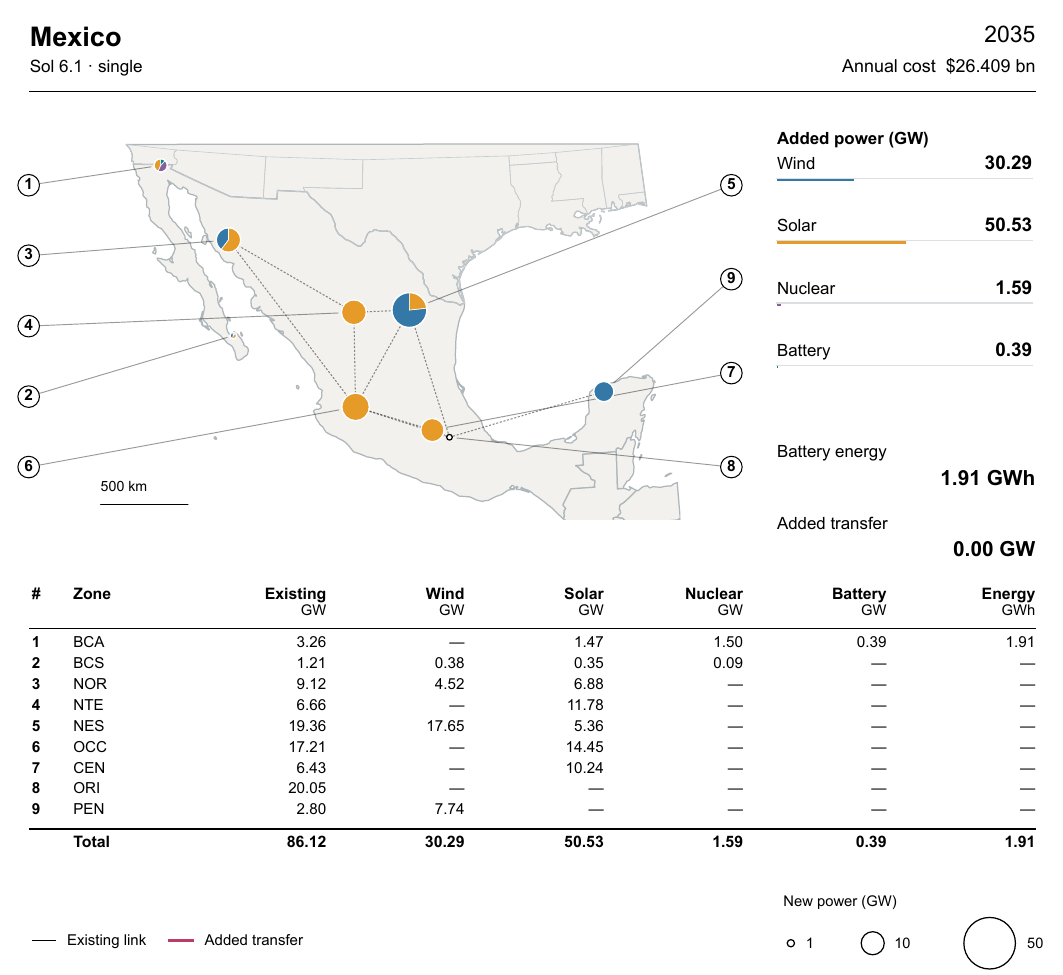}}
\caption{\textbf{Lowest-cost accepted 2035 design for Mexico.} Lowest observed selected total annual cost among four physically accepted trajectories using identical 2035 demand and NASA resource inputs. The displayed portfolio is GPT-6.1 Sol, single-agent, at USD 26.409 billion per year. Map symbols follow Fig.~\ref{fig:geographic-planning}; numbers link to the zonal ledger. Existing is inherited generation; Energy is added battery energy; dashes are below 0.005 in the displayed units. No additional transfer capacity is selected. All represented feeder checks pass. Locations and corridors have model-zone resolution. The basemap uses Natural Earth.}
\label{fig:atlas-mexico}
\end{figure}

\clearpage
\begin{figure}[p]
\centering
\makebox[\textwidth][c]{\includegraphics[width=170mm]{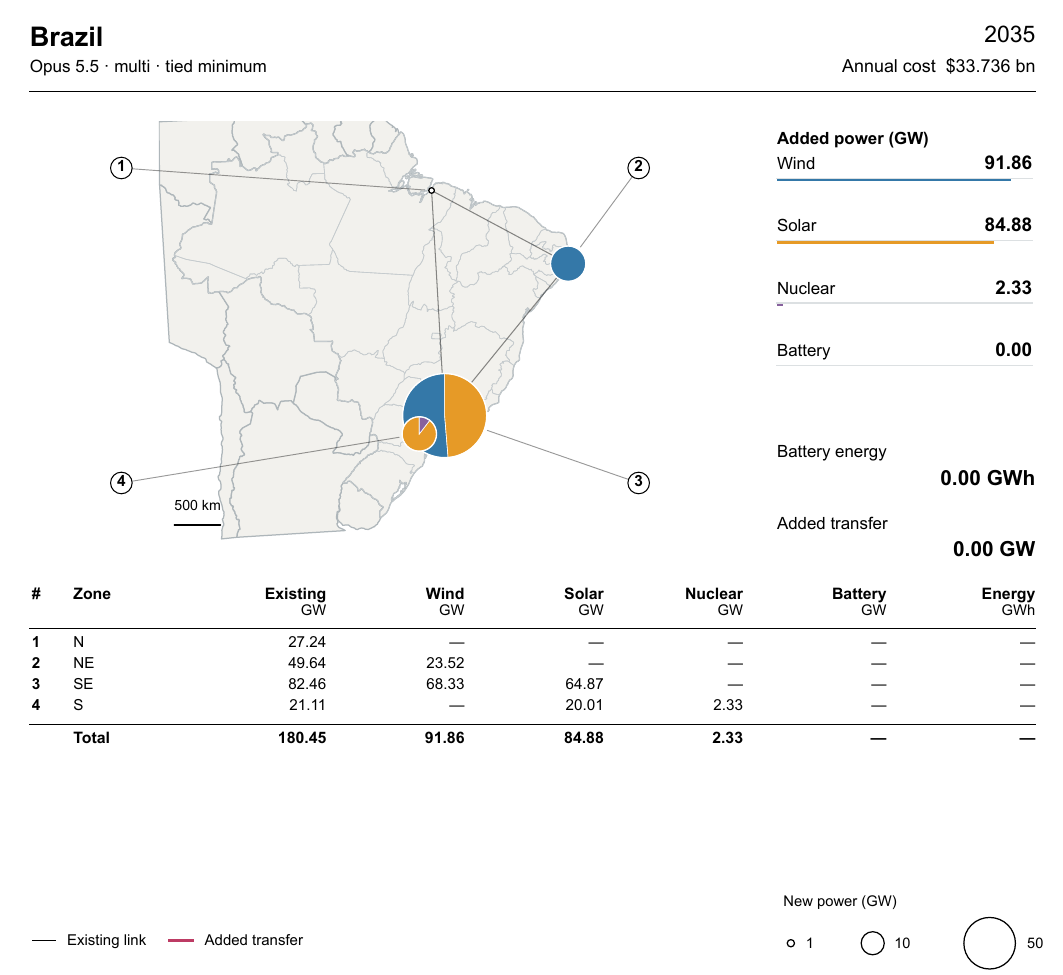}}
\caption{\textbf{Lowest-cost accepted 2035 design for Brazil.} Lowest observed selected total annual cost among four physically accepted trajectories using identical 2035 demand and NASA resource inputs. The displayed portfolio is Claude Opus 5.5, multi-agent, at USD 33.736 billion per year. All four trajectories have the same reported minimum cost; this portfolio is a stable representative of the tie. Map symbols follow Fig.~\ref{fig:geographic-planning}; numbers link to the zonal ledger. Existing is inherited generation; Energy is added battery energy; dashes are below 0.005 in the displayed units. No additional transfer capacity is selected. All represented feeder checks pass. Locations and corridors have model-zone resolution. The basemap uses Natural Earth.}
\label{fig:atlas-brazil}
\end{figure}
\clearpage
\section{Search histories and token usage for the loose-cap reference}\label{app:search-tokens}
\setcounter{figure}{0}
This section reports the ninety-six trajectories under the original 1,000~Mt annual ceiling.
Tables~\ref{tab:search-sol} and~\ref{tab:search-opus} retain every trajectory. Each paired entry is single/multi for the same system and year. Assessments include feeder-only rechecks; their selected index identifies the first occurrence of the retained candidate. Failed assessments fail the recorded distribution check. Probes are uncommitted system-only verifications without feeder acceptance. Repeated designs remain separate assessments. Input tokens include cache reads and writes, and output includes the reasoning subset. Full-precision counts, native round/turn indices, role attribution and response provenance are supplied in the frozen evidence archive.

\begin{tableorg}[htbp]
\caption{\textbf{Assessment histories and tokens for all 48 GPT-6.1 Sol trajectories.} Entries are single/multi. Input is in millions of tokens and output in thousands; displayed values are rounded. Selected is the source assessment index, not a native turn or cycle.}
\label{tab:search-sol}
\centering\small
\makebox[\textwidth][c]{\begin{tabular*}{170mm}{@{\extracolsep{\fill}}lrrrrrrr@{}}
\toprule
System & Year & Assessments & Selected & Failed & Probes & Input (M) & Output (k) \\
\midrule
MISO & 2023 & 3/1 & 1/1 & 0/0 & 0/2 & 0.582/0.509 & 2.133/11.211\\
 & 2030 & 8/1 & 8/1 & 0/0 & 0/2 & 3.131/0.453 & 4.114/10.040\\
 & 2035 & 11/1 & 11/1 & 0/0 & 0/2 & 6.279/0.454 & 4.805/10.811\\
\addlinespace[3pt]
ERCOT & 2023 & 3/1 & 1/1 & 0/0 & 0/2 & 0.691/0.504 & 2.802/10.468\\
 & 2030 & 13/18 & 13/17 & 6/6 & 0/20 & 9.287/23.873 & 8.033/276.833\\
 & 2035 & 12/13 & 10/13 & 5/3 & 0/21 & 7.897/15.469 & 5.235/188.850\\
\addlinespace[3pt]
NYISO & 2023 & 5/2 & 4/2 & 2/1 & 0/2 & 1.963/1.661 & 4.347/27.141\\
 & 2030 & 4/1 & 1/1 & 0/0 & 0/2 & 1.549/0.731 & 3.147/12.452\\
 & 2035 & 3/1 & 1/1 & 0/0 & 0/2 & 0.865/0.729 & 2.188/12.638\\
\addlinespace[3pt]
South Africa & 2023 & 8/14 & 8/14 & 4/5 & 0/11 & 5.379/19.473 & 6.205/189.135\\
 & 2030 & 10/15 & 10/15 & 3/3 & 0/18 & 8.130/24.199 & 8.328/221.683\\
 & 2035 & 12/12 & 12/12 & 4/3 & 0/18 & 9.852/18.279 & 8.143/188.018\\
\addlinespace[3pt]
Australia & 2023 & 4/1 & 1/1 & 0/0 & 0/2 & 0.746/0.317 & 2.300/10.856\\
 & 2030 & 3/1 & 1/1 & 0/0 & 0/2 & 0.443/0.400 & 1.955/11.656\\
 & 2035 & 8/2 & 6/2 & 0/0 & 0/4 & 2.455/1.045 & 3.996/24.353\\
\addlinespace[3pt]
New Zealand & 2023 & 3/1 & 1/1 & 0/0 & 0/2 & 0.370/0.276 & 2.235/10.645\\
 & 2030 & 5/1 & 1/1 & 0/0 & 0/2 & 0.965/0.346 & 2.729/11.050\\
 & 2035 & 4/1 & 1/1 & 0/0 & 0/2 & 0.577/0.296 & 1.744/10.541\\
\addlinespace[3pt]
Mexico & 2023 & 8/6 & 5/6 & 4/4 & 0/6 & 3.622/5.220 & 4.641/73.423\\
 & 2030 & 8/1 & 8/1 & 0/0 & 0/2 & 3.890/0.486 & 3.727/10.554\\
 & 2035 & 9/1 & 9/1 & 0/0 & 0/2 & 4.921/0.485 & 4.045/10.476\\
\addlinespace[3pt]
Brazil & 2023 & 4/1 & 1/1 & 0/0 & 0/2 & 0.521/0.256 & 1.563/10.022\\
 & 2030 & 4/1 & 1/1 & 0/0 & 0/2 & 0.539/0.273 & 1.829/10.824\\
 & 2035 & 2/1 & 1/1 & 0/0 & 0/2 & 0.203/0.273 & 1.423/11.110\\
\bottomrule
\end{tabular*}}
\end{tableorg}

\begin{tableorg}[htbp]
\caption{\textbf{Assessment histories and tokens for all 48 Opus 5.5 trajectories.} Entries are single/multi. Input is in millions of tokens and output in thousands; displayed values are rounded. Selected is the source assessment index, not a native turn or cycle.}
\label{tab:search-opus}
\centering\small
\makebox[\textwidth][c]{\begin{tabular*}{170mm}{@{\extracolsep{\fill}}lrrrrrrr@{}}
\toprule
System & Year & Assessments & Selected & Failed & Probes & Input (M) & Output (k) \\
\midrule
MISO & 2023 & 1/1 & 1/1 & 0/0 & 0/2 & 0.087/0.621 & 1.213/26.456\\
 & 2030 & 7/3 & 4/3 & 0/0 & 0/6 & 1.575/3.093 & 3.093/113.202\\
 & 2035 & 7/1 & 5/1 & 0/0 & 0/2 & 1.577/0.656 & 3.177/31.438\\
\addlinespace[3pt]
ERCOT & 2023 & 3/1 & 1/1 & 0/0 & 0/2 & 0.520/0.745 & 2.055/30.383\\
 & 2030 & 4/6 & 4/6 & 2/1 & 0/9 & 0.609/8.323 & 5.636/290.225\\
 & 2035 & 3/7 & 3/4 & 2/2 & 0/14 & 0.379/9.984 & 3.018/293.209\\
\addlinespace[3pt]
NYISO & 2023 & 2/2 & 2/2 & 1/1 & 0/1 & 0.458/2.303 & 2.211/66.623\\
 & 2030 & 1/1 & 1/1 & 0/0 & 0/2 & 0.230/0.898 & 2.282/31.254\\
 & 2035 & 1/1 & 1/1 & 0/0 & 0/2 & 0.229/0.898 & 1.543/36.060\\
\addlinespace[3pt]
South Africa & 2023 & 8/5 & 6/5 & 3/2 & 0/8 & 3.120/8.695 & 9.034/219.023\\
 & 2030 & 3/5 & 3/4 & 1/1 & 0/6 & 0.813/9.487 & 6.051/245.520\\
 & 2035 & 2/7 & 2/7 & 1/1 & 0/11 & 0.479/13.025 & 3.970/349.532\\
\addlinespace[3pt]
Australia & 2023 & 3/1 & 1/1 & 0/0 & 0/2 & 0.398/0.471 & 1.781/25.197\\
 & 2030 & 3/1 & 1/1 & 0/0 & 0/2 & 0.332/0.512 & 2.120/33.855\\
 & 2035 & 5/3 & 2/3 & 0/0 & 0/6 & 0.872/2.405 & 3.195/122.218\\
\addlinespace[3pt]
New Zealand & 2023 & 1/1 & 1/1 & 0/0 & 0/2 & 0.099/0.417 & 1.247/33.305\\
 & 2030 & 1/1 & 1/1 & 0/0 & 0/0 & 0.107/0.366 & 1.110/29.497\\
 & 2035 & 1/1 & 1/1 & 0/0 & 0/2 & 0.062/0.448 & 1.309/33.979\\
\addlinespace[3pt]
Mexico & 2023 & 4/4 & 4/4 & 2/2 & 0/7 & 0.591/4.643 & 2.874/157.947\\
 & 2030 & 4/1 & 1/1 & 0/0 & 0/2 & 0.740/0.704 & 2.061/26.903\\
 & 2035 & 6/1 & 4/1 & 0/0 & 0/2 & 1.390/0.699 & 3.143/27.783\\
\addlinespace[3pt]
Brazil & 2023 & 1/1 & 1/1 & 0/0 & 0/2 & 0.052/0.387 & 1.208/26.849\\
 & 2030 & 1/1 & 1/1 & 0/0 & 0/2 & 0.098/0.418 & 1.386/29.177\\
 & 2035 & 1/1 & 1/1 & 0/0 & 0/2 & 0.098/0.415 & 1.147/32.233\\
\bottomrule
\end{tabular*}}
\end{tableorg}

\clearpage
\section{Regional data source catalogue}\label{app:regional-sources}
\setcounter{figure}{0}
This catalogue groups the regional source documents by country. The Methods specify data definitions, model mappings and comparison boundaries. Preparation and comparison manifests retain extraction coordinates, workbook cells, source hashes and archived copies. Links identify the editions used in the experiments.

\paragraph{United States.} Demand and fleet records use \href{https://www.eia.gov/electricity/gridmonitor/about}{EIA-930} and \href{https://www.eia.gov/electricity/data/eia860/}{EIA-860}, with \href{https://www.ercot.com/news/mediakit/maps/index}{ERCOT weather-zone maps} and \href{https://www.nyiso.com/documents/20142/37320118/2023-NYCA-Generators.xlsx/145ca922-064c-133f-b3e8-3b4a30ed2845}{2023 NYCA generator-zone assignments}. Demand outlooks are \href{https://cdn.misoenergy.org/20260413\%20LTLF\%20Workshop\%20Supplemental\%20Materials\%20-\%20Regional750492.xlsx}{MISO 2026 LTLF regional supplementary data}, \href{https://www.ercot.com/files/docs/2025/04/08/ErcotAdjustedForecast.xlsb}{ERCOT-adjusted 2025 LTLF} and the \href{https://www.nyiso.com/documents/20142/57858057/2026-Gold-Book-Baseline-Forecast-Tables.xlsx/c8895dee-3d9f-0b3e-52fb-6a9ea5fea535}{2026 NYISO Gold Book forecast tables} and \href{https://www.nyiso.com/documents/20142/2226333/2026-Gold-Book-Public.pdf/a8fd42fe-5a8c-88cb-5052-f93dfd6423b8}{report}. NYISO network sources are the \href{https://www.nyiso.com/documents/20142/48283847/2024-RNA-Appendices.pdf/87c9ea6c-89eb-bcc0-a705-0d5ca17dd7df}{2024 RNA appendices}, \href{https://www.nyiso.com/documents/20142/3691300/Summer2025-OperatingStudy-APPROVED.pdf/3aaa5eb6-9fa4-53b5-181a-189e3ac52fcb}{Summer 2025 Operating Study} and \href{https://www.nyiso.com/documents/20142/2248481/2025-2034-Comprehensive-Reliability-Plan.pdf/61984e49-f7a2-eeda-bdd3-176c5ae40ba9}{final 2025--2034 CRP}. MISO limits use the \href{https://cdn.misoenergy.org/PY\%202025-2026\%20LOLE\%20Study\%20Report685316.pdf}{2025--2026 LOLE study}, \href{https://cdn.misoenergy.org/2025\%20PRA\%20Results\%20Posting\%2020250529_Corrections694160.pdf}{corrected PRA results} and \href{https://cdn.misoenergy.org/20250226\%20RASC\%20Item\%2003c\%20SRIC\%20SREC\%202025\%202026\%20PRA680238.pdf}{SRIC/SREC posting}; projects use \href{https://cdn.misoenergy.org/MTEP21\%20Addendum-LRTP\%20Tranche\%201\%20Report\%20with\%20Executive\%20Summary625790.pdf}{LRTP Tranche 1}, \href{https://cdn.misoenergy.org/LRTP\%20Tranche\%202.1666573.pdf}{Tranche 2.1} and \href{https://cdn.misoenergy.org/MTEP25\%20Chapter\%202\%20-\%20Regional_Long\%20Range\%20Transmission\%20Planning720398.pdf}{MTEP25 chapter 2}. ERCOT network assumptions refer to the \href{https://www.ercot.com/files/docs/2025/08/05/2025-Regional-Transmission-Plan-RTP-Economic-Study-Stability-Interface-Limits-Update-8_04_25-.pdf}{2025 RTP economic study}, \href{https://www.ercot.com/files/docs/2024/06/28/Reliability-Plan-For-The-Permian-Basin-Region.pdf}{Permian Basin reliability plan}, \href{https://ftp.puc.texas.gov/public/puct-info/agency/resources/pubs/news/2025/PUCT_Approves_Plan_for_the_First_Extra_High_Voltage_Transmission_Lines_in_ERCOT_Region.pdf}{PUCT 765 kV approval} and \href{https://www.ercot.com/files/docs/2025/12/01/14.1-25RPG025-AEP-Texas-CPS-Energy-Oncor-and-CNP-Texas-765-kV-STEP-Eastern-Backbone-Project.pdf}{STEP Eastern Backbone recommendation}.

\paragraph{South Africa.} Demand uses the \href{https://doi.org/10.25375/uct.26942134}{ESRG Hourly Demand Model v2025 by Province} and \href{https://github.com/GregoryIreland/Demand-Spatialization-South-Africa}{Demand Spatialization South Africa} corrected regional outputs. The spatialization notebook retains 2017 calibration inputs; the experimental base demand year is 2023. Fleet totals use the \href{https://www.eskom.co.za/heritage/wp-content/uploads/2024/03/2023_Annual_Report.pdf}{Eskom Integrated Report 2023} snapshot of 31 March 2023; global plant coordinates and allocation weights are described in Methods. Network records use \href{https://www.ntcsa.co.za/wp-content/uploads/2024/06/Shapefiles-1.zip}{NTCSA GCCA line shapefiles} and \href{https://www.ntcsa.co.za/wp-content/uploads/2025/02/TDP-2024-Public-Report_Rev2.pdf}{TDP 2025--2034 (TDP 2024), Public Report Rev.\,2}. The expert comparison uses \href{https://www.dee.gov.za/wp-content/uploads/2026/08/User-Friendly-IRP-2025.pdf}{User-Friendly IRP 2025, Table 1}, printed page 31 (PDF page 41), combining the end-2025 base with scheduled 2026--2035 additions while retaining separate distributed-generation, CSP and hybrid categories.

\paragraph{Australia.} The baseline uses \href{https://aemo.com.au/energy-systems/electricity/national-electricity-market-nem/data-nem/aggregated-data}{AEMO 2023 price-and-demand records} and the \href{https://www.aemo.com.au/-/media/files/electricity/nem/security_and_reliability/dispatch/policy_and_process/demand-terms-in-emms-data-model.pdf}{EMMS demand-term definitions}; five-minute \texttt{TOTALDEMAND} is averaged hourly. Fleet selection uses the archived \href{https://aemo.com.au/-/media/files/electricity/nem/participant_information/nem-registration-and-exemption-list.xlsx}{Registration and Exemption List} and \href{https://nemweb.com.au/Data_Archive/Wholesale_Electricity/MMSDM/2023/MMSDM_2023_12/MMSDM_Historical_Data_SQLLoader/DATA/PUBLIC_DVD_DUDETAILSUMMARY_202312010000.zip}{December 2023 MMSDM DUDETAILSUMMARY} records. Network assumptions use \href{https://www.aemo.com.au/-/media/files/electricity/nem/security_and_reliability/congestion-information/2024/interconnector-capabilities.pdf}{AEMO}'s \textit{Interconnector Capabilities} (\href{https://www.scribd.com/document/1051598892/Interconnector-Capabilities}{version 5 of 25 September 2025}) and \href{https://www.aemo.com.au/-/media/files/major-publications/isp/2024/appendices/a5-network-investments.pdf}{2024 ISP Appendix 5}. Demand growth uses \href{https://www.aemo.com.au/-/media/files/electricity/nem/planning_and_forecasting/nem_esoo/2026/2026-esoo-results-workbook.xlsx}{2026 NEM ESOO results, Step Change} with the \href{https://www.aemo.com.au/-/media/files/electricity/nem/planning_and_forecasting/nem_esoo/2026/2026-esoo-data-centre-forecasting-overview.pdf}{data-centre outlook}; financial-year growth is mapped to the calendar-2023 demand template as stated in Methods.

\paragraph{New Zealand.} Demand, fleet, hydro generation and connection records use the \href{https://www.emi.ea.govt.nz/Wholesale/Datasets}{Electricity Authority EMI wholesale datasets}. Network assumptions use the \href{https://static.transpower.co.nz/public/uncontrolled_docs/TPR\%202023\%20-\%20Complete.pdf}{Transmission Planning Report 2023}; the 1,400 MW northward HVDC target and assumed southward increment retain their distinct status. Future demand uses \href{https://static.transpower.co.nz/public/bulk-upload/documents/2026\%20SOSA\%20-\%20Final\%20Report.pdf?VersionId=Lp87J8gdqEEv1qiWPGOQrC6ax06BixFp}{Security of Supply Assessment 2026} and its \href{https://static.transpower.co.nz/public/bulk-upload/documents/2026\%20SOSA\%20-\%20Final\%20Supplementary\%20Data\%20-\%20Final\%20Version.xlsx?VersionId=U0jsFRXOz4d2ZmYgOFSYhlm9bRI.pgzm}{supplementary data}, Medium Demand April--September grid-exit-point offtake.

\paragraph{Mexico.} Baseline hourly demand uses \href{https://www.cenace.gob.mx/Paginas/SIM/Reportes/EstimacionDemandaReal.aspx}{CENACE daily real-demand settlement reports for 2023}. Fleet and network records use \href{http://base.energia.gob.mx/PRODESEN2024/prodesen24-38A1.PDF}{PRODESEN 2024--2038, Annex 1}; future demand uses \href{http://base.energia.gob.mx/PRODESEN2024/prodesen24-38cap3.PDF}{chapter 3, Table 3.5}, retaining the nine control-area mapping and the stated voltage-to-rating rule.

\paragraph{Brazil.} Baseline load, fleet, hydro generation and interchange records use \href{https://dados.ons.org.br/}{ONS open datasets}; \href{https://dadosabertos.aneel.gov.br/dataset/siga-sistema-de-informacoes-de-geracao-da-aneel}{ANEEL SIGA} supplies archived coordinates. Network and project descriptions use \href{https://www.ons.org.br/AcervoDigitalDocumentosEPublicacoes/Apresenta\%C3\%A7\%C3\%A3o\%20PEN\%202023.pdf}{PEN 2023} and \href{https://www.epe.gov.br/sites-pt/areas-de-atuacao/energia-eletrica/ArquivosGrupoEstudosTransmissao/EPE-DEE-RE-021-2023-rev0\%20-\%20Diagn\%C3\%B3stico\%20-\%20PDE\%202032\%20-\%20GET\%20Norte.pdf}{PDE 2032 GET Norte, Volume I}; the represented 4 GW HVDC bipole remains a model assumption. Future demand uses \href{https://www.epe.gov.br/sites-pt/publicacoes-dados-abertos/publicacoes/Documents/PDE\%202035_Relat\%C3\%B3rio\%20Final_Aprovado.pdf}{PDE 2035 final report, Table 2-3}. Capacity benchmarks use its \href{https://www.epe.gov.br/sites-pt/publicacoes-dados-abertos/publicacoes/Documents/PDE\%202035\_Dados\_Relat\%C3\%B3rio\%20Final.zip}{chapter-three workbook}, \texttt{Figura 3-25}, December 2035 row 8, retaining wind E8, centralized solar F8, separate MMGD G8 and batteries I8.

\paragraph{Operational carbon-budget sources.}

NYISO 2023 uses \href{https://www.rggi.org/sites/default/files/Uploads/Allowance-Tracking/2023\_Allowance-Distribution.xlsx}{RGGI Distribution of 2023 CO$_2$ Allowances}. sheet1, New York row 14; base budget B14, adjustment C14, adjusted budget D14. The archived source value is 27295284 US short tons CO$_2$/year; the adopted model ceiling is 24.761865118766~MtCO$_2$ per year.\par

NYISO 2030 uses \href{https://dec.ny.gov/sites/default/files/2026-08/part242\_et\_adopted.pdf}{New York adopted 6 NYCRR Part 242 Express Terms (2026 amendments)}. \S{}242-5.1(d); PDF p. 47 of 84. Short-ton definition: \S{}242-1.2 definition 71, PDF pp. 20--21. The archived source value is 13696558 US short tons CO$_2$/year; the adopted model ceiling is 12.425308408125~MtCO$_2$ per year.\par

NYISO 2035 uses \href{https://dec.ny.gov/sites/default/files/2026-08/part242\_et\_adopted.pdf}{New York adopted 6 NYCRR Part 242 Express Terms (2026 amendments)}. \S{}242-5.1(i); PDF p. 47 of 84. Short-ton definition: \S{}242-1.2 definition 71, PDF pp. 20--21. The archived source value is 3913302 US short tons CO$_2$/year; the adopted model ceiling is 3.550087857411~MtCO$_2$ per year.\par

MISO 2030 uses \href{https://cdn.misoenergy.org/20250924\%20Workshop\%20Item\%2002\%20Futures\%201-4\%20Preliminary\%20Results\%20update\%2020250923719306.pdf}{MISO Futures 1--4 Preliminary Results, September 24, 2025 workshop (updated September 23)}. Slide 66, Future 2 MISO aggregate constraint, year 2030; use exact table value rather than rounded slide 65 chart. The archived source value is 220499016 US short tons CO$_2$/year; the adopted model ceiling is 200.033342500216~MtCO$_2$ per year.\par

MISO 2035 uses \href{https://cdn.misoenergy.org/20250924\%20Workshop\%20Item\%2002\%20Futures\%201-4\%20Preliminary\%20Results\%20update\%2020250923719306.pdf}{MISO Futures 1--4 Preliminary Results, September 24, 2025 workshop (updated September 23)}. Slide 66, Future 2 MISO aggregate constraint, year 2035; use exact table value rather than rounded slide 65 chart. The archived source value is 181486560 US short tons CO$_2$/year; the adopted model ceiling is 164.641837747094~MtCO$_2$ per year.\par

South Africa 2030 uses \href{https://www.gov.za/sites/default/files/gcis\_document/202510/53596gon6767.pdf}{South Africa Integrated Resource Plan 2025, Government Gazette 53596, Notice 6767}. Figure 21, printed p. 35 / PDF p. 44, balanced-plan electricity CO$_2$ trajectory, 2030. The archived source value is 168 MtCO$_2$/year; the adopted model ceiling is 168.000000000000~MtCO$_2$ per year.\par

South Africa 2035 uses \href{https://www.gov.za/sites/default/files/gcis\_document/202510/53596gon6767.pdf}{South Africa Integrated Resource Plan 2025, Government Gazette 53596, Notice 6767}. Figure 21, printed p. 35 / PDF p. 44, balanced-plan electricity CO$_2$ trajectory, 2035. The archived source value is 142 MtCO$_2$/year; the adopted model ceiling is 142.000000000000~MtCO$_2$ per year.\par

Mexico 2030 uses \href{https://sidof.segob.gob.mx/notas/docFuente/5776838}{Mexico Programa Sectorial de Energía 2025--2030 (PROSENER)}. Electricity-generation emissions paragraph: net-injected-electricity emissions peak in 2027, then decline 23.9\% from 156.42 to 119 MtCO$_2$e by the 2030 planning horizon. The archived source value is 119 MtCO$_2$e/year; the adopted model ceiling is 119.000000000000~MtCO$_2$ per year.\par

Brazil 2030 uses \href{https://www.epe.gov.br/sites-pt/publicacoes-dados-abertos/publicacoes/Documents/PDE\%202035\_Relat\%C3\%B3rio\%20Final\_Aprovado.pdf}{Brazil EPE/MME Plano Decenal de Expansão de Energia 2035, approved final report}. Table 10-1, p. 393, SIN row, 2030; scope/hydrology discussion pp. 394--395 and explicit absence of model GHG limits p. 396. The archived source value is 27 MtCO$_2$e/year; the adopted model ceiling is 27.000000000000~MtCO$_2$ per year.\par

Brazil 2035 uses \href{https://www.epe.gov.br/sites-pt/publicacoes-dados-abertos/publicacoes/Documents/PDE\%202035\_Relat\%C3\%B3rio\%20Final\_Aprovado.pdf}{Brazil EPE/MME Plano Decenal de Expansão de Energia 2035, approved final report}. Table 10-1, p. 393, SIN row, 2035; scope/hydrology discussion pp. 394--395 and explicit absence of model GHG limits p. 396. The archived source value is 29 MtCO$_2$e/year; the adopted model ceiling is 29.000000000000~MtCO$_2$ per year.\par

\end{appendices}
\clearpage

\setlength{\bibsep}{0.8em}
\bibliography{sn-bibliography}
\end{document}